\documentclass[12pt]{article}
\usepackage{amsmath}
\usepackage{amssymb}
\usepackage{graphicx}
\usepackage{enumerate}
\usepackage[authoryear]{natbib}
\usepackage{url}
 \usepackage{dsfont}
 
\usepackage[toc,page]{appendix}
 
\RequirePackage{amsthm,mathrsfs,multirow,morefloats,booktabs,subfigure,pdfpages,color,algorithm,array,hhline,makecell,epstopdf,algorithm}
\RequirePackage[colorlinks,citecolor=blue,urlcolor=blue]{hyperref}
\RequirePackage{algpseudocode}
\algrenewcommand\algorithmicrequire{\textbf{Input:}}
\algrenewcommand\algorithmicensure{\textbf{Output:}}

\newcommand{\blind}{1}

\numberwithin{equation}{section}
\theoremstyle{plain}
\newtheorem{thm}{Theorem}[section]

\newtheorem{lem}[thm]{Lemma}

\begin{document}
	\def\spacingset#1{\renewcommand{\baselinestretch}%
		{#1}\small\normalsize} \spacingset{1}

	%%%%%%%%%%%%%%%%%%%%%%%%%%%%%%%%%%%%%%%%%%%%%%%%%%%%%%%%%%%%%%%%%%%%%%%%%%%%%%
	
\if1\blind
	{
		\title{\bf Dual regularized Laplacian  spectral clustering methods on community detection}%Applications of dual regularized Laplacian matrix for community detection}
		\author{Huan Qing\\%\thanks{
				%The authors gratefully acknowledge \textit{please remember to list all relevant funding sources in the unblinded version}}\hspace{.2cm}\\
			Department of Mathematics, China University of Mining and Technology\\
			and \\
			Jingli Wang \\
			School of Statistics and Data Science, Nankai University\\
		Email: jlwang@nankai.edu.cn}
		\maketitle
	} \fi
	
	\if0\blind
	{
		\bigskip
		\bigskip
		\bigskip
		\begin{center}
			{\LARGE\bf Applications of dual regularized Laplacian matrix for community detection}
		\end{center}
		\medskip
	} \fi
	
	\bigskip
	\begin{abstract}
		Spectral clustering methods are widely used for detecting clusters in networks for community detection, while a small change on the graph Laplacian matrix could bring a dramatic improvement. In this paper, we propose a dual regularized graph Laplacian matrix and then employ it to three classical spectral clustering approaches  under the degree-corrected stochastic block model. If the number of  communities is known as $K$, we consider more than $K$ leading eigenvectors and weight them by their corresponding eigenvalues in the spectral clustering procedure to improve the performance.  Three improved spectral clustering methods are  dual regularized spectral clustering (DRSC) method, dual regularized spectral clustering on Ratios-of-eigenvectors (DRSCORE) method, and dual regularized symmetrized Laplacian inverse matrix (DRSLIM) method.
		Theoretical analysis of DRSC  and DRSLIM show that under mild conditions DRSC and DRSLIM yield stable consistent community detection, moreover, DRSCORE returns perfect clustering under the ideal case.  We compare the performances of DRSC, DRSCORE and DRSLIM with several spectral methods by substantial simulated networks and eight real-world networks.
	\end{abstract}
	
	\noindent%
	{\it Keywords:}  Spectral clustering; degree-corrected stochastic block model; Network structure data %3 to 6 keywords, that do not appear in the title
	\vfill
	
	\newpage
	%\spacingset{1.5} % DON'T change the spacing!
\section{Introduction}
The network has inadvertently penetrated into every aspect of our life, such as friendship networks, work networks, social networks, biochemical networks and so on. In past few decades, researchers have developed a variety of approaches to explore network structures for understanding more underlying mechanisms \citep{lorrain1971structural, burt1976positions, doreian1985structural, doreian1994partitioning, borgatti1997network}. Recently, the community detection problem appeals to the ascending number of attention in computer science, physics and statistics \citep{DCSBM, MN2006, nepusz2008fuzzy, PJBAC, SCORE}.  The stochastic blockmodel (SBM)  \citep{SBM} is a well-known generative model for communities in networks, and it also be taken as a posteriori blockmodeling to discover the network structure. In the SBM, the nodes within each community are assumed to have the same expected degrees, that is to say, the distribution of degrees within each community follows a Poisson distribution. Under this model, one can easily produce abundant different network structures  \citep{snijders1997estimation,daudin2008a,bickel2009a}. Unfortunately, due to the restrictive assumptions of vertices, the simple blockmodel does not work well in many applications to real-world networks \citep{DCSBM}. In fact, there are some extensions of SBM to make it more flexible to empirical networks. For instance, \cite{wang1987stochastic} constructed a blockmodel for directed networks with arbitrary expected degrees; \cite{reichardt2007role} proposed a density-based stochastic block model for complex networks; \cite{airoldi2008mixed} introduced a mixed membership stochastic blockmodel based on hierarchical Bayes; \cite{DCSBM} established the degree-corrected stochastic blockmodel (DCSBM) by assuming the degrees follow a power-law distribution, which allows the degree varies among different communities.
Doubtless, DCSBM is the most popular one of the generalizations of SBM.  And a number of methods have been designed based on DCSBM, such as, \cite{newman2016structure, SCORE, RSC,OCCAM,CMM,LSCD}.  In this paper, we construct spectral clustering methods under the framework of DCSBM.

The spectral clustering algorithm is a fast and efficient method for community detection. In general, assuming there are $K$ clusters, a spectral clustering method usually has four steps \citep{Tutorial}: (1) obtain the adjacency matrix or its variants of the given network; (2) compute the first leading $K$ eigenvectors of the adjacency matrix or its variants; (3) combine eigenvectors by column to a matrix; (4) apply clustering methods (e.g. K-means clustering, K-median clustering) to the row vectors of the eigenvector matrix to detect communities. Some algorithms apply a normalizing or regularizing procedure for the graph matrix, and a scaling-invariant mapping to each row of eigenvector matrix to reduce the heterogeneity \citep{ng2002spectral,Tutorial, SCORE}.  Under DCSBM, substantial spectral clustering algorithms are developed recently. The regularized spectral clustering (RSC) method \citep{RSC} is established based on the general spectral clustering method with a regularized graph Laplacian matrix and normalized eigenvector matrix. From RSC we can find that a small change of graph matrix can lead to dramatically better results. \cite{SCORE} proposed a spectral clustering method which is called SCORE in short, using the entry-wise ratios between the first leading eigenvector and each of the other leading eigenvectors for clustering.  Recently, \cite{SLIM} employed a symmetrized Laplacian inverse matrix (SLIM) on the spectral clustering method to measure the closeness between nodes. In fact, the procedures for most spectral clustering methods are similar, but the graph Laplacian matrices vary. In this paper, we focus on the construction of the graph Laplacian matrix to improve the performance of spectral clustering methods for community detection.

For a network, the Laplacian matrix plays a key role since it is consistent with the matrix in spectral geometry and random walks \citep{chung1997spectral}. Many people think the quantities based on this matrix may more faithfully reflect the structure and properties of a graph \citep{chen2007resistance}. Actually, a number of authors have constructed algorithms based on the (regularized/normalized) Laplacian matrix. For example, \cite{RSC} proposed a spectral clustering method with a regularized Laplacian matrix; \cite{SCORE+} have discussed the reason why the regularized Laplacian matrix should be used as a pre-procedure for their PCA-based clustering method; \cite{dong2016learning} studied the smooth graph signal representation with a Laplacian matrix. Moreover, there are also many attractive methods established based on the dual Laplacian matrix. \cite{yankelevsky2016dual} applied dual Laplacian regularization to dictionary learning.  \cite{xiao2018graph} and \cite{tang2019dual}  developed biological methods based on dual Laplacian regularization microRNA-disease associations prediction. It is worth to mention that there are a variety of forms for Laplacian matrices, and we should be careful to chose the suitable one for using \citep{von2007tutorial, chaudhuri2012spectral}. The motivation of this paper comes from the fact that \cite{amini2013pseudo,RSC, joseph2016impact} study that regularization on the Laplacian matrix can improve the performance of spectral clustering, therefore a question comes naturally, whether spectral clustering approaches designed based on the dual regularization or even multiple regularization on the Laplacian matrix can give satisfactory performances. In this paper we construct our detection methods based on a dual regularized Laplacian matrix. However, our proposed dual regularized Laplacian methods differ from the above mentioned dual Laplacian methods. In our dual Laplacian methods, we first build a regularized Laplacian matrix on the adjacency matrix and then reapply regularized Laplacian on the first Laplacian matrix, while other dual Laplacian approaches just use two Laplacian matrix in a same function/model.

% In the final step of spectral clustering methods, k-means algorithm is a popular choice, such as \cite{SCORE,RSC,von2007tutorial}. However, it is not the only choice. Instead of k-means, one can also use other techniques to construct the final solution for clustering, such as, \cite{bach2004learning,lang2006fixing}. In this paper we just use the k-means method to extract the correct clusters in the last step.

In this paper we propose three spectral clustering methods to detect communities. The three spectral clustering methods are designed based on combining the dual regularized Laplacian matrix $L_{\tau_{2}}$ (which is defined in section \ref{sec3}) with two traditional spectral clustering methods and one recent spectral clustering method. Next we briefly introduce our three approaches one by one. For the dual regularized spectral clustering (DRSC) method:  as mentioned above we first construct a dual regularized Laplacian matrix to reduce the noise of the adjacency matrix of a network, and then unlike other spectral clustering methods, we compute the $(K+1)$ leading eigenvalues and its corresponding eigenvectors with unit-norm on the obtained dual regularized Laplacian matrix. After a row-normalization step aiming at reducing the noise caused by degree heterogeneity, the final estimated clusters are determined by K-means. In fact, the dual regularized Laplacian can be taken as the foundation of this proposed method DRSC and our DRSC is designed based on the traditional regularized spectral clustering (RSC) method \citep{RSC}. For our dual regularized spectral clustering on ratios-of-eigenvectors (DRSCORE) method which is designed based on the spectral clustering on ratios-of-eigenvectors (SCORE) method \citep{SCORE}: after obtaining the production of the leading $(K+1)$ eigenvectors and eigenvalues of $L_{\tau_{2}}$, there is a step to obtain the ratio of entry-wise matrix and then apply K-means to this matrix for clustering.  %In fact, our DRSCORE is designed based on the application of $L_{\tau_{2}}$ on the traditional spectral clustering on ratios-of-eigenvectors (SCORE) method \cite{SCORE}.
 For our  dual regularized symmetrized Laplacian inverse matrix (DRSLIM) method which is based on the recent symmetric Laplacian inverse matrix (SLIM) method \citep{SLIM}: we obtain the symmetric Laplacian inverse matrix based on $L_{\tau_{2}}$, then apply K-means on the row-normalization of the production of the leading $(K+2)$ eigenvectors and eigenvalues of the symmetric Laplacian inverse matrix. %In fact, our DRSLIM is designed based on the application of $L_{\tau_{2}}$ on the recent symetric Laplacian inverse matrix (SLIM) method \cite{SLIM}.
More importantly, our proposed method use more than $K$ eigenvalues and eigenvectors, which enables that DRSC, DRSCORE and DRSLIM could deal with some weak signal networks, such as Simmons and Caltech where the two weak signal networks are discussed in \cite{SCORE+}. In the numerical studies, we also construct other multiple regularized spectral clustering methods with multiple regularized Laplacian. Unfortunately, their performances are not as good as DRSC and DRSLIM.

In Section \ref{sec2}, we set up the community detection problem under DCSBM. In Section \ref{sec3}, we propose our three methods DRSC, DRSCORE and DRSLIM. Section \ref{sec4} presents theoretical framework of these three approaches where we show the consistency of DRSC and DRSLIM, and we provide population analysis for DRSCORE. Section \ref{sec5} investigates the performances of DRSC, DRSCORE and DRSLIM via comparing with four spectral clustering methods on both numerical networks and eight empirical data sets. Section \ref{sec5} also studies the effect of different choice of the two regularizers $\tau_{1}$ and $\tau_{2}$ on the performance of DRSC. Meanwhile, we also compare the performances of DRSC, DRSCORE and DRSLIM with the multiple regularization spectral clustering methods MRSC, MRSCORE and MRSLIM in Section \ref{sec5}, respectively. Section \ref{sec6} concludes.
%\newpage
\section{Problem setup}\label{sec2}
%In this section, we set up the community detection problem under DCSBM, and then describe our algorithm DRSC.

The following notations will be used throughout the paper: $\|\cdot\|_{F}$ for a matrix denotes the Frobenius norm,  $\|\cdot\|$ for a matrix denotes the spectral norm,  and $\|\cdot\|$ for a vector denotes the $l_{2}$-norm. For convenience, when we say ``leading eigenvalues'' or ``leading eigenvectors'', we are comparing the \emph{magnitudes} of the eigenvalues and their respective eigenvectors with unit-norm. For any matrix or vector $x$, $x'$ denotes the transpose of $x$.

Consider an undirected, no-loops, and un-weighted connected network $\mathcal{N}$ with $n$ nodes and let $A$ be its adjacency matrix such that $A_{ij}=1$ if there is an edge between node $i$ and $j$, $A_{ij}=0$ otherwise. Since there is no-loops in $\mathcal{N}$, all diagonal entries of $A$ are zero. Let $\mathcal{C}$ denote the set containing all nodes in $\mathcal{N}$, assume that there exist $K$ disjoint clusters $\mathcal{C}^{(1)}, \mathcal{C}^{(2)}, \ldots, \mathcal{C}^{(K)}$ and each node belongs to exactly one cluster (i.e., $\mathcal{C}=\cup_{i=1}^{K}\mathcal{C}_{i}$, and $\mathcal{C}_{i}\cap\mathcal{C}_{j}=\emptyset$ for any distinct $i,j$ such that $1\leq i,j\leq K$).  The number of clusters $K$ is assumed to be known. Let $\ell$ be an $n\times 1$ vector such that $\ell(i)$ takes values from $\{1, 2, \ldots, K\}$ and $\ell(i)$ is the true community label for node $i$. In this paper, our goal is designing spectral clustering algorithms via applying information by the given $(A, K)$ of the network $\mathcal{N}$ to estimate $\ell$.

In this paper, we consider the degree-corrected stochastic blockmodel (DCSBM). Under DCSBM, an $n\times 1$ degree heterogeneity vector $\theta = (\theta_{1}, \cdots, \theta_{n})^\prime$ is introduced to control the node degrees, where $\theta_{i}>0$ for each node $i$, $i = 1, \cdots, n$. To facilitate theoretical analysis, assume that all elements of $\theta$ are in $[0, 1]$. Under the DCSBM model, the probability of generating an edge between node $i$ and node $j$ is assumed to follow a Bernoulli distribution such that $\mathrm{Pr}(A_{ij}=1)=\theta_{i}\theta_{j}P_{g_{i}g_{j}}$ where $P$ is a $K\times K$ symmetric matrix with full rank (called mixing matrix) and its entries are in $[0,1]$, and $g_{i}$ denotes the cluster that node $i$ belongs to (i.e., $g_{i}=\ell(i)$). %In this paper, $P$ is symmetric with full rank.
% Let $P$ be a $K\times K$ symmetric matrix whose elements are the probability of generating an edge between distinct nodes and the entries are in $[0,1]$, and we call $P$ mixing matrix for convenience.  To facilitate theoretical analysis, assume that all elements of $\theta$ are in $[0, 1]$  and $P$ is symmetric with full rank in this paper. Under the DCSBM model, the probability of generating an edge between node $i$ and node $j$ is assumed to follow a Bernoulli distribution such that $\mathrm{Pr}(A_{ij}=1)=\theta_{i}\theta_{j}P_{g_{i}g_{j}}$ where $g_{i}$ denotes the cluster that node $i$ belongs to (i.e., $g_{i}=\ell(i)$). When $g_{i}=g_{j}$ for any two distinct nodes $i,j$ such that $\theta_{i}=\theta_{j}$, DCSBM degenerates to SBM.
%
%Under DCSBM, since $\mathrm{Pr}(A_{ij}=1)$ follows a Bernoulli distribution with probability $\theta_{i}\theta_{j}P_{g_{i}g_{j}}$, the expectation of $\mathrm{Pr}(A_{ij}=1)$ is $\theta_{i}\theta_{j}P_{g_{i}g_{j}}$.
Then we define the expectation matrix of the adjacency matrix $A$ as $\Omega \overset{\bigtriangleup}{=}\mathop{{}\mathbb{E}}[A]$ such that $\Omega_{ij}=\mathrm{Pr}(A_{ij}=1)=\theta_{i}\theta_{j}P_{g_{i}g_{j}}$. From \cite{SCORE, DCSBM} and \cite{ RSC}, $\Omega$ can be expressed as
\begin{align*}
  \Omega=\Theta ZPZ'\Theta,
\end{align*}
where the $n\times n$ diagonal matrix $\Theta$'s $i$-th diagonal element is $\theta_{i}$ and
the $n\times K$ membership matrix $Z$ directly contains information about the true nodes labels such that  $Z_{ik}=1$ if and only if node $i$ belongs to block $k$ (i.e., $g_{i}=k$), otherwise $Z_{ik}=0$.

Given $(n, P, \Theta, Z)$, we can generate the random adjacency matrix $A$ under DCSBM, therefore we denote the DCSBM model as $DCSBM(n, P, \Theta, Z)$ for convenience in this paper. For community detection under DCSBM, spectral clustering algorithms are always designed based on analyzing the properties of $\Omega$ or its variants since it can be expressed by the true membership matrix $Z$.
%\section{The algorithms: DRSC, DRSCORE and DRSLIM}\label{sec3}
\section{Methodology and algorithms}\label{sec3}
In this section, under DCSBM we first introduce a dual regularized Laplacian matrix based spectral clustering method in general case. And then we apply our dual regularization ideology to three published spectral clustering methods to improve their performances. These published spectral clustering methods are RSC \citep{RSC}, SCORE \citep{SCORE} and SLIM \citep{SLIM},  and their refinements are called dual regularized spectral clustering (DRSC for short), dual regularized spectral clustering on ratios-of-eigenvectors (DRSCORE for short) and dual regularized symmetrized Laplacian inverse matrix (DRSLIM for short).% as refinements of three typical spectral clustering approaches RSC \citep{RSC}, SCORE \citep{SCORE} and SLIM \citep{SLIM}, without losing their appealing features.
\subsection{Dual regularized Laplacian-based spectral clustering}\label{DualLSC}
The regularized Laplacian matrix is defined as
\begin{align*}
  L_{\tau_{1}}=D_{\tau_{1}}^{-1/2}AD_{\tau_{1}}^{-1/2},
\end{align*}
where $D_{\tau_{1}}=D_{1}+\tau_{1}I$, $D_{1}$ is an $n\times n$ diagonal matrix whose $i$-th diagonal entry is given by $D_{1}(i,i)=\sum_{j}A(i,j)$, $I$ is an $n\times n$ identity matrix, and the regularizer  $\tau_{1}$ is a nonnegative number.

The dual regularized Laplacian matrix is defined as
\begin{align*}
  L_{\tau_{2}}=D_{\tau_{2}}^{-1/2}L_{\tau_{1}}D_{\tau_{2}}^{-1/2},
\end{align*}
where $D_{\tau_{2}}=D_{2}+\tau_{2}I$, $D_{2}$ is an $n\times n$ diagonal matrix whose $i$-th diagonal entry is given by $D_{2}(i,i)=\sum_{j}L_{\tau_{1}}(i,j)$, and the regularizer $\tau_{2}$ is nonnegative\footnote{Note that in this paper, without causing confusion, matrices or vectors with subscript 1 or $\tau_{1}$ are always computed based on the regularized Laplacian matrix  or the population version of the regularized Laplacian matrix in next sections, while matrices or vectors with subscript 2 or $\tau_{2}$ are always computed based on the dual regularized Laplacian matrix  or the population version of the dual regularized Laplacian matrix in next sections.}.

In traditional spectral clustering methods, once people have adjacent matrix or its variants, we should compute the leading $K$ eigenvectors and combine them by column to a matrix, and then one may normalize the matrix by row or without this normalizing procedure. Finally, K-means is applied to the eigenvector-matrix to detect communities. Different from the traditional methods, we employ the weighted $(K+K_0)$ leading eigenvectors of the dual regularized Laplacian matrix, where $K_0$ is a positive integer. In fact the idea of more than $K$ leading eigenvectors is first proposed by \cite{SCORE+} since they find that the $(K+1)$-th eigenvector may contain some label information for some weak signal networks. In their method they apply the $K$ or $(K+1)$ leading eigenvectors depending on some conditions, but we release their conditions and directly use these $(K+K_0)$ leading eigenvectors for all cases.
In this paper, unless specified, let $\{\hat{\lambda}_{i}\}_{i=1}^{K+K_0}$ be the leading $(K+K_0)$ eigenvalues of $L_{\tau_{2}}$, and $\{\hat{\eta}_{i}\}_{i=1}^{K+K_0}$ be the respective eigenvectors with unit-norm. Then we construct a weighted eigenvector matrix $\hat{X}$ based on $L_{\tau_{2}}$ (or $\check{X}$ based on $M$ defined in Section \ref{DRSLIM}), where the weights are their corresponding eigenvalues. This matrix can be presented as $\hat{X}=[\hat{\eta}_{1},\hat{\eta}_{2}, \ldots, \hat{\eta}_{K}, \hat{\eta}_{K+K_0}]\cdot \mathrm{diag}(\hat{\lambda}_{1}, \hat{\lambda}_{2}, \ldots, \hat{\lambda}_{K}, \hat{\lambda}_{K+K_0})$. $\check{X}$ shares similar forms as $\hat{X}$ except that it consists of eigenvectors and eigenvalues of $M$. The penultimate step is normalizing $\hat{X}$ and $\check{X}$ by row or by entry-wise ratio. Finally, K-means method is applied to the normalization versions of $\hat{X}$ and $\check{X}$ to detect the communities.

There are 3 key points in our proposed methods: one is the idea of dual regularized Laplacian matrix; one is using the first $(K+K_0)$ leading eigenvectors; one is taking the eigenvalues as weights for the eigenvectors. These three points contribute a lot to the performances of spectral clustering methods. Note that the $(K+1)$ or $(K+2)$ leading eigenvectors are enough for detecting strong and weak signal networks. For example, we use $(K+1)$ leading eigenvectors for DRSC and DRSCORE algorithms, and present the algorithm of DRSLIM with $(K+2)$ leading eigenvectors.

\subsection{The DRSC algorithm}
The detail of the DRSC method proceeds as in Algorithm \ref{alg:DRSC}.
\begin{algorithm}
\caption{\textbf{Dual Regularized Spectral Clustering algorithm} (\textbf{DRSC})}
\label{alg:DRSC}
\begin{algorithmic}[1]
\Require $A,K$ and regularizers $\tau_{1}, \tau_{2}$.
\Ensure node labels $\hat{\ell}$
\State Compute $L_{\tau_{1}}$ (the default $\tau_{1}$ for DRSC is the average degree, i.e., $\tau_{1}=\sum_{i,j}A(i,j)/n$).
\State Compute $L_{\tau_{2}}$ (the default $\tau_{2}$ for DRSC is $\tau_{2}=\sum_{i,j}L_{\tau_{1}}(i,j)/n$).
\State Compute the $n\times (\textbf{K+1})$ matrix
$\hat{X}=[\hat{\eta}_{1},\hat{\eta}_{2}, \ldots, \hat{\eta}_{K}, \hat{\eta}_{K+1}]\cdot \mathrm{diag}(\hat{\lambda}_{1}, \hat{\lambda}_{2}, \ldots, \hat{\lambda}_{K}, \hat{\lambda}_{K+1})$.
\State  Compute $\hat{X}^{*}$ by normalizing each of $\hat{X}$'s rows to have unit length.
\State Treat each row of $\hat{X}^{*}$ as a point in $\mathcal{R}^{K+1}$, and apply K-means to $\hat{X}^{*}$ with $K$ clusters to obtain $\hat{\ell}$.
\end{algorithmic}
\end{algorithm}

%By analyzing the five steps of DRSC, we can find that our DRSC differs from traditional spectral clustering methods such as SCORE and RSC in the following aspects a) compared with RSC, our DRSC applies the eigenvectors and eigenvalues of the so-called dual Laplacian matrix $L_{\tau_{2}}$ to construct $\hat{X}$. b) compared with RSC and SCORE, our DRSC always apply the production of the leading (K+1) eigenvectors and the leading (K+1) eigenvalues to construct $\hat{X}$, instead of simply applying the leading $K$ eigenvectors as RSC and SCORE. c) though there are two ridge regularizers $\tau_{1}$ and $\tau_{2}$ in our DRSC, numerical results on eight real-world datasets in Section \ref{secreal8} shows that our DRSC is insensitive to the choice of $\tau_{1}$ and $\tau_{2}$. For convenience, unless specified, the default values for $\tau_{1}$ and $\tau_{2}$ for DRSC are set as in Algorithm \ref{alg:DRSC} in this paper.
 Though there are two ridge regularizers $\tau_{1}$ and $\tau_{2}$ in our DRSC, numerical results on eight real-world datasets in Section \ref{secreal8} shows that our DRSC is insensitive to the choice of $\tau_{1}$ and $\tau_{2}$. For convenience, unless specified, the default values for $\tau_{1}$ and $\tau_{2}$ for DRSC are set as in Algorithm \ref{alg:DRSC} in this paper.
\subsection{The DRSCORE algorithm}
The detail of the DRSCORE method proceeds as in Algorithm \ref{alg:DRSCORE}.
\begin{algorithm}
\caption{\textbf{Dual Regularized Spectral Clustering On Ratios-of-Eigenvectors algorithm} (\textbf{DRSCORE})}
\label{alg:DRSCORE}
\begin{algorithmic}[1]
\Require $A,K$ and regularizers $\tau_{1}, \tau_{2}$.
\Ensure node labels $\hat{\ell}$
\State Compute $L_{\tau_{1}}$ (the default $\tau_{1}$ for DRSCORE is $\tau_{1}=\sum_{i,j}A(i,j)$).
\State Compute $L_{\tau_{2}}$ (the default $\tau_{2}$ for DRSCORE is $\tau_{2}=\frac{\sum_{i,j}L_{\tau_{1}}(i,j)}{nK}$).
\State Compute
$\hat{X}=[\hat{\eta}_{1},\hat{\eta}_{2}, \ldots, \hat{\eta}_{K}, \hat{\eta}_{K+1}]\cdot \mathrm{diag}(\hat{\lambda}_{1}, \hat{\lambda}_{2}, \ldots, \hat{\lambda}_{K}, \hat{\lambda}_{K+1})$. Set $\hat{X}_{i}$ as the $i$-th column of $\hat{X}, 1\leq i\leq K+1$.
\State Compute the $n\times \textbf{K}$ matrix $\hat{R}$  of entry-wise eigen-ratios such that
$\hat{R}(i,k)=\hat{X}_{k+1}(i)/\hat{X}_{1}(i), 1\leq i\leq n, 1\leq k\leq K$.
\State Treat each row of $\hat{R}$ as a point in $\mathcal{R}^{K}$, and apply K-means to $\hat{R}$ with $K$ clusters to obtain $\hat{\ell}$.
\end{algorithmic}
\end{algorithm}

%By analyzing the four steps of DRSCORE, we can find that our DRSCORE differs from SCORE in the following aspects a) our DRSCORE always apply the production of the leading (K+1) eigenvectors and the leading (K+1) eigenvalues of $L_{\tau_{2}}$ to construct $\hat{R}$, instead of simply applying the leading $K$ eigenvectors of $A$ to construct $\hat{R}$ as SCORE does. b)
 In the original SCORE method, there is a threshold parameter $T_{n}$ (the default $T_{n}$ is $\mathrm{log}(n)$) to control the eigen-ratios. In our DRSCORE, we release this condition. By Lemma 2.5 in \cite{SCORE}, since we only consider connected network in this paper, $\hat{\lambda}_{1}$ is nonzero and all elements of $\hat{\eta}_{1}$ are nonzero, which means that $\hat{\lambda}_{1}\hat{\eta}_{1}(i)$ can be in the denominator, therefore the eigen-ratio matrix $\hat{R}$ is well defined.
 Meanwhile, there are two regularizers $\tau_{1}$ and $\tau_{2}$ in DRSCORE. Though numerical results in Section \ref{secreal8} show that our DRSCORE is sensitive to the choice of $\tau_{1}$ and $\tau_{2}$, as long as we set $\tau_{1}=\sum_{i,j}A(i,j)$ and $\tau_{2}=\frac{\sum_{i,j}L_{\tau_{1}}(i,j)}{nK}$, our DRSCORE has excellent performances and always outperforms SCORE.  For convenience, unless specified, the default values for $\tau_{1}$ and $\tau_{2}$ for DRSCORE are set as in Algorithm \ref{alg:DRSCORE} in this paper.
\subsection{The DRSLIM algorithm}\label{DRSLIM}
The detail of the DRSLIM method proceeds as in Algorithm \ref{alg:DRSLIM}.
\begin{algorithm}
\caption{\textbf{Dual Regularized Symmetrized Laplacian Inverse Matrix algorithm} (\textbf{DRSLIM})}
\label{alg:DRSLIM}
\begin{algorithmic}[1]
\Require $A,K$ and regularizers $\tau_{1}, \tau_{2}$ and tuning parameter $\gamma$.
\Ensure node labels $\hat{\ell}$
\State Compute $L_{\tau_{1}}$ (the default $\tau_{1}$ for DRSCORE is $\tau_{1}=\sum_{i,j}A(i,j)/n$).
\State Compute $L_{\tau_{2}}$ (the default $\tau_{2}$ for DRSCORE is $\tau_{2}=\sum_{i,j}L_{\tau_{1}}(i,j)/n$).
\State Compute the inverse dual regularized Laplacian matrix
$\hat{W}=(I-e^{-\gamma}D_{\tau_{2}}^{-1}L_{\tau_{2}})^{-1}$ (the default $\gamma$ is 0.25). Calculate $\hat{M}=(\hat{W}+\hat{W}')/2$ and force $\hat{M}$'s diagonal entries to be 0.
\State Compute the $n\times \textbf{(K+2)}$ matrix $\check{X}$ such that $\check{X}=[\check{\eta}_{1},\check{\eta}_{2}, \ldots, \check{\eta}_{K+1}, \check{\eta}_{K+2}]\cdot \mathrm{diag}(\check{\lambda}_{1}, \check{\lambda}_{2}, \ldots, \check{\lambda}_{K+1}, \check{\lambda}_{K+2})$, where $\{\check{\lambda}_{i}\}_{i=1}^{K+2}$ are the leading eigenvalues of $\hat{M}$, $\{\check{\eta}_{i}\}_{i=1}^{K+2}$  are the respective eigenvectors with unit-norm.
\State  Compute $\check{X}^{*}$ by normalizing each of $\check{X}$'s rows to have unit length.
\State Treat each row of $\check{X}^{*}$ as a point in $\mathcal{R}^{K}$, and apply K-means to $\check{X}^{*}$ with $K$ clusters to obtain $\hat{\ell}$.
\end{algorithmic}
\end{algorithm}

There is a row-normalization step of DRSLIM to obtain $\check{X}^{*}$, and then apply K-means on $\check{X}^{*}$, instead of simply applying K-means on $\check{X}$ as in SLIM.  Though there are two regularizers $\tau_{1}$ and $\tau_{2}$ in DRSLIM, it is shown in Section \ref{secreal8} that DRSLIM is  insensitive to the choice of $\tau_{1}$ and $\tau_{2}$ as long as they are lager than 1. For convenience, unless specified, the default values for $\tau_{1}$ and $\tau_{2}$ for DRSLIM are set as in Algorithm \ref{alg:DRSLIM} in this paper. As argued in \cite{SLIM}, we always set the default value for $\gamma$ as 0.25 since it has been found to be a good choice in both simulated and real-world networks under both SBM and DCSBM. This choice of $\gamma$ is applied for all numerical studies in this paper.
%By analyzing the six steps of DRSLIM, we can find that our DRSLIM differs from SLIM in the following aspects a) $\hat{W}$ in DRSLIM is computed based on $L_{\tau_{2}}$ instead of the $A$ in SLIM. b) our DRSLIM always apply the production of the leading (K+2) eigenvectors and the leading (K+2) eigenvalues of $\hat{M}$ to construct $\check{X}$, instead of simply applying the leading $K$ eigenvectors as SLIM does. c) there is a row-normalization step of DRSLIM to obtain $\check{X}^{*}$, and then apply K-means on $\check{X}^{*}$, instead of simply applying K-means on $\check{X}$ as in SLIM. d) though there are two regularizers $\tau_{1}$ and $\tau_{2}$ in DRSLIM, it is shown in Section \ref{secreal8} that DRSLIM is  insensitive to the choice of $\tau_{1}$ and $\tau_{2}$ as long as they are lager than 1. For convenience, unless specified, the default values for $\tau_{1}$ and $\tau_{2}$ for DRSLIM are set as in Algorithm \ref{alg:DRSLIM} in this paper. As argued in \cite{SLIM}, we always set the default value for $\gamma$ as 0.25 since it has been found to be a good choice in both simulated and real-world networks under both SBM and DCSBM. This choice of $\gamma$ is applied for all numerical studies in this paper.

\section{Theoretical Results}\label{sec4}
This section builds theoretical frameworks  for DRSC, DRSCORE and DRSLIM to show that they yields stable consistent community detection under mild conditions.
Before propose the details of the theoretical analysis for the proposed methods, first we define the dual regularized population Laplacian matrix $\mathscr{L}_{\tau_{2}}$ and present some useful properties of it.

Define the diagonal matrix $\mathscr{D}_{1}$ to contain the expected node degrees such that
$\mathscr{D}_{1}(i,i)=\sum_{j}^{n}\Omega(i,j)$ for $1\leq i\leq n$, and define $\mathscr{D}_{\tau_{1}}$ such that
$\mathscr{D}_{\tau_{1}}=\mathscr{D}_{1}+\tau_{1}I$. Then the regularized population Laplacian $\mathscr{L}_{\tau_{1}}$ in $\mathcal{R}^{n\times n}$ can be presented as:
\begin{align*}
\mathscr{L}_{\tau_{1}}=\mathscr{D}_{\tau_{1}}^{-0.5}\Omega\mathscr{D}_{\tau_{1}}^{-0.5}.
\end{align*}
Then we define the diagonal matrix $\mathscr{D}_{2}$ such that its $i$-th diagonal element is defined as
$\mathscr{D}_{2}(i,i)=\sum_{j}^{n}\mathscr{L}_{\tau_{1}}(i,j)$, and define $\mathscr{D}_{\tau_{2}}$ such that
$\mathscr{D}_{\tau_{2}}=\mathscr{D}_{2}+\tau_{2}I$. Then the dual regularized population Laplacian $\mathscr{L}_{\tau_{2}}$ can be written in the following way:
\begin{align*}
\mathscr{L}_{\tau_{2}}=\mathscr{D}_{\tau_{2}}^{-0.5}\mathscr{L}_{\tau_{1}}\mathscr{D}_{\tau_{2}}^{-0.5}.
\end{align*}
\subsection{Analysis of $\mathscr{L}_{\tau_{2}}$}
For convenience, we denote  $\delta_{\mathrm{min}}=\mathrm{min}_{i}\mathscr{D}_{1}(i,i), \delta_{\mathrm{max}}=\mathrm{max}_{i}\mathscr{D}_{1}(i,i), \Delta_{\mathrm{min}}=\mathrm{min}_{i}D_{1}(i,i), \mathrm{and~} \Delta_{\mathrm{max}}=\mathrm{max}_{i}D_{1}(i,i)$.
And we set $\varpi_{a}, \varpi_{b}$ as
\begin{align*}
&\varpi_{a}=\mathrm{max}(\sqrt{\frac{\tau_{2}+\frac{\Delta_{\mathrm{max}}}{\tau_{1}+\Delta_{\mathrm{min}}}}{\tau_{2}+\frac{\delta_{\mathrm{min}}}{\tau_{1}+\delta_{\mathrm{max}}}}}-1, 1-\sqrt{\frac{\tau_{2}+\frac{\Delta_{\mathrm{min}}}{\tau_{1}+\Delta_{\mathrm{max}}}}{\tau_{2}+\frac{\delta_{\mathrm{max}}}{\tau_{1}+\delta_{\mathrm{min}}}}}),\\
&\varpi_{b}=\mathrm{max}(\frac{\tau_{2}+\frac{\Delta_{\mathrm{max}}}{\tau_{1}+\Delta_{\mathrm{min}}}}{\tau_{2}+\frac{\delta_{\mathrm{min}}}{\tau_{1}+\delta_{\mathrm{max}}}}-1, 1-\frac{\tau_{2}+\frac{\Delta_{\mathrm{min}}}{\tau_{1}+\Delta_{\mathrm{max}}}}{\tau_{2}+\frac{\delta_{\mathrm{max}}}{\tau_{1}+\delta_{\mathrm{min}}}}).
\end{align*}

%The following lemma gives the bounds of $\|L_{\tau_{1}}\|, \|\mathscr{L}_{\tau_{1}}\|, \|D_{\tau_{2}}\|, \|\mathscr{D}_{\tau_{2}}\|, \|L_{\tau_{2}}\|,  \|\mathscr{L}_{\tau_{2}}\|, \|D_{\tau_{2}}^{1/2}-\mathscr{D}_{\tau_{2}}^{-1/2}\|, \mathrm{and~} \|D_{\tau_{2}}-\mathscr{D}_{\tau_{2}}^{-1}\|$, which are powerful for theoretical analysis for DRSC, DRSCORE, and DRSLIM.
%\begin{lem}\label{boundTAU12}
%Under $DCSBM(n,P,\Theta,Z)$, we have
%\begin{align*}
%&\|L_{\tau_{1}}\|\leq \frac{\Delta_{\mathrm{max}}}{\tau_{1}+\Delta_{\mathrm{max}}}, ~~~ \|\mathscr{L}_{\tau_{1}}\|\leq \frac{\delta_{\mathrm{max}}}{\tau_{1}+\delta_{\mathrm{max}}} ,\\
%&\|D^{-1}_{\tau_{2}}\|\leq \frac{1}{\tau_{2}+\frac{\Delta_{\mathrm{min}}}{\tau_{1}+\Delta_{\mathrm{max}}}},~~~\|\mathscr{D}^{-1}_{\tau_{2}}\|\leq \frac{1}{\tau_{2}+\frac{\delta_{\mathrm{min}}}{\tau_{1}+\delta_{\mathrm{max}}}},\\
%&\|L_{\tau_{2}}\|\leq \frac{\Delta_{\mathrm{max}}}{\tau_{1}\tau_{2}+\tau_{2}\Delta_{\mathrm{max}}+\Delta_{\mathrm{min}}}, ~~~\|\mathscr{L}_{\tau_{2}}\|\leq \frac{\delta_{\mathrm{max}}}{\tau_{1}\tau_{2}+\tau_{2}\delta_{\mathrm{max}}+\delta_{\mathrm{min}}},\\
%&\|D_{\tau_{2}}^{-\frac{1}{2}}-\mathscr{D}_{\tau_{2}}^{-\frac{1}{2}}\|\leq\frac{1}{\sqrt{\tau_{2}+\frac{\Delta_{\mathrm{min}}}{\tau_{1}+\Delta_{\mathrm{max}}}}}\varpi_{a}, \mathrm{and~}\|D_{\tau_{2}}^{-1}-\mathscr{D}_{\tau_{2}}^{-1}\|\leq\frac{1}{\tau_{2}+\frac{\Delta_{\mathrm{min}}}{\tau_{1}+\Delta_{\mathrm{max}}}}\varpi_{b}.
%\end{align*}
%\end{lem}
Define a $K\times n$ matrix $Q_{\tau_{1}}$ as $Q_{\tau_{1}}=PZ'\Theta$ (note that though $Q_{\tau_{1}}$ is not defined based on $\tau_{1}$, we use this subscript to mark that it is not designed based on dual procedures to distinguish it from $Q_{\tau_{2}}$, the same nomenclature holds for $D_{P}^{\tau_{1}}, \tilde{P}_{\tau_{1}}$ defined below), and define the
$K\times K$ diagonal matrix $D^{\tau_{1}}_{P}$ as $D^{\tau_{1}}_{P}(i,i)=\sum_{j=1}^{n}Q_{\tau_{1}}(i,j), i=1,2,\ldots,K$.
The next lemma gives an explicit form for $\mathscr{L}_{\tau_{1}}$ as a product of the parameter matrices.
\begin{lem}\label{ExplicitformforLtau1}
(\emph{Explicit form for} $\mathscr{L}_{\tau_{1}}$) Under $DCSBM(n, P, \Theta, Z)$,
 define $\theta_{\tau_{1}}(i)$ as $\theta_{\tau_{1}}(i)=\theta_{i}\frac{\mathscr{D}_{1}(i,i)}{\mathscr{D}_{1}(i,i)+\tau_{1}}$,
let $\Theta_{\tau_{1}}$ be a diagonal matrix whose
$ii$'th entry is $\theta_{\tau_{1}}(i)$. Define $\tilde{P}_{\tau_{1}}$ as $\tilde{P}_{\tau_{1}}=(D^{\tau_{1}}_{P})^{-0.5}P(D^{\tau_{1}}_{P})^{-0.5}$, then $\mathscr{L}_{\tau_{1}}$ can be written as
\begin{align*}
  \mathscr{L}_{\tau_{1}}=\mathscr{D}_{\tau_{1}}^{-0.5}\Omega \mathscr{D}_{\tau_{1}}^{-0.5}=\Theta_{\tau_{1}}^{0.5}Z\tilde{P}_{\tau_{1}}Z'\Theta_{\tau_{1}}^{0.5}.
\end{align*}
\end{lem}

In fact, we can rewrite $\mathscr{L}_{\tau_{1}}$ as follows:
\begin{align*}
\mathscr{L}_{\tau_{1}}=\|\tilde{\theta}_{\tau_{1}}\|^{2}\tilde{\Gamma}_{\tau_{1}}\tilde{D}_{\tau_{1}}\tilde{P}_{\tau_{1}}\tilde{D}_{\tau_{1}}\tilde{\Gamma}_{\tau_{1}}',
\end{align*}
where $\tilde{\theta}_{\tau_{1}}$ is an $n\times 1$ vector with $\tilde{\theta}_{\tau_{1}}(i)=\sqrt{\theta_{\tau_{1}}(i)}, $ for $1\leq i\leq n,$ $\tilde{D}_{\tau_{1}}$ is a $K\times K$ diagonal matrix of the \emph{overall degree intensities} with $\tilde{D}_{\tau_{1}}(k,k)=\|\tilde{\theta}_{\tau_{1}}^{(k)}\|(\|\tilde{\theta}_{\tau_{1}}\|)^{-1}, 1\leq k\leq K,$ $\tilde{\theta}_{\tau_{1}}^{(k)}$ is an $n\times 1$  vector such that for $1\leq i\leq n, 1\leq k\leq K$, $\tilde{\theta}_{\tau_{1}}^{(k)}(i) = \tilde{\theta}_{\tau_{1}}(i)\mathds{1}_{\{g_{i}=k\}} $, and the $n\times K$ matrix $\tilde{\Gamma}_{\tau_{1}}$ such that
\begin{align*}
\tilde{\Gamma}_{\tau_{1}}=[\frac{\tilde{\theta}_{\tau_{1}}^{(1)}}{\|\tilde{\theta}_{\tau_{1}}^{(1)}\|} ~ \frac{\tilde{\theta}_{\tau_{1}}^{(2)}}{\|\tilde{\theta}_{\tau_{1}}^{(2)}\|} ~\ldots~\frac{\tilde{\theta}_{\tau_{1}}^{(K)}}{\|\tilde{\theta}_{\tau_{1}}^{(K)}\|}].
\end{align*}
%Then define the $n\times 1$ vector $\tilde{\theta}_{\tau_{1}}$ as
%\begin{align*}
%  \tilde{\theta}_{\tau_{1}}(i)=\sqrt{\theta_{\tau_{1}}(i)},\qquad 1\leq i\leq n.
%\end{align*}
%Let $\tilde{\theta}_{\tau_{1}}^{(k)}$ be the $n\times 1$  vectors such that for $1\leq i\leq n, 1\leq k\leq K$,
%
%Let $\tilde{D}_{\tau_{1}}$ be the $K\times K$ diagonal matrix of the \emph{overall degree intensities}
%\begin{align*}
%  \tilde{D}_{\tau_{1}}(k,k)=\frac{\|\tilde{\theta}_{\tau_{1}}^{(k)}\|}{\|\tilde{\theta}_{\tau_{1}}\|},\qquad 1\leq k\leq K.
%\end{align*}
%Now we define the $n\times K$ matrix $\tilde{\Gamma}_{\tau_{1}}$ such that
%\begin{align*}
%  \tilde{\Gamma}_{\tau_{1}}=[\frac{\tilde{\theta}_{\tau_{1}}^{(1)}}{\|\tilde{\theta}_{\tau_{1}}^{(1)}\|} ~ \frac{\tilde{\theta}_{\tau_{1}}^{(2)}}{\|\tilde{\theta}_{\tau_{1}}^{(2)}\|} ~\ldots~\frac{\tilde{\theta}_{\tau_{1}}^{(K)}}{\|\tilde{\theta}_{\tau_{1}}^{(K)}\|}]
%\end{align*}

%Then according to the special form of $\tilde{\Gamma}_{\tau_{1}}$, we have $\tilde{\Gamma}_{\tau_{1}}'\tilde{\Gamma}_{\tau_{1}}=I$, where $I$ is a $K\times K$ identity matrix. %Now we can rewrite $\mathscr{L}_{\tau_{1}}$ as
%\begin{align*}
%\mathscr{L}_{\tau_{1}}=\|\tilde{\theta}_{\tau_{1}}\|^{2}\tilde{\Gamma}_{\tau_{1}}\tilde{D}_{\tau_{1}}\tilde{P}_{\tau_{1}}\tilde{D}_{\tau_{1}}\tilde{\Gamma}_{\tau_{1}}'.
%\end{align*}

To conduct the explicit form of $\mathscr{L}_{\tau_{2}}$, we define a $K\times n$ matrix $Q_{\tau_{2}}$ as $Q_{\tau_{2}}=\tilde{P}_{\tau_{1}}Z'\Theta_{\tau_{1}}^{0.5}$, and define a
$K\times K$ diagonal matrix $D^{\tau_{2}}_{P}$ as $D^{\tau_{2}}_{P}(i,i)=\sum_{j=1}^{n}Q_{\tau_{2}}(i,j), i=1,2,\ldots,K$.
The next lemma gives an explicit form for $\mathscr{L}_{\tau_{2}}$ as a product of the parameter matrices.
\begin{lem}\label{ExplicitformforLtau2}
(\emph{Explicit form for} $\mathscr{L}_{\tau_{2}}$) Under $DCSBM(n, P, \Theta, Z)$,
 define $\theta_{\tau_{2}}(i)$ as $\theta_{\tau_{2}}(i)=(\theta_{\tau_{1}}(i))^{0.5}\frac{\mathscr{D}_{2}(i,i)}{\mathscr{D}_{2}(i,i)+\tau_{2}}$,
let $\Theta_{\tau_{2}}$ be a diagonal matrix whose
$ii$'th entry is $\theta_{\tau_{2}}(i)$. Define $\tilde{P}_{\tau_{2}}$ as $\tilde{P}_{\tau_{2}}=(D^{\tau_{2}}_{P})^{-0.5}\tilde{P}_{\tau_{1}}(D^{\tau_{2}}_{P})^{-0.5}$, then $\mathscr{L}_{\tau_{2}}$ can be written as
\begin{align*}
  \mathscr{L}_{\tau_{2}}=\mathscr{D}_{\tau_{2}}^{-0.5}\mathscr{L}_{\tau_{1}} \mathscr{D}_{\tau_{2}}^{-0.5}=\Theta_{\tau_{2}}^{0.5}Z\tilde{P}_{\tau_{2}}Z'\Theta_{\tau_{2}}^{0.5}.
\end{align*}
\end{lem}

Similarly, we can represent $\mathscr{L}_{\tau_{2}}$ to the following form:
\begin{align*}
\mathscr{L}_{\tau_{2}}=\|\tilde{\theta}_{\tau_{2}}\|^{2}\tilde{\Gamma}_{\tau_{2}}\tilde{D}_{\tau_{2}}\tilde{P}_{\tau_{2}}\tilde{D}_{\tau_{2}}\tilde{\Gamma}_{\tau_{2}}',
\end{align*}
where $\tilde{\theta}_{\tau_{2}} = (\tilde{\theta}_{\tau_{2}}(1), \cdots, \tilde{\theta}_{\tau_{2}}(n))^{\prime}$, $\tilde{\theta}_{\tau_{2}}(i)=\sqrt{\theta_{\tau_{2}}(i)}, 1\leq i\leq n,$ $\tilde{D}_{\tau_{2}}$ is a $K\times K$ diagonal matrix with $\tilde{D}_{\tau_{2}}(k,k)=\|\tilde{\theta}_{\tau_{2}}^{(k)}\|(\|\tilde{\theta}_{\tau_{2}}\|)^{-1}, 1\leq k\leq K$, $\tilde{\theta}_{\tau_{2}}^{(k)}$ is an $n\times 1$  vector such that for $1\leq i\leq n, 1\leq k\leq K$, $\tilde{\theta}_{\tau_{2}}^{(k)}(i)= \tilde{\theta}_{\tau_{2}}(i)\mathds{1}_{\{g_{i}=k\}}$, and $
\tilde{\Gamma}_{\tau_{2}}=[\frac{\tilde{\theta}_{\tau_{2}}^{(1)}}{\|\tilde{\theta}_{\tau_{2}}^{(1)}\|} ~ \frac{\tilde{\theta}_{\tau_{2}}^{(2)}}{\|\tilde{\theta}_{\tau_{2}}^{(2)}\|} ~\ldots~\frac{\tilde{\theta}_{\tau_{2}}^{(K)}}{\|\tilde{\theta}_{\tau_{2}}^{(K)}\|}].$

%Then define the $n\times 1$ vector $\tilde{\theta}_{\tau_{2}}$ as
%\begin{align*}
%  \tilde{\theta}_{\tau_{2}}(i)=\sqrt{\theta_{\tau_{2}}(i)},\qquad 1\leq i\leq n.
%\end{align*}
%Let $\tilde{\theta}_{\tau_{2}}^{(k)}$ be the $n\times 1$  vectors such that for $1\leq i\leq n, 1\leq k\leq K$,
%\begin{equation*}
%  \tilde{\theta}_{\tau_{2}}^{(k)}(i)=\begin{cases}
%                    \tilde{\theta}_{\tau_{2}}(i), & \mbox{if } g_{i}=k, \\
%                    0, & \mbox{otherwise}.
%                  \end{cases}
%\end{equation*}
%Let $\tilde{D}_{\tau_{2}}$ be the $K\times K$ diagonal matrix of the \emph{overall degree intensities}
%\begin{align*}
%  \tilde{D}_{\tau_{2}}(k,k)=\frac{\|\tilde{\theta}_{\tau_{2}}^{(k)}\|}{\|\tilde{\theta}_{\tau_{2}}\|},\qquad 1\leq k\leq K.
%\end{align*}
%Now we define the $n\times K$ matrix $\tilde{\Gamma}_{\tau_{2}}$ such that
%\begin{align*}
%  \tilde{\Gamma}_{\tau_{2}}=[\frac{\tilde{\theta}_{\tau_{2}}^{(1)}}{\|\tilde{\theta}_{\tau_{2}}^{(1)}\|} ~ \frac{\tilde{\theta}_{\tau_{2}}^{(2)}}{\|\tilde{\theta}_{\tau_{2}}^{(2)}\|} ~\ldots~\frac{\tilde{\theta}_{\tau_{2}}^{(K)}}{\|\tilde{\theta}_{\tau_{2}}^{(K)}\|}].
%\end{align*}
%Then according to the special form of $\tilde{\Gamma}_{\tau_{2}}$, we have $\tilde{\Gamma}_{\tau_{2}}'\tilde{\Gamma}_{\tau_{2}}=I$.% Now we can rewrite $\mathscr{L}_{\tau_{2}}$ as
%\begin{align*}
%\mathscr{L}_{\tau_{2}}=\|\tilde{\theta}_{\tau_{2}}\|^{2}\tilde{\Gamma}_{\tau_{2}}\tilde{D}_{\tau_{2}}\tilde{P}_{\tau_{2}}\tilde{D}_{\tau_{2}}\tilde{\Gamma}_{\tau_{2}}'.
%\end{align*}

By basic knowledge of algebra, we know that the rank of $\mathscr{L}_{\tau_{2}}$ is $K$ when there are $K$ clusters, therefore $\mathscr{L}_{\tau_{2}}$ has $K$ nonzero eigenvalues. We give the expressions of the leading $K$ eigenvectors of $\mathscr{L}_{\tau_{2}}$ in Lemma \ref{eigenvectors}.
%By basic knowledge of algebra, we know that the rank of $\mathscr{L}_{\tau_{2}}$ is $K$ when there are $K$ clusters, therefore $\mathscr{L}_{\tau_{2}}$ has $K$ nonzero eigenvalues.
%Meanwhile, based on the observation that $\mathscr{L}_{\tau_{1}}=\|\tilde{\theta}_{\tau_{1}}\|^{2}\tilde{\Gamma}_{\tau_{1}}\tilde{D}_{\tau_{1}}\tilde{P}_{\tau_{1}}\tilde{D}_{\tau_{1}}\tilde{\Gamma}_{\tau_{1}}'$ and $\mathscr{L}_{\tau_{2}}=\|\tilde{\theta}_{\tau_{2}}\|^{2}\tilde{\Gamma}_{\tau_{2}}\tilde{D}_{\tau_{2}}\tilde{P}_{\tau_{2}}\tilde{D}_{\tau_{2}}\tilde{\Gamma}_{\tau_{2}}'$,
%we argue that $\mathscr{L}_{\tau_{2}}$ may share similar properties as $\mathscr{L}_{\tau_{1}}$, and this statement is supported by the next Lemma which gives the expressions of the leading $K$ eigenvectors of $\mathscr{L}_{\tau_{2}}$ \footnote{Though \cite{RSC} does not give the expressions of the leading $K$ eigenvectors of $\mathscr{L}_{\tau_{1}}$, we can find such expressions simply by setting $\tau_{2}$ as $\tau_{1}$ in Lemma \ref{eigenvectors}, this is the reason that we argue that $\mathscr{L}_{\tau_{2}}$ may share similar properties as $\mathscr{L}_{\tau_{1}}$ based on the observation that they have similar expression forms.}.
\begin{lem}\label{eigenvectors}
Under $DCSBM(n, P, \Theta, Z)$, suppose all eigenvalues of $\tilde{D}_{\tau_{2}}\tilde{P}_{\tau_{2}}\tilde{D}_{\tau_{2}}$ are simple. Let $\lambda_{1}/\|\tilde{\theta}_{\tau_{2}}\|^{2}, \lambda_{2}/\|\tilde{\theta}_{\tau_{2}}\|^{2},\ldots, \lambda_{K}/\|\tilde{\theta}_{\tau_{2}}\|^{2}$ be such eigenvalues, arranged in the descending order of the magnitudes, and let $a_{1}, a_{2}, \ldots, a_{K}$ be the associated (unit-norm) eigenvectors. Then the $K$ nonzero eigenvalues of $\mathscr{L}_{\tau_{2}}$ are $\lambda_{1}, \lambda_{2}, \ldots, \lambda_{K}$, with the associated (unit-norm) eigenvectors being
\begin{align*}
  \eta_{k}=\sum_{i=1}^{K}[a_{k}(i)/\|\tilde{\theta}_{\tau_{2}}^{(i)}\|]\cdot \tilde{\theta}_{\tau_{2}}^{(i)}, k=1,2,\ldots, K.
\end{align*}
\end{lem}
Let $\hat{\lambda}_{1},\hat{\lambda}_{2},\ldots, \hat{\lambda}_{K}$ be the $K$ leading eigenvalues of $L_{\tau_{2}}$.  The theoretical bound for the spectral norm $\|L_{\tau_{2}}-\mathscr{L}_{\tau_{2}}\|$  is given in Lemma \ref{bounddualL} which is helpful in the theoretical analysis for DRSC and DRSLIM.

For convenience, we set one parameter $err_{n}$ as:
\begin{align*}
err_{n}=\frac{4\sqrt{\frac{3\mathrm{log}(4n/\epsilon)}{\delta_{\mathrm{min}}+\tau_{1}}}}{\tau_{2}+\frac{\delta_{\mathrm{min}}}{\delta_{\mathrm{max}}+\tau_{1}}}+(\frac{1}{\sqrt{\tau_{2}+\frac{\Delta_{\mathrm{min}}}{\tau_{1}+\Delta_{\mathrm{max}}}}}+\frac{1}{\sqrt{\tau_{2}+\frac{\delta_{\mathrm{min}}}{\tau_{1}+\delta_{\mathrm{max}}}}})\frac{\frac{\Delta_{\mathrm{max}}}{\tau_{1}+\Delta_{\mathrm{max}}}}{\sqrt{\tau_{2}+\frac{\Delta_{\mathrm{min}}}{\Delta_{\mathrm{max}}+\tau_{1}}}}\varpi_{a}.
\end{align*}
\begin{lem}\label{bounddualL}(Concentration of the dual regularized Graph Laplacian) Under $DCSBM(n, P, \Theta, Z)$, if $\tau_{1}+\delta_{\mathrm{min}}>3\mathrm{log}(4n/\epsilon)$, with probability at least $1-\epsilon$, we have
\begin{align*}
&\|L_{\tau_{2}}-\mathscr{L}_{\tau_{2}}\|\leq err_{n}.
\end{align*}
\end{lem}
\begin{figure}%[H]
	\centering
\includegraphics[width=11cm,height=7cm]{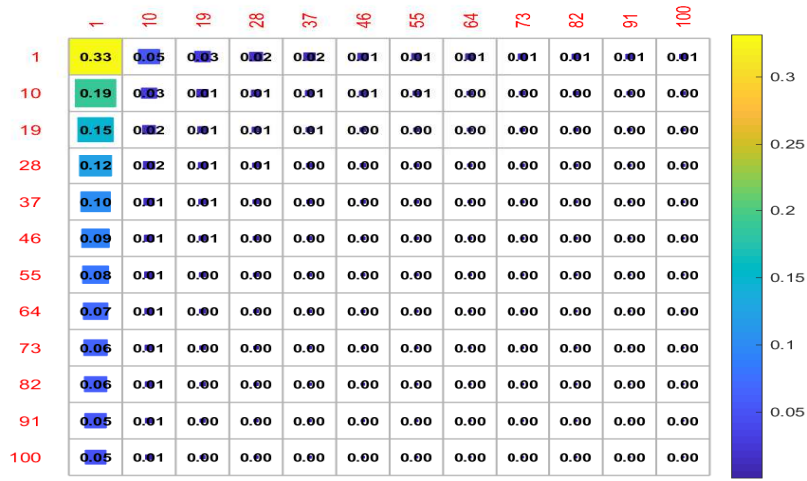}
\caption{The effect of $\tau_{1}$ and $\tau_{2}$ on  $\|L_{\tau_{2}}-\mathscr{L}_{\tau_{2}}\|$. x-axis: $\tau_{1}$. y-axis: $\tau_{2}$. For the figure, we take $n=400, K=2$, and let nodes belong to one of the clusters with equal probability, set $\theta_{i}=0.3$ for $g_{i}=1$, $\theta_{i}=0.7$ for $g_{i}=2$, and the mixing matrix $P$ with 0.1 as diagonal entries and 0.05 as off-diagonal entries. Given the above settings of parameters, we can generate one $A$ and $\Omega$ with equal size, then set $\tau_{1},\tau_{2}\in\{1, 10, 19, \ldots, 100\}$, for each $\tau_{1}$ and $\tau_{2}$, we can obtain $L_{\tau_{2}}$ and $\mathscr{L}_{\tau_{2}}$, then we can obtain the heatmap of $\|L_{\tau_{2}}-\mathscr{L}_{\tau_{2}}\|$ against $\tau_{1}$ and $\tau_{2}$.
}
\label{HeatmapdisL}
\end{figure}

\begin{figure}%[H]
	\centering
\includegraphics[width=8cm,height=6cm]{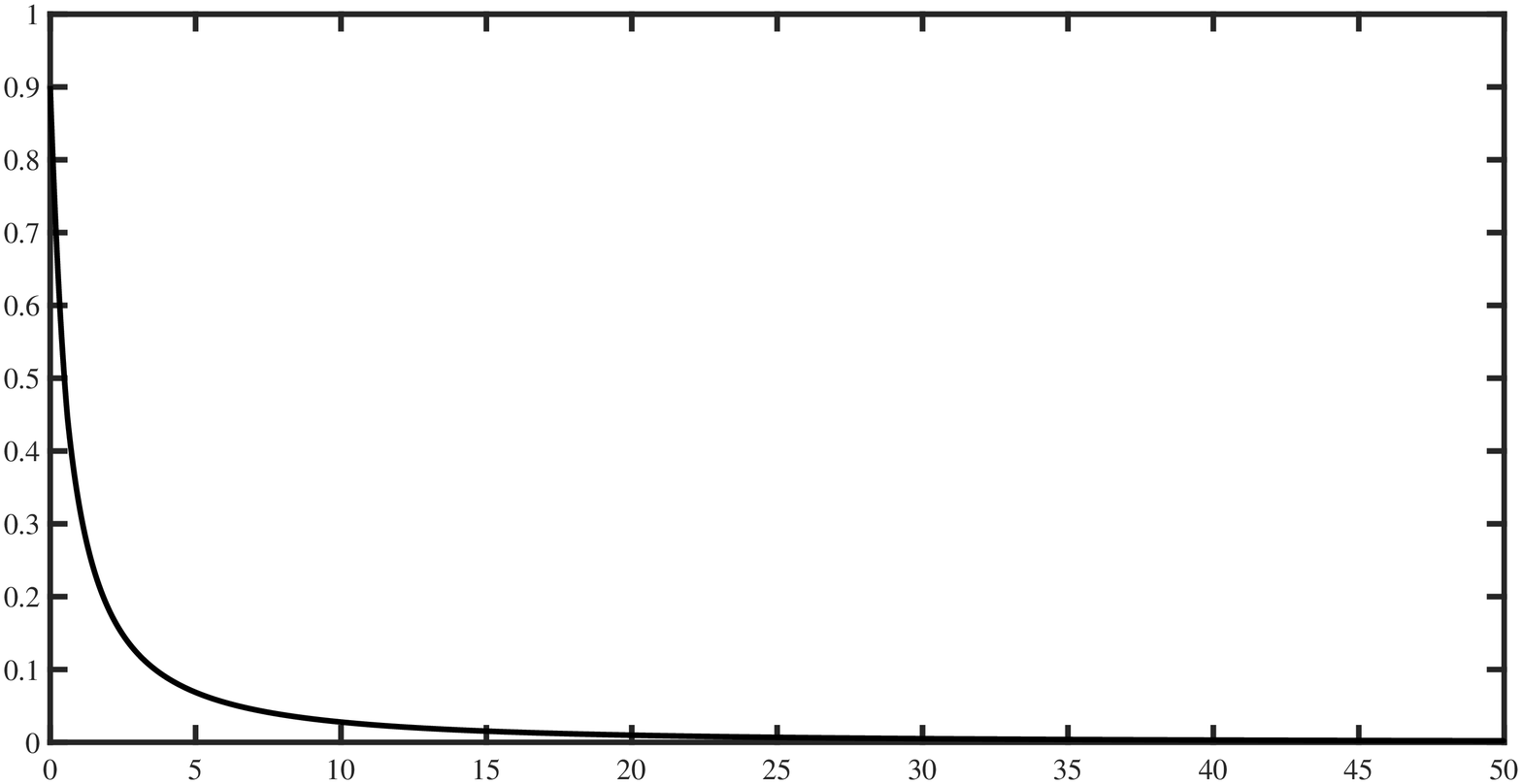}
\caption{The effect of $\tau$ on  $\|L_{\tau_{2}}-\mathscr{L}_{\tau_{2}}\|$ with $\tau_{1}= \tau_{2}=\tau$. $\tau\in\{0,0.01,0.02,\ldots,50\}$ and other settings are same as in Figure \ref{HeatmapdisL}. x-axis: $\tau$. y-axis: $\|L_{\tau_{2}}-\mathscr{L}_{\tau_{2}}\|$. }
\label{disL}
\end{figure}
Figure \ref{HeatmapdisL} demonstrates the effect of $\tau_{1}$ and $\tau_{2}$ on  $\|L_{\tau_{2}}-\mathscr{L}_{\tau_{2}}\|$, and the color indicates $\|L_{\tau_{2}}-\mathscr{L}_{\tau_{2}}\|$ on given $\tau_{1}$ and $\tau_{2}$. We also consider the case when $\tau_{1}=\tau_{2}$ in Figure \ref{disL}.
From Figure \ref{HeatmapdisL} and Figure \ref{disL}, we can find that $\|L_{\tau_{2}}-\mathscr{L}_{\tau_{2}}\|$ decreases to zero when $\tau_{1}$ and/or $\tau_{2}$ increase(s). This is consistent with Lemma \ref{bounddualL}, since when increasing the two regularizers or one of the two regularizers, $err_{n}$ decreases to zero, and hence $\|L_{\tau_{2}}-\mathscr{L}_{\tau_{2}}\|$ goes to zero.  Meanwhile, $\|L_{\tau_{2}}-\mathscr{L}_{\tau_{2}}\|$ is more sensitive on $\tau_{1}$ than on $\tau_{2}$, since smaller $\tau_{1}$ indicates large $\|L_{\tau_{2}}-\mathscr{L}_{\tau_{2}}\|$  while $\|L_{\tau_{2}}-\mathscr{L}_{\tau_{2}}\|$ is quite small for any $\tau_{2}$ as long as $\tau_{1}$ is larger than 0. Therefore, to make $\|L_{\tau_{2}}-\mathscr{L}_{\tau_{2}}\|$ small enough, $\tau_{1}$ should be larger than 0. Numerical results in Section \ref{sec4} also supports this statement.
%%%%%%%%%%%%%%%%%%%%%%%%%%%%%%%%%%%%

We can also bound the differences of the estimated eigenvalues and the population eigenvalues by $err_n$. Since $L_{\tau_{2}}$ and $\mathscr{L}_{\tau_{2}}$ are two symmetric matrices, Weyl's inequality \citep{weyl1912das} holds. The following lemma is a direct result of Lemma \ref{bounddualL} and the Weyl's inequality, the proof of which is omitted.
\begin{lem}\label{boundeigenvalue}
  Under $DCSBM(n, P, \Theta, Z)$, if $\tau_{1}+\delta_{\mathrm{min}}>3\mathrm{log}(4n/\epsilon)$, then with probability at least $1-\epsilon$,
\begin{align*}
\underset{1\leq k\leq K}{\mathrm{max}}\{|\hat{\lambda}_{k}-\lambda_{k}|\}\leq err_{n}.
\end{align*}
\end{lem}

The following lemma gives a bound for difference of matrices of eigenvectors and also constitutes the key component of the proof of Theorem \ref{boundstar} for DRSC.
\begin{lem}\label{boundeigenvector}
Let $\hat{V}_{K}$ be the $n\times K$ matrix such that its $i$-th column is $\hat{\eta}_{i}$, let $V_{K}$ be the $n\times K$ matrix such that its $i$-th column is $\eta_{i}$ for $1\leq i\leq K$, then under $DCSBM(n, P, \Theta, Z)$, if we assume that (a) $\tau_{1}+\delta_{\mathrm{min}}>3\mathrm{log}(4n/\epsilon)$, (b) $err_{n}\leq \frac{\lambda_{K}}{2}$, with probability at least $1-\epsilon$, we have
\begin{align*}
\|\hat{V}_{K}-V_{K}\|_{F}\leq \frac{8err_{n}\sqrt{K}}{\lambda_{K}}.
\end{align*}
\end{lem}
Note that assumption (b) also means that we assume all nonzero eigenvalues of $\mathscr{L}_{\tau_{2}}$ are positive.
\subsection{Main Results for DRSC}
\subsubsection{Population analysis of DRSC}
Recall that we have obtained the expressions of the leading $(K+1)$ eigenvalues and the leading $K$ eigenvectors of $\mathscr{L}_{\tau_{2}}$, then we can write down the Ideal DRSC algorithm which can be obtained via using $\Omega$ to replace $A$ in the DRSC algorithm. %The Ideal DRSC algorithm proceeds as follows:

 \textbf{Ideal DRSC}. Input: $\Omega$. Output: $\ell$.

 \textbf{Step 1:} Obtain $\mathscr{L}_{\tau_{1}}$.

 \textbf{Step 2:} Obtain $\mathscr{L}_{\tau_{2}}$.

 \textbf{Step 3:} Obtain $X=[\eta_{1},\eta_{2}, \ldots, \eta_{K}, \eta_{K+1}]\cdot \mathrm{diag}(\lambda_{1}, \lambda_{2}, \ldots, \lambda_{K}, \lambda_{K+1})=[\lambda_{1}\eta_{1},\lambda_{2}\eta_{2}, \ldots, \lambda_{K}\eta_{K}, 0]$ since $\mathscr{L}_{\tau_{2}}$ only has $K$ nonzero eigenvalues.

 \textbf{Step 4:} Obtain $X^{*}$, the row-normalized version of $X$.

 \textbf{Step 5:} Apply K-means to $X^{*}$ assuming there are $K$ clusters.

The population analysis of DRSC aims at confirming that the Ideal DRSC algorithm can return perfect clustering. Lemma \ref{population} guarantees that the Ideal DRSC algorithm surely provides true nodes labels.
\begin{lem}\label{population}
Under $DCSBM(n, P, \Theta, Z)$, $X$ has $K$ distinct rows and for any two distinct nodes $i,j$ if $g_{i}=g_{j}$, the $j$-th row of $X$ equals the $i$-th row of it.
\end{lem}
Applying K-means to $X$ leads to the true community labels of
each node. The above population analysis for DRSC method presents a direct understanding of
why the DRSC algorithm works and guarantees that DRSC returns perfect clustering results under the ideal case.
%\subsubsection{Characterization of the matrix $X^{*}$}
%For the full theoretical work for DRSC, we need to bound the Hamming error rate (defined in (\ref{DefinHamm})) of DRSC. The theoretical bound of Hamming error rate is obtained based on the bound of $\|\hat{X}^{*}-X^{*}\|_{F}$. Therefore,
%in this sub-section, we aim at bounding $\|\hat{X}^{*}-X^{*}\|_{F}$, which is obtained by the bound of $\|\hat{X}-X\|_{F}$ by basic algebra. The objective of bounding $\|\hat{X}-X\|_{F}$ can be achieved by Lemma  \ref{bounddualL} which gives the theoretical bound of the difference between eigenvalues of $L_{\tau_{2}}$ and $\mathscr{L}_{\tau_{2}}$.
%
%%Theorem \ref{boundstar} provides the bound of $\|\hat{X}^{*}-X^{*}\|_{F}$, which is the corner stone to characterize the behavior of our DRSC approach.
%\begin{thm}\label{boundstar}
%Under $DCSBM(n, P, \Theta, Z)$, define $m_{a}=\mathrm{min}_{i}\{\mathrm{min}\{\|\hat{X}_{i}\|, \|X_{i}\|\}\}$ as the length of the shortest row in $\hat{X}$ and $X$. Then, for any $\epsilon >0$ and sufficiently large $n$,  assume that assumptions (a) and (b) in Lemma \ref{boundeigenvector} hold,
%then with probability at least $1-\epsilon$, the following holds
%\begin{align*}
%&\|\hat{X}-X\|_{F}\leq \sqrt{Kerr^{2}_{n}+\hat{\lambda}^{2}_{K+1}}+\frac{8Kerr_{n}}{\lambda_{K}},\\
%&\|\hat{X}^{*}-X^{*}\|_{F}\leq \frac{1}{m_{a}}(\sqrt{Kerr^{2}_{n}+\hat{\lambda}^{2}_{K+1}}+\frac{8Kerr_{n}}{\lambda_{K}}).
%\end{align*}
%\end{thm}
\subsubsection{Bound of Hamming error rate of DRSC}
Hamming error rate \cite{SCORE} is a proper criterion to measure the performance of $\hat{\ell}$. This section bounds the Hamming error rate of DRSC under DCSBM to show that it stably yields consistent community detection under certain conditions.

The Hamming error rate of $\hat{\ell}$ is defined as:
 \begin{align*}
   \mathrm{Hamm}_{n}(\hat{\ell},\ell)=\underset{\pi\in S_{K}}{\mathrm{min~}}H_{p}(\hat{\ell},\pi(\ell))/n,
 \end{align*}
 where $S_{K}=\{\pi:\pi \mathrm{~is~a~permutation~of~the~set~}\{1,2,\ldots, K\}\}$ \footnote{Due to the fact that the clustering errors should not
depend on how we tag each of the K communities, it is necessary for us to take permutation of labels into account to measure the performances of DRSC.}, $\pi(\ell)(i)=\pi(\ell(i))$ for $1\leq i\leq n$,  and $H_{p}(\hat{\ell},\ell)$ is the expected number of mismatched labels which is defined as
 \begin{align*}
   H_{p}(\hat{\ell},\ell)=\sum_{i=1}^{n}P(\hat{\ell}(i)\neq \ell(i)).
 \end{align*}
Therefore, a direct understanding of Hamming error rate is that it is the ratio between the expected number of nodes where the estimated label does not match with the true label and the number of nodes in a given network \cite{SCORE}.

 The theoretical bound of Hamming error rate is obtained based on the bound of $\|\hat{X}^{*}-X^{*}\|_{F}$. Therefore, we first bound $\|\hat{X}-X\|_{F}$ and $\|\hat{X}^{*}-X^{*}\|_{F}$ in Theorem \ref{boundstar}.
\begin{thm}\label{boundstar}
	Under $DCSBM(n, P, \Theta, Z)$, define $m_{a}=\mathrm{min}_{i}\{\mathrm{min}\{\|\hat{X}_{i}\|, \|X_{i}\|\}\}$ as the length of the shortest row in $\hat{X}$ and $X$. Then, for any $\epsilon >0$ and sufficiently large $n$,  assume that assumptions (a) and (b) in Lemma \ref{boundeigenvector} hold,
	then with probability at least $1-\epsilon$, the following holds
	\begin{align*}
	&\|\hat{X}-X\|_{F}\leq \sqrt{Kerr^{2}_{n}+\hat{\lambda}^{2}_{K+1}}+\frac{8Kerr_{n}}{\lambda_{K}},\\
	&\|\hat{X}^{*}-X^{*}\|_{F}\leq \frac{1}{m_{a}}(\sqrt{Kerr^{2}_{n}+\hat{\lambda}^{2}_{K+1}}+\frac{8Kerr_{n}}{\lambda_{K}}).
	\end{align*}
\end{thm}

 The following theorem is the main theoretical result of this paper which bounds the Hamming error rate of our DRSC method under mild conditions and shows that the DRSC method stably yields consistent community detection under several  conditions if we assume that the adjacency matrix $A$ are generated from the DCSBM model.
\begin{thm}\label{mainDRSC}
Under the DCSBM with parameters $\{n, P, \Theta, Z\}$ and the same assumptions as in Theorem \ref{boundstar} hold, suppose as $n\rightarrow \infty$, we have
\begin{align*}
 \frac{4}{m_{a}^{2}}(\sqrt{Kerr^{2}_{n}+\hat{\lambda}^{2}_{K+1}}+\frac{8Kerr_{n}}{\lambda_{K}})^{2}
/\mathrm{min~}\{n_{1}, n_{2}, \ldots, n_{K}\}\rightarrow 0,
\end{align*}
where $n_{k}$ is the size of the $k$-th community for $1\leq k\leq K$.
For the estimates label vector $\hat{\ell}$ by DRSC, such that with probability at least $1-\epsilon$, we have
\begin{align*}
  \mathrm{Hamm}_{n}(\hat{\ell},\ell) \leq\frac{4}{nm_{a}^{2}}(\sqrt{Kerr^{2}_{n}+\hat{\lambda}^{2}_{K+1}}+\frac{8Kerr_{n}}{\lambda_{K}})^{2}.
\end{align*}
\end{thm}
Note that by the assumption (b), we can room $\mathrm{Hamm}_{n}(\hat{\ell},\ell)$ as
\begin{align*}
\mathrm{Hamm}_{n}(\hat{\ell},\ell)\leq\frac{(\sqrt{K\lambda_{K}^{2}+4\hat{\lambda}^{2}_{K+1}}+8K)^{2}}{nm_{a}^{2}}.
\end{align*}
We can find that as $n$ goes on increasing while keeping other parameters fixed, the Hamming error rates of DRSC decreases to zero. A larger $K$ suggests a larger error bound, which means that it becomes harder to detect communities for DRSC when $K$ increases. We can also find that our DRSC procedure can detect both strong signal networks and weak signal networks since the theoretical bound of the Hamming error rate for DRSC depends on $\hat{\lambda}_{K+1}$ and $\lambda_{K}$. When a network is strong signal, suggesting that $\hat{\lambda}_{K+1}$ is much smaller than $\hat{\lambda}_{K}$, and hence much smaller than $\lambda_{K}$ by Lemma \ref{boundeigenvalue}, therefore the bound of Hamming error rate for strong signal networks mainly depends on the $K$-th leading eigenvalue $\hat{\lambda}_{K}$.
When dealing with weak signal networks, $\hat{\lambda}_{K+1}$ is quite close to $\hat{\lambda}_{K}$, and hence it is also close to $\lambda_{K}$ by Lemma \ref{boundeigenvalue}, which means that the bound of Hamming error rate for weak signal networks mainly depends on the $(K+1)$-th leading eigenvalue $\hat{\lambda}_{K+1}$. %The above analysis provides an idea on why our DRSC method  can detect communities for both strong signal and weak signal networks.
\subsection{Main Results for DRSCORE}
\subsubsection{Population analysis of DRSCORE}
Similar as the procedures of theoretical analysis for DRSC, for the population analysis of DRSCORE, first we present its  ideal case, the Ideal DRSCORE algorithm:

 \textbf{Ideal DRSCORE}. Input: $\Omega$. Output: $\ell$.

 \textbf{Step 1:} Obtain $\mathscr{L}_{\tau_{1}}$.

 \textbf{Step 2:} Obtain $\mathscr{L}_{\tau_{2}}$.

 \textbf{Step 3:} Obtain $X=[\eta_{1},\eta_{2}, \ldots, \eta_{K}, \eta_{K+1}]\cdot \mathrm{diag}(\lambda_{1}, \lambda_{2}, \ldots, \lambda_{K}, \lambda_{K+1})=[\lambda_{1}\eta_{1},\lambda_{2}\eta_{2}, \ldots, \lambda_{K}\eta_{K}, 0]$ since $\mathscr{L}_{\tau_{2}}$ only has $K$ nonzero eigenvalues. Set $X_{i}$ as the $i$-th column of $X, 1\leq i\leq K+1$.

 \textbf{Step 4:} Obtain the $n\times \textbf{K}$ matrix $R$  of entry-wise eigen-ratios such that
$R(i,k)=X_{k+1}(i)/X_{1}(i), 1\leq i\leq n, 1\leq k\leq K$.

 \textbf{Step 5:} Apply K-means to $R$ assuming there are $K$ clusters.

Note that, by Lemma 2.5 in \cite{SCORE}, since $\mathscr{L}_{\tau_{2}}$ is a connected matrix (i.e., it does not have dis-connected parts), $\lambda_{1}$ is nonzero and all elements of $\eta_{1}$ are nonzero, hence $\lambda_{1}\eta_{1}(i)$ can be in the denominator, and $R$ is well defined. Applying K-means to $R$ leads to the true community labels of
each node. The following population analysis for DRSCORE method presents a direct understanding of
why the DRSCORE algorithm works and guarantees that DRSCORE returns perfect clustering results under the ideal case.

%The population analysis of DRSCORE aims at confirming that the Ideal DRSCORE algorithm can return perfect clustering. Lemma \ref{population_DRSCORE} guarantees that the Ideal DRSCORE algorithm surely provides true nodes labels.
\begin{lem}\label{population_DRSCORE}
Under $DCSBM(n, P, \Theta, Z)$, $R$ has $K$ distinct rows and for any two distinct nodes $i,j$ if $g_{i}=g_{j}$, then the $j$-th row of $R$ equals the $i$-th row of it.
\end{lem}

\subsection{Main Results for DRSLIM}
\subsubsection{Characterization of the matrix $\hat{M}$}
Similar as the procedures of theoretical analysis for DRSC and DRSCORE, for the population analysis of DRSLIM, first we present its  ideal case:

 \textbf{Ideal DRSLIM}. Input: $\Omega$. Output: $\ell$.

 \textbf{Step 1:} Obtain $\mathscr{L}_{\tau_{1}}$.

 \textbf{Step 2:} Obtain $\mathscr{L}_{\tau_{2}}$.

 \textbf{Step 3:} Obtain $W=(I-e^{-\gamma}\mathscr{D}^{-1}_{\tau_{2}}\mathscr{L}_{\tau_{2}})^{-1}$.
Calculate $M=(W+W')/2$ and force $M$'s diagonal entries to be 0.

 \textbf{Step 4:} Obtain the $n\times \textbf{(K+2)}$ matrix $\tilde{X}$  such that
$\tilde{X}=[\tilde{\eta}_{1},\tilde{\eta}_{2}, \ldots, \tilde{\eta}_{K+1}, \tilde{\eta}_{K+2}]\cdot \mathrm{diag}(\tilde{\lambda}_{1}, \tilde{\lambda}_{2}, \ldots, \tilde{\lambda}_{K+1}, \tilde{\lambda}_{K+2})$, where $\{\tilde{\lambda}_{i}\}_{i=1}^{K+2}$ are the leading eigenvalues of $M$, $\{\tilde{\eta}_{i}\}_{i=1}^{K+2}$  are the respective eigenvectors with unit-norm.

\textbf{Step 5:} Obtain $\tilde{X}^{*}$ by normalizing each of $\tilde{X}$'s rows to have unit length.

 \textbf{Step 6:} Apply K-means to $\tilde{X}^{*}$ assuming there are $K$ clusters.

 Since $W$ is nonsingular, clearly $M$ is also nonsingular, which indicates that the $n\times n$ matrix $M$ has $n$ nonzero eigenvalues. Therefore, $M$ does not share similar explicit expressions as $\mathscr{L}_{\tau_{1}}$ and $\mathscr{L}_{\tau_{2}}$, so it is challenging to find the explicit expressions of the eigenvectors of $M$. %Instead, we will give the concentration of $\hat{M}$ to bound $\|\hat{M}-M\|$ after introducing Lemma \ref{boundW} which bounds $\|\hat{W}\|$ and $\|W\|$, then we bound $\|\check{X}-\tilde{X}\|_{F}$ after bounding the differences between eigenvalues and eigenvectors of $\hat{M}$ and $M$, and finally we propose the bound of the Hamming error rate for DRSLIM.

 For convenience, in DRSLIM, we set $\varsigma=e^{-\gamma}$ and $Err_{n}$ as
%\begin{lem}\label{boundW}
%Under $DCSBM(n, P, \Theta, Z)$, if we assume that (c)
%\begin{align*}
%\varsigma< \mathrm{min}\{\frac{(\tau_{2}\Delta_{\mathrm{max}}+\tau_{1}\tau_{2}+\Delta_{\mathrm{min}})^{2}}{\tau_{1}\Delta_{\mathrm{max}}+\Delta^{2}_{\mathrm{max}}}, \frac{(\tau_{2}\delta_{\mathrm{max}}+\tau_{1}\tau_{2}+\delta_{\mathrm{min}})^{2}}{\tau_{1}\delta_{\mathrm{max}}+\delta^{2}_{\mathrm{max}}}\},
%\end{align*}
%then we have
%\begin{align*}
%\|\hat{W}\|\leq \frac{1}{1-\varsigma\frac{\tau_{1}\Delta_{\mathrm{max}}+\Delta^{2}_{\mathrm{max}}}{(\tau_{2}\Delta_{\mathrm{max}}+\tau_{1}\tau_{2}+\Delta_{\mathrm{min}})^{2}}
%}\mathrm{~and~}\|W\|\leq \frac{1}{1-\varsigma\frac{\tau_{1}\delta_{\mathrm{max}}+\delta^{2}_{\mathrm{max}}}{(\tau_{2}\delta_{\mathrm{max}}+\tau_{1}\tau_{2}+\delta_{\mathrm{min}})^{2}}
%}.
%\end{align*}
%\end{lem}
%For notation convenience, we set one parameter $Err_{n}$ as below
\begin{align*}
Err_{n}=&\varsigma (\frac{1}{1-\varsigma\frac{\tau_{1}\Delta_{\mathrm{max}}+\Delta^{2}_{\mathrm{max}}}{(\tau_{2}\Delta_{\mathrm{max}}+\tau_{1}\tau_{2}+\Delta_{\mathrm{min}})^{2}}
})(\frac{1}{1-\varsigma\frac{\tau_{1}\delta_{\mathrm{max}}+\delta^{2}_{\mathrm{max}}}{(\tau_{2}\delta_{\mathrm{max}}+\tau_{1}\tau_{2}+\delta_{\mathrm{min}})^{2}}})\\
&\times (\frac{err_{n}}{\tau_{2}+\frac{\Delta_{\mathrm{min}}}{\tau_{1}+\Delta_{\mathrm{max}}}}+\frac{(\tau_{1}+\Delta_{\mathrm{max}})\delta_{\mathrm{max}}\varpi_{b}}{(\tau_{1}\tau_{2}+\tau_{2}\Delta_{\mathrm{max}}+\Delta_{\mathrm{min}})(\tau_{1}\tau_{2}+\tau_{2}\delta_{\mathrm{max}}+\delta_{\mathrm{min}})}).
\end{align*}
%Next lemma bounds $\|\hat{M}-M\|$.
%\begin{lem}\label{boundM}
%Under $DCSBM(n, P, \Theta, Z)$, if assumptions (a) and (c) hold, with probability at least $1-\epsilon$, we have
%\begin{align*}
%\|\hat{M}-M\|\leq Err_{n}.
%\end{align*}
%\end{lem}
In the theoretical analysis for DRSLIM, without causing confusion, we set  four matrices as follows
\begin{align*}
&\check{V}=[\check{\eta}_{1}, \check{\eta}_{2}, \ldots, \check{\eta}_{K}, \check{\eta}_{K+1}, \check{\eta}_{K+2}], \tilde{V}=[\tilde{\eta}_{1}, \tilde{\eta}_{2}, \ldots, \tilde{\eta}_{K}, \tilde{\eta}_{K+1}, \tilde{\eta}_{K+2}],\\
&\check{E}=\mathrm{diag}(\check{\lambda}_{1}, \check{\lambda}_{2}, \ldots, \check{\lambda}_{K}, \check{\lambda}_{K+1}, \check{\lambda}_{K+2}), \tilde{E}=\mathrm{diag}(\tilde{\lambda}_{1}, \tilde{\lambda}_{2}, \ldots, \tilde{\lambda}_{K}, \tilde{\lambda}_{K+1}, \tilde{\lambda}_{K+2}).
\end{align*}
Similar as Lemma \ref{boundeigenvalue} and \ref{boundeigenvector}, we consider the bounds for  $\underset{1\leq k\leq K+2}{\mathrm{max}}\{|\check{\lambda}_{k}-\tilde{\lambda}_{k}|\}$ and $\|\check{V}-\tilde{V}\|_{F}$.
\begin{lem}\label{boundeigenvalueDRSLIM}
Under $DCSBM(n, P, \Theta, Z)$, if assumptions (a) and (c)
\begin{align*}
\varsigma< \mathrm{min}\{\frac{(\tau_{2}\Delta_{\mathrm{max}}+\tau_{1}\tau_{2}+\Delta_{\mathrm{min}})^{2}}{\tau_{1}\Delta_{\mathrm{max}}+\Delta^{2}_{\mathrm{max}}}, \frac{(\tau_{2}\delta_{\mathrm{max}}+\tau_{1}\tau_{2}+\delta_{\mathrm{min}})^{2}}{\tau_{1}\delta_{\mathrm{max}}+\delta^{2}_{\mathrm{max}}}\},
\end{align*}
hold, then with probability at least $1-\epsilon$, we have
\begin{align*}
\underset{1\leq k\leq K+2}{\mathrm{max}}\{|\check{\lambda}_{k}-\tilde{\lambda}_{k}|\}\leq Err_{n}.
\end{align*}
\end{lem}
%The proof of Lemma \ref{boundeigenvalueDRSLIM} is similar as lemma \ref{boundeigenvalue}, hence we omit the detail here.

\begin{lem}\label{boundVDRSLIM}
Under $DCSBM(n, P, \Theta, Z)$, if assumptions (a) and (c) hold, and assume that (d) $\tilde{\lambda}_{1}\geq \ldots \geq \tilde{\lambda}_{K+2}>0, \check{\lambda}_{1}\geq \ldots \geq \check{\lambda}_{K+2}>0$, and $\tilde{\lambda}_{K+2}>|\tilde{\lambda}_{i}|, \check{\lambda}_{K+2}>|\check{\lambda}_{i}|$ for $i=K+3,K+4,\ldots, n$ hold, then with probability at least $1-\epsilon$, we have
\begin{align*}
\|\check{V}-\tilde{V}\|_{F}\leq \frac{\sqrt{8(K+2)}Err_{n}}{\tilde{\lambda}_{k+2}-\tilde{\lambda}_{k+3}}.
\end{align*}
\end{lem}
Theorem \ref{boundstarDRSLIM} provides the bound of $\|\check{X}^{*}-\tilde{X}^{*}\|_{F}$, which is the corner stone to characterize the behavior of our DRSC approach.
\begin{thm}\label{boundstarDRSLIM}
Under $DCSBM(n, P, \Theta, Z)$, define $m_{b}=\mathrm{min}_{i}\{\mathrm{min}\{\|\check{X}_{i}\|, \|\tilde{X}_{i}\|\}\}$ as the length of the shortest row in $\check{X}$ and $\tilde{X}$. Then, for any $\epsilon >0$ and sufficiently large $n$,  assume that assumptions (a), (c) and (d) in Lemma \ref{boundVDRSLIM} hold,
then with probability at least $1-\epsilon$, the followings hold
\begin{align*}
&\|\check{X}-\tilde{X}\|_{F}\leq Err_{n}\sqrt{K+2}+\frac{2^{3/2}(K+2)\tilde{\lambda}_{1}Err_{n}}{\tilde{\lambda}_{k+2}-\tilde{\lambda}_{k+3}},\\
&\|\check{X}^{*}-\tilde{X}^{*}\|_{F}\leq \frac{1}{m_{b}}(Err_{n}\sqrt{K+2}+\frac{2^{3/2}(K+2)\tilde{\lambda}_{1}Err_{n}}{\tilde{\lambda}_{k+2}-\tilde{\lambda}_{k+3}}).
\end{align*}
\end{thm}
\subsubsection{Bound of Hamming error rate of DRSLIM}
The following theorem is the main theoretical result which bounds the Hamming error rate of our DRSLIM method under mild conditions and shows that the DRSLIM method stably yields consistent community detection under several conditions.% if we assume that the adjacency matrix $A$ are generated from the DCSBM model.
\begin{thm}\label{mainDRSLIM}
Under the DCSBM with parameters $\{n, P, \Theta, Z\}$ and the same assumptions as in Theorem \ref{boundstarDRSLIM} hold, suppose as $n\rightarrow \infty$, we have
\begin{align*}
 \frac{4}{m^{2}_{b}}(Err_{n}\sqrt{K+2}+\frac{2^{3/2}(K+2)\tilde{\lambda}_{1}Err_{n}}{\tilde{\lambda}_{k+2}-\tilde{\lambda}_{k+3}})^{2}
/\mathrm{min~}\{n_{1}, n_{2}, \ldots, n_{K}\}\rightarrow 0.
\end{align*}
For the estimates label vector $\check{\ell}$ by DRSLIM, such that with probability at least $1-\epsilon$, we have
\begin{align*}
  \mathrm{Hamm}_{n}(\check{\ell},\ell)\leq \frac{4}{nm^{2}_{b}}(Err_{n}\sqrt{K+2}+\frac{2^{3/2}(K+2)\tilde{\lambda}_{1}Err_{n}}{\tilde{\lambda}_{k+2}-\tilde{\lambda}_{k+3}})^{2}.
  \end{align*}
\end{thm}
%Since the proof of Theorem \ref{mainDRSLIM} follows a similar procedure as that of Theorem \ref{mainDRSC}, we omit the detail here.
\section{Numerical Results}\label{sec5}
We compare our DRSC, DRSCORE and DRSLIM with a few recent methods: RSC \citep{RSC}, SCORE \citep{SCORE}, SLIM \citep{SLIM} and OCCAM \citep{OCCAM} via synthetic data and eight real-world networks. It needs to mention that overlapping continuous community assignment model (OCCAM) is designed for overlapping communities by \cite{OCCAM}  which is also a spectral clustering algorithm via applying K-median method for clustering instead of K-means.

For each procedure, the clustering error rate is measured by
\begin{align*}
  \mathrm{min}_{\{\pi: \mathrm{permutation~over~}\{1,2,\ldots, K\}\}}\frac{1}{n}\sum_{i=1}^{n}1\{\pi(\hat{\ell}_{i})\neq \ell_{i}\},
\end{align*}
where $\ell_{i}$ and $\hat{\ell}_{i}$ are the true and estimated labels of node $i$.
\subsection{Synthetic data experiment}
In this subsection, we use three simulated
experiments to investigate the performance of these approaches.

\textbf{\texttt{Experiment 1}}. In this experiment, we  investigate performances of these approaches under SBM when $K=2$ and 3 by increasing $n$. Set $n\in \{50, 100, 150, \ldots, 500\}$. For each fixed $n$, we record the mean of the error rate of 50 repetitions.

\emph{Experiment 1(a)}.  When $K=2$, we generate $\ell$ by setting each node belonging to one of the clusters with equal probability (i.e., $\ell_{i}-1 \overset{\mathrm{i.i.d.}}{\sim}\mathrm{Bernoulli}(1/2)$). Set the mixing matrix $P_{1(a)}$ as
 \[
P_{1(a)}
=
\begin{bmatrix}
    1&0.6\\
    0.6&1\\
\end{bmatrix}.
\]
Generate $\theta$ as $\theta(i)=0.3$ for $g_{i}=1$, $\theta(i)=0.7$ for $g_{i}=2$.

\emph{Experiment 1(b)}. When $K=3$, we generate $\ell$ by setting each node belonging to one of the clusters with equal probability. The mixing matrix $P_{1(b)}$ is
 \[
P_{1(b)}
=
\begin{bmatrix}
    1&0.6&0.6\\
    0.6&1&0.6\\
    0.6&0.6&1\\
\end{bmatrix}.
\]
Generate $\theta$ as $\theta(i)=0.3$ for $g_{i}=1$, $\theta(i)=0.5$ for $g_{i}=2$, and $\theta(i)=0.7$ for $g_{i}=3$.

\begin{figure}[H]
	\centering
\includegraphics[width=12cm,height=4cm]{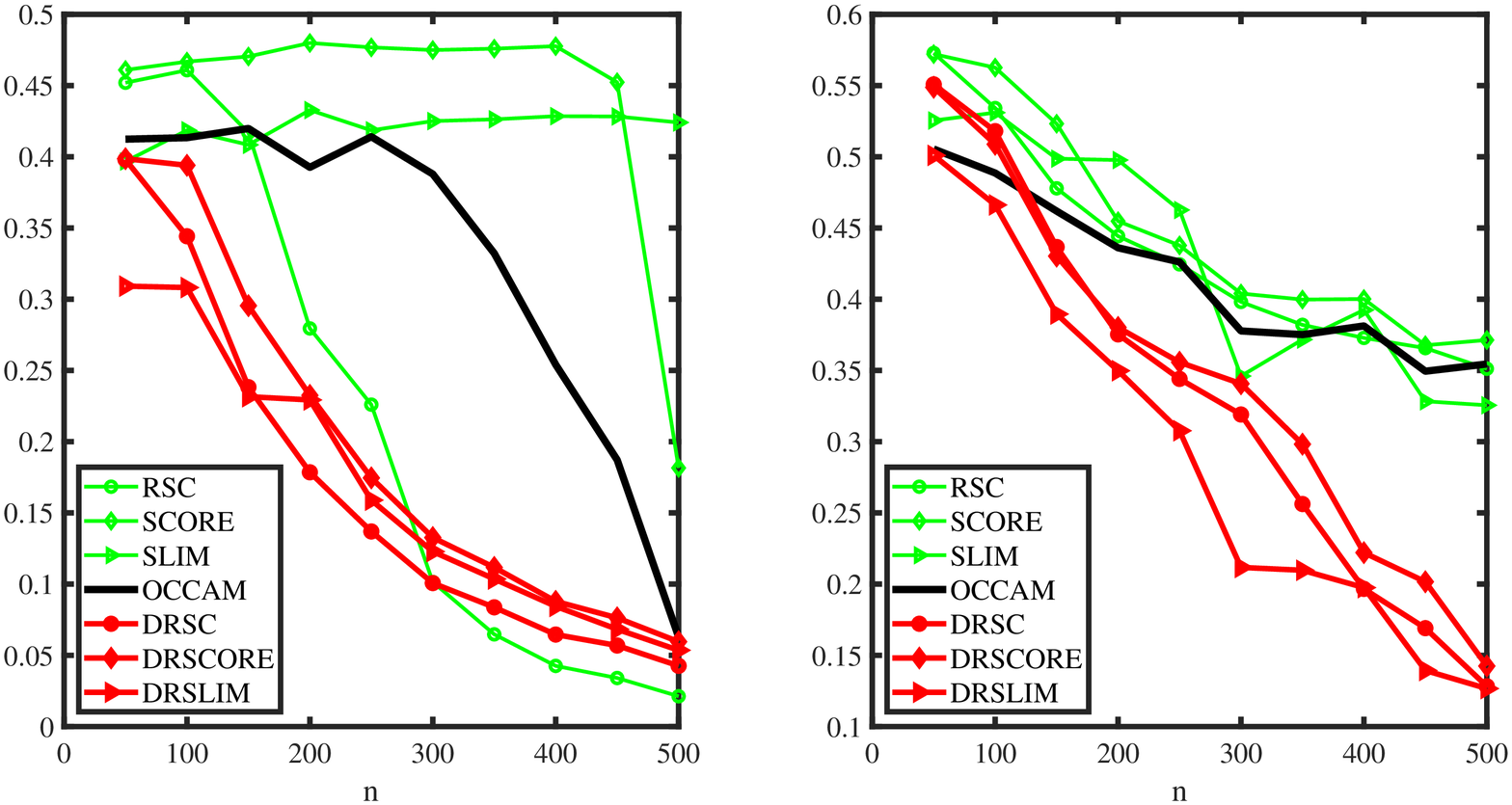}
\caption{Numerical results of Experiment 1. Left panel: Experiment 1(a). Right panel: Experiment 1(b). y-axis: error rates.}\label{Ex1}
\end{figure}

The numerical results of Experiment 1 are shown in Figure \ref{Ex1} from which we can have the following conclusions. When the number of clusters is 2, as $n$ increases, error rates of RSC, DRSC, DRSCORE, and DRSLIM decrease rapidly, while error rates for SCORE, SLIM and OCCAM decrease quite slowly. When we increase the number of clusters from 2 to 3, the error rates of RSC, SCORE, SLIM and OCCOM are always quite large though $n$ increases. Overall, our DRSC, DRSCORE and DRSLIM have comparable performances and almost always outperform the other four procedures obviously in this Experiment.
\textbf{\texttt{Experiment 2}}. In this experiment, we study how the ratios of
 $\theta$, the ratios of the size between different communities and the ratios of diagonal and off-diagonal entries of the mixing matrix impact behaviors of our proposed approaches when $n=500$.

\emph{Experiment 2(a)}. We study the influence of ratios of $\theta$ when $K=2$ under SBM. Set the proportion $a_{0}$ in $\{1, 1.2, 1.4, \ldots, 4\}$. We generate $\ell$  by setting each node belonging to one of the clusters with equal probability. And the mixing matrix $P_{2(a)}$ is
 \[\renewcommand{\arraystretch}{0.75}
P_{2(a)}
=
\begin{bmatrix}
    0.5&0.3\\
    0.3&0.5\\
\end{bmatrix}.
\]
Generate $\theta$ as $\theta(i)=1$ if $g_{i}=1$, $\theta(i)=1/a_{0}$ if $g_{i}=2$. Note that for each fixed $a_{0}$, $\theta(i)$ is a fixed number for all nodes in the same community and hence this is a SBM case. Meanwhile, for each fixed $a_{0}$, we record the mean of clustering error rates of 50 sampled networks.

\emph{Experiment 2(b)}. We study how the sparsity of networks affects the performance of these methods under SBM in this sub-experiment. Let the proportion $b_{0}$  in $\{1,1.5, 2, \ldots, 5\}$. We generate $\ell$  by setting each node belonging to one of the clusters with equal probability. Set $\theta$ as $\theta(i)=1$  for all $i$. The mixing matrix $P_{2(b)}$ is set as:
 \[\renewcommand{\arraystretch}{0.75}
P_{2(b)}
=
b_{0}\begin{bmatrix}
   0.2& 0.15\\
    0.15 & 0.2\\
\end{bmatrix}.
\]
For each fixed $b_{0}$, we record the average of the clustering error rates of 50 simulated networks. Note that as $b_{0}$ increases, more edges are generated, therefore the generated network is more dense.

\emph{Experiment 2(c)}. All parameters are the same as in Experiment 2(b), except that the mixing matrix $P_{2(c)}$ is as follows:
 \[\renewcommand{\arraystretch}{0.75}
P_{2(c)}
=
b_{0}\begin{bmatrix}
   0.15& 0.2\\
    0.2 & 0.15\\
\end{bmatrix}.
\]
Therefore, networks generated from Experiment 2(c) are dis-associative where dis-associative networks denote networks generated from the mixing matrix such that the off-diagonal entries are larger than the diagonal entries, i.e., there are more edges between nodes from distinct clusters than from the same cluster.

\emph{Experiment 2(d)}. We study how the proportion between the size of clusters influences the performance of these methods under SBM in this sub-experiment. We set the proportion $c_{0}$ in $\{1,2,\ldots,10\}$. Set $n_{1}= \mathrm{round}(\frac{n}{c_{0}+1})$
as the number of nodes in cluster 1 where $\mathrm{round}(x)$ denotes the nearest integer for any real number $x$. We generate $\ell$ such that $ g_{i}=1$ for $i=1, \cdots, n_{1},$ and $g_{i}=2$ for $i=(n_{1}+1), \cdots, n$. Note that $c_{0}$ is the ratio \footnote{Number of nodes in cluster 2 is $n-\mathrm{round}(\frac{n}{c_{0}+1})\approx n-\frac{n}{c_{0}+1}=c_{0}\frac{n}{c_{0}+1}$, therefore
number of nodes in cluster 2 is $c_{0}$ times of that in cluster 1.} of the size of
cluster 2 and cluster 1. And the mixing matrix $P_{2(d)}$ is set as:
  \[\renewcommand{\arraystretch}{0.75}
P_{2(d)}
=
\begin{bmatrix}
    0.9&0.6\\
    0.6&0.8
\end{bmatrix}.
\]
Let $\theta$ be $\theta(i)=0.6$ if $g_{i}=1$ and $\theta(i)=0.9$ otherwise. For each fixed $c_{0}$, we record the average for the clustering error rates of 50 simulated networks.

\emph{Experiment 2(e)}. All parameters are the same as in Experiment 2(d), except that we set $\theta$ as $\theta_{i}=0.6+0.4(i/n)^{2}$ for $1\leq i\leq n$ (i.e., Experiment 2(e) is the DCSBM case).

\emph{Experiment 2(f)}. We study how the probability of nodes belong to distinct communities influences the performance of these methods when $K=2$ in this sub-experiment. Set the proportion $d_{0}$ in $\{0, 0.05, 0.1,\ldots,0.45\}$. We generate $\ell$ such that nodes belong to cluster 1 with probability $0.5-d_{0}$, nodes belong to cluster 2 with probability $0.5+d_{0}$. Therefore, as $d_{0}$ increases from 0 to 0.45, number of nodes in cluster 1 decreases, so this is a case that is more challenge to detect. The mixing matrix is same as in Experiment 2(d). Let $\theta$ be $\theta(i)=0.4$ if $g_{i}=1$ and $\theta(i)=0.6$ otherwise. For each fixed $d_{0}$, we record the average for the clustering error rates of 50 simulated networks.

\begin{figure}[H]
	\centering
\includegraphics[width=14cm,height=8cm]{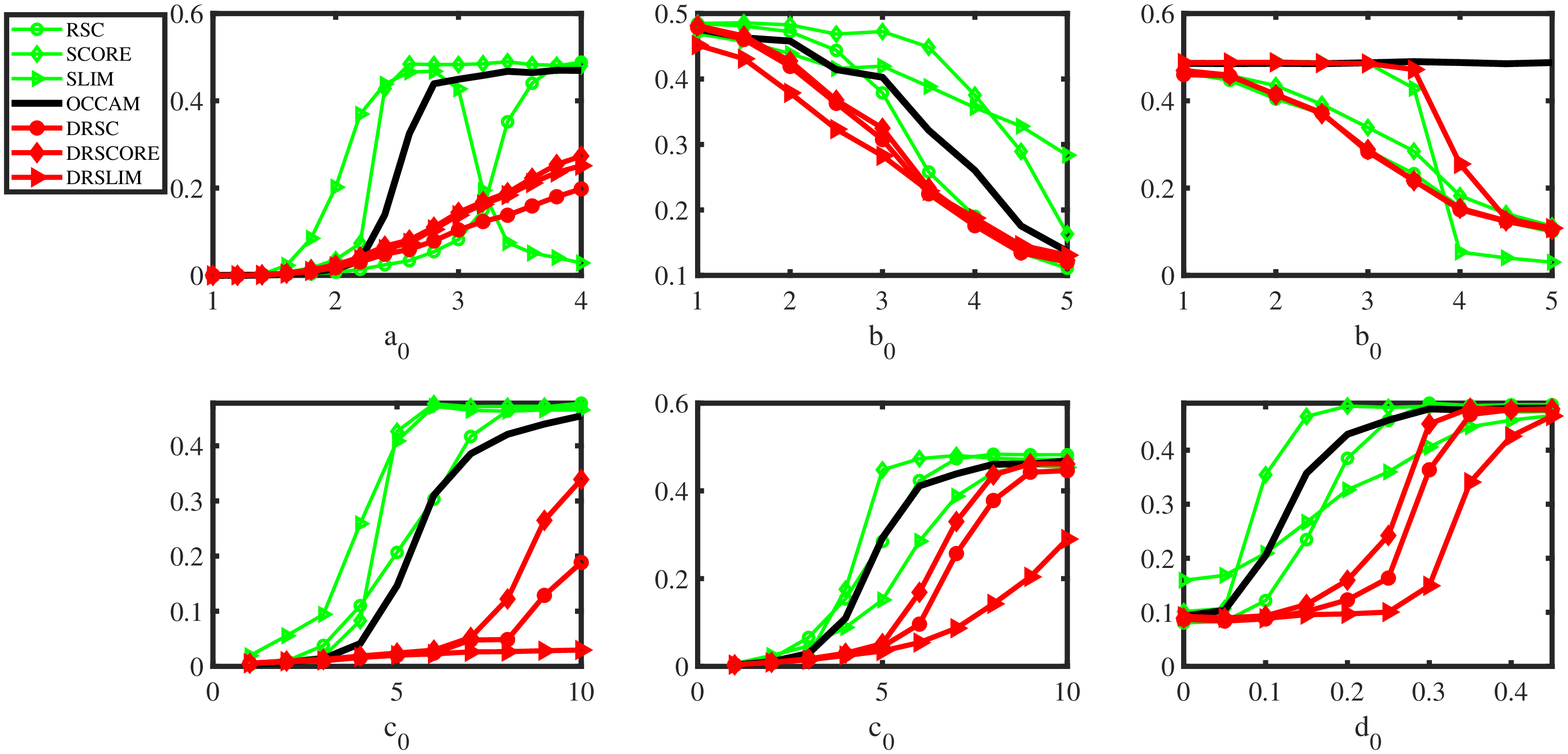}
\caption{Numerical results of Experiment 2. Top three panels (from left to right): Experiment 2(a), Experiment 2(b), and Experiment 2(c). Bottom three panels (from left to right): Experiment 2(d), Experiment 2(e), and Experiment 2(f). y-axis: error rates.  }\label{Ex2}%Experiment 2 shares the same legends as Experiment 1.
\end{figure}

The numerical results of Experiment 2 are shown by Figure \ref{Ex2}.
 In Experiment 2(a), as $a_{0}$ increases, $\theta(i)$ decreases for node $i$ in cluster 2, which suggests that there are fewer edges generated in cluster 2, hence it become more challenging to detect cluster 2. Numerical results of Experiment 2(a) suggests that as the variability of degree increases, all approaches (except SLIM) perform poorer while our DRSC, DRSCORE and DRSLIM perform better than the three traditional spectral clustering methods  RSC, SCORE and OCCAM.
 However, it is interesting to find that SLIM has abnormal performances: its error rate decreases when $a_{0}$ increases from 3 to 4. We can not explain the abnormal behavior of SLIM at present, and we leave it for our future work. In Experiment 2(b), all methods perform better when the simulated network becomes denser, meanwhile, our three approaches DRSC, DRSCORE and DRSLIM have better performances than that of RSC, SCORE, OCCAM and SLIM. Numerical results of Experiment 2(c) say that all methods can detect dis-associative networks while OCCAM fails. When the off-diagonal entries of the mixing matrix are close to its diagonal entries, all methods have poor performances. Numerical results of Experiment 2(d), 2(e), and 2(f) tell us that though it becomes challenging for all approaches to have satisfactory detection performances for a fixed size network when the size of one of the cluster decreases, our approaches DRSC, DRSCORE and DRSLIM always outperform the other four comparison approaches. %, suggesting that our three methods can successfully detect communities of a given network with various size of clusters.
  In all, our DRSC, DRSCORE and DRSLIM always outperform other approaches in this experiment.

\emph{Remark:} \textit{Recall that in the theoretical analysis for DRSC and DRSLIM, we assume that $\lambda_{1}\geq \ldots\geq \lambda_{K}>0$ and $\tilde{\lambda}_{1}\geq \ldots\geq \tilde{\lambda}_{K}>0$ (i.e., the networks generated under DCSBM should be associative network), combine with DRSC and DRSLIM's satisfactory performances in Experiment 2(c), we argue that our DRSC and DRSLIM can detect dis-associative networks as well. This phenomenon suggests that the Davia-Kahan theorem  may be extended to the case that $S$ (defined in Lemma \ref{QinDK}) has several disjoint sets, and we leave the study of this conjecture for future work.}

\textbf{\texttt{Experiment 3}}. This experiment contains two sub-experiments. In both two sub-experiments, we set $n=500, K=4$, generate $\ell$ by setting each node belonging to one of the clusters with equal probability, and set $\theta_{i}=1$ if $g_{i}=1$, $\theta_{i}=0.8$ if $g_{i}=2$, $\theta_{i}=0.6$ if $g_{i}=3$, and $\theta_{i}=0.4$ if $g_{i}=4$. %In experiment 3(a), we use simulated data to evaluate the performances of our methods by increasing the diagonal entries of $P$ while off-diagonal entries are fixed under SBM. In experiment 3(b), to generate denser networks, we increase all entries of $P$.
 Set $\alpha, \beta$  in $\{0, 1/20, 2/20, 3/20,\ldots, 12/20\}$, and for each fixed $\alpha$ (as well as $\beta$), we record the mean of the error rate of 50 simulated samples.

\emph{Experiment 3(a)}. We  increase the diagonal entries of $P$ while off-diagonal entries are fixed under SBM. The mixing matrix $P_{3(a)}$ is set as
  \[\renewcommand{\arraystretch}{0.75}
P_{3(a)}
=
\begin{bmatrix}
    0.4+\alpha&0.4&0.2&0.2\\
    0.4&0.4+\alpha&0.2&0.2\\
    0.2&0.2&0.4+\alpha&0.4\\
    0.2&0.2&0.4&0.4+\alpha\\
\end{bmatrix}.
\]
When $\alpha=0$, the networks have only two communities, and as $\alpha$ increases, the four communities become more distinguishable.

\emph{Experiment 3(b)}. We increase all entries of $P$  to generate denser networks. The mixing matrix $P_{3(b)}$ is set as
  \[\renewcommand{\arraystretch}{0.75}
P_{3(b)}
=
\begin{bmatrix}
    0.4+\beta&0.4&0.2+\beta&0.2+\beta\\
    0.4&0.4+\beta&0.2+\beta&0.2+\beta\\
    0.2+\beta&0.2+\beta&0.4+\beta&0.4\\
    0.2+\beta&0.2+\beta&0.4&0.4+\beta\\
\end{bmatrix}.
\]
When $\beta$ increases from 0 to 0.6, the four communities become more distinguishable and the simulated networks are more denser.
\begin{figure}[H]
	\centering
\includegraphics[width=12cm,height=4cm]{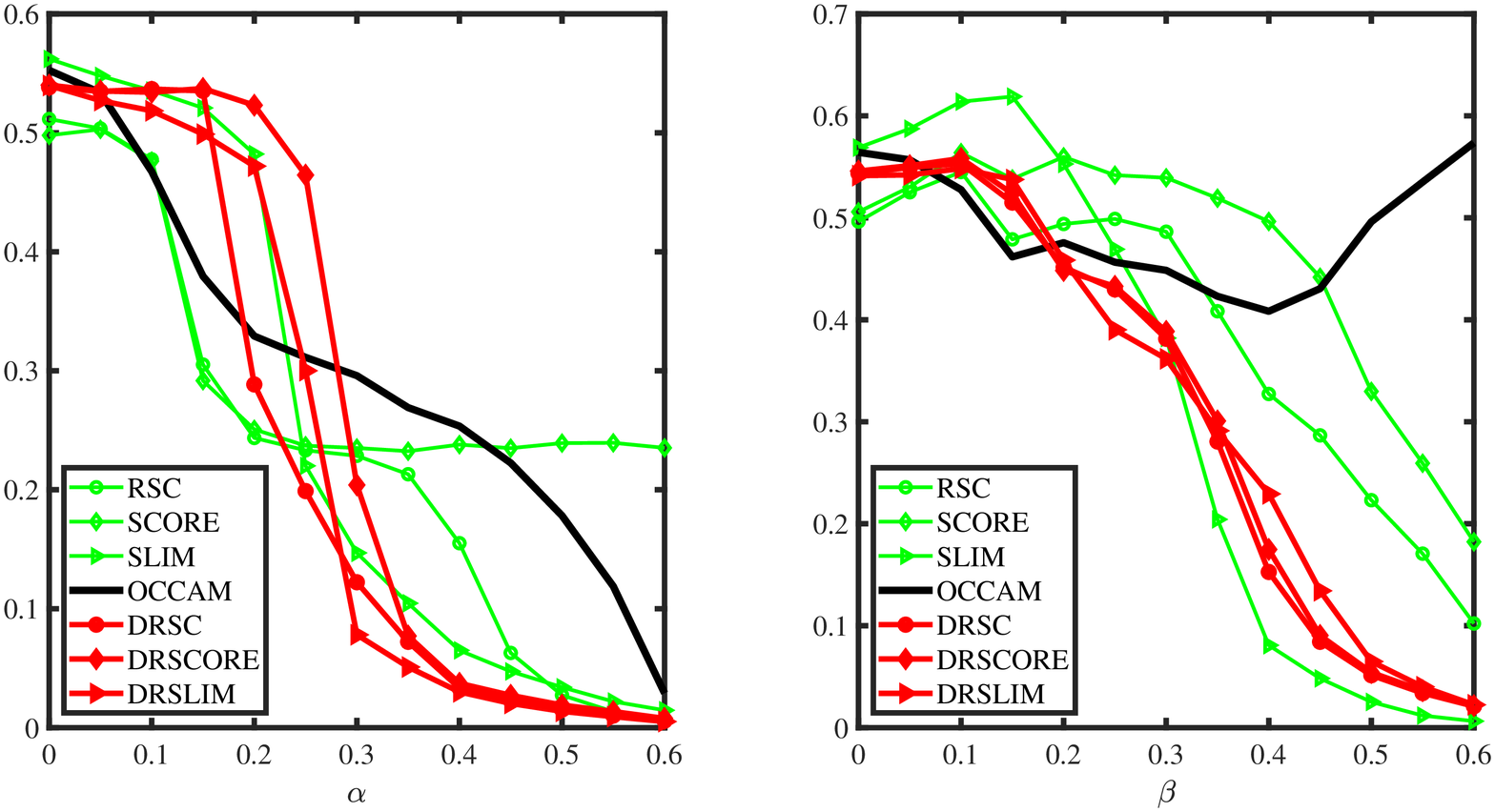}
\caption{Numerical results of Experiment 3. Left panel: Experiment 3(a). Right panel: Experiment 3(b). y-axis: error rates. }\label{Ex3}
\end{figure}
Numerical results of Experiment 3 is demonstrated by Figure \ref{Ex3}. In experiment 3(a), all methods perform poor when $\alpha$ is small, that is, a smaller $\alpha$ means that the four communities are more difficult to be distinguished.
 All methods except SCORE perform better as $\alpha$ increases for Experiment 3(a), while the performances of SCORE almost do not change when $\alpha$ increases from 0.3 to 0.6.
  Our DRSC, DRSCORE and DRSLIM obviously outperform RSC, SCORE and OCCAM in Experiment 3(b), however, SLIM performs similar as the proposed methods. It is interesting to find that OCCAM has abnormal behaviors that it performs even poorer when $\beta$ increases which gives a more denser simulated network, while the other six procedures can handle with a denser network.

\subsection{Application to real-world datasets}\label{secreal8}
In this paper, eight real-world network datasets are analyzed to test the performances of our DRSC, DRSCORE and DRSLIM.
The eight datasets are used in the paper \cite{SCORE+} and can be downloaded directly from
\url{http://zke.fas.harvard.edu/software.html}. Table \ref{real8} presents some basic information about the eight datasets. These eight datasets are networks with known labels for all nodes where the true label information is surveyed by researchers. From Table \ref{real8}, we can see that $d_{\mathrm{min}}$ and $d_{\mathrm{max}}$\footnote{Where $d_{\mathrm{min}}=\underset{i}{\mathrm{min}}D_{1}(i,i)$ and $d_{\mathrm{max}}=\underset{i}{\mathrm{max}}D_{1}(i,i)$.} are always quite different for any one of the eight real-world datasets, which suggests a DCSBM case. Readers who are interested in the background information of the eight real-world networks can refer to Appendix \ref{dereal8} for details.
\begin{table}[h!]
%\footnotesize
\centering
\caption{Eight real-world data sets with known label information analyzed in this paper.}
\label{real8}
\resizebox{\columnwidth}{!}{
\begin{tabular}{cccccccccc}
\toprule
\#&Karate&Dolphins&Football&Polbooks&UKfaculty&Polblogs&Simmons&Caltech\\
\midrule
$n$&34&62&110&92&79&1222&1137&590\\
$K$&2&2&11&2&3&2&4&8\\
$d_{\mathrm{min}}$&1&1&7&1&2&1&1&1\\
$d_{\mathrm{max}}$&17&12&13&24&39&351&293&179\\
\bottomrule
\end{tabular}}
\end{table}

\begin{table}[h!]
%\scriptsize
\centering
\caption{Error rates on the eight empirical data sets.}
\label{real8errors}
\resizebox{\columnwidth}{!}{
\begin{tabular}{cccccccccc}
\toprule
\textbf{ Methods} &Karate&Dolphins&Football&Polbooks&UKfaculty&Polblogs&Simmons&Caltech\\
\midrule
RSC&\textbf{0/34}&1/62&5/110&3/92&\textbf{0/79}&64/1222&244/1137&170/590\\
SCORE&\textbf{0/34}&\textbf{0/62}&5/110&\textbf{1/92}&1/79&58/1222&268/1137&180/590\\
SLIM&1/34&\textbf{0/62}&6/110&2/92&1/79&\textbf{51/1222}&275/1137&147/590\\
OCCAM&\textbf{0/34}&1/62&4/110&3/92&5/79&60/1222&268/1137&192/590\\
\hline
DRSC&\textbf{0/34}&1/62&5/110&3/92&2/79&63/1222&124/1137&95/590\\
$\mathrm{DRSC}_{K+2}$&\textbf{0/34}&\textbf{0/62}&\textbf{3/110}&2/92&2/79&63/1222&121/1137&98/590\\
DRSCORE&\textbf{0/34}&4/62&6/110&4/92&3/79&65/1222&117/1137&99/590\\
$\mathrm{DRSCORE}_{K+2}$&11/34&6/62&\textbf{3/110}&26/92&3/79&336/1222&271/1137&105/590\\
$\mathrm{DRSLIM}_{K+1}$&\textbf{0/34}&1/62&\textbf{3/110}&2/92&2/79&58/1222&186/1137&\textbf{92/590}\\
DRSLIM&\textbf{0/34}&\textbf{0/62}&\textbf{3/110}&2/92&2/79&59/1222&\textbf{115/1137}&98/590\\
\bottomrule
\end{tabular}}
\end{table}
Next we study the performances of these methods on the eight real-world networks. Recall that in Section \ref{DualLSC}, we said that we can apply the leading $(K+K_{0})$ eigenvectors and eigenvalues of $L_{\tau_{2}}$ and $M$ in our three methods, where $K_{0}$ can be 1 or 2. Here, if $K_{0}$ is 2 for DRSC and DRSCORE, then we call the two new methods as $\mathrm{DRSC}_{K+2}$ and $\mathrm{DRSCORE}_{K+2}$, and if $K_{0}$ is 1 for DRSLIM, we call the new method as $\mathrm{DRSLIM}_{K+1}$.   Table \ref{real8errors} records the error rates on the eight real-world networks, from which we can see that for Karate, Dolphins, Football, Polboks, UKfaculty, and Polblogs, our DRSC, DRSCORE, DRSLIM, $\mathrm{DRSC}_{K+2}$ and $\mathrm{DRSLIM}_{K+1}$ have similar performances as that of RSC, SCORE, OCCAM and SLIM. While,  for Simmons and Caltech,  our DRSC, DRSCORE, DRSLIM, $\mathrm{DRSC}_{K+2}$ and $\mathrm{DRSLIM}_{K+1}$ approaches detect clusters with much lower error numbers than RSC, SCORE, OCCAM and SLIM. This phenomenon occurs since Simmons and Caltech are two weak signal networks which is suggested by \cite{SCORE+} where weak signal networks are defined as network whose leading (K+1)-th eigenvalue of $A$ or its variants is close to its leading $K$-th eigenvalue, while the other six real-world datasets are deemed as strong signal networks.
% \cite{SCORE+} argues that the leading $(K+1)$-th eigenvector of $A$ or its variants may also contain information about nodes labels, therefore we'd apply one more eigenvectors for clustering when dealing with weak signal networks.
 Recall that our DRSC and DRSCORE apply the leading $(K+1)$ eigenvectors to construct $\hat{X}$ while our DRSLIM applies the leading $(K+2)$ eigenvectors of $\hat{M}$ to construct $\check{X}$, this is the reason that our three approaches outperform the other four approaches when detecting Simmons and Caltech. Though DRSCORE performs satisfactory on the eight empirical datasets, $\mathrm{DRSCORE}_{K+2}$ fail to detect Karate, Polbooks and Polblogs, which suggests that $K_{0}$ should be 1 for DRSCORE. Meanwhile, $\mathrm{DRSC}_{K+2}$ has similar performances as DRSC, hence $K_{0}$ can be 1 or 2 for DRSC, and we set $K_{0}$ as 1 in DRSC for simplicity. Finally ,though $\mathrm{DRSLIM}_{K+1}$ performs similar as DRSLIM generally, it performs poor on Simmons network, and this is the reason we set $K_{0}$ as 2 for our DRSLIM algorithm.
\subsection{Discussion on the choice of tuning parameters}
\begin{figure}
	\centering
\includegraphics[width=12cm,height=6cm]{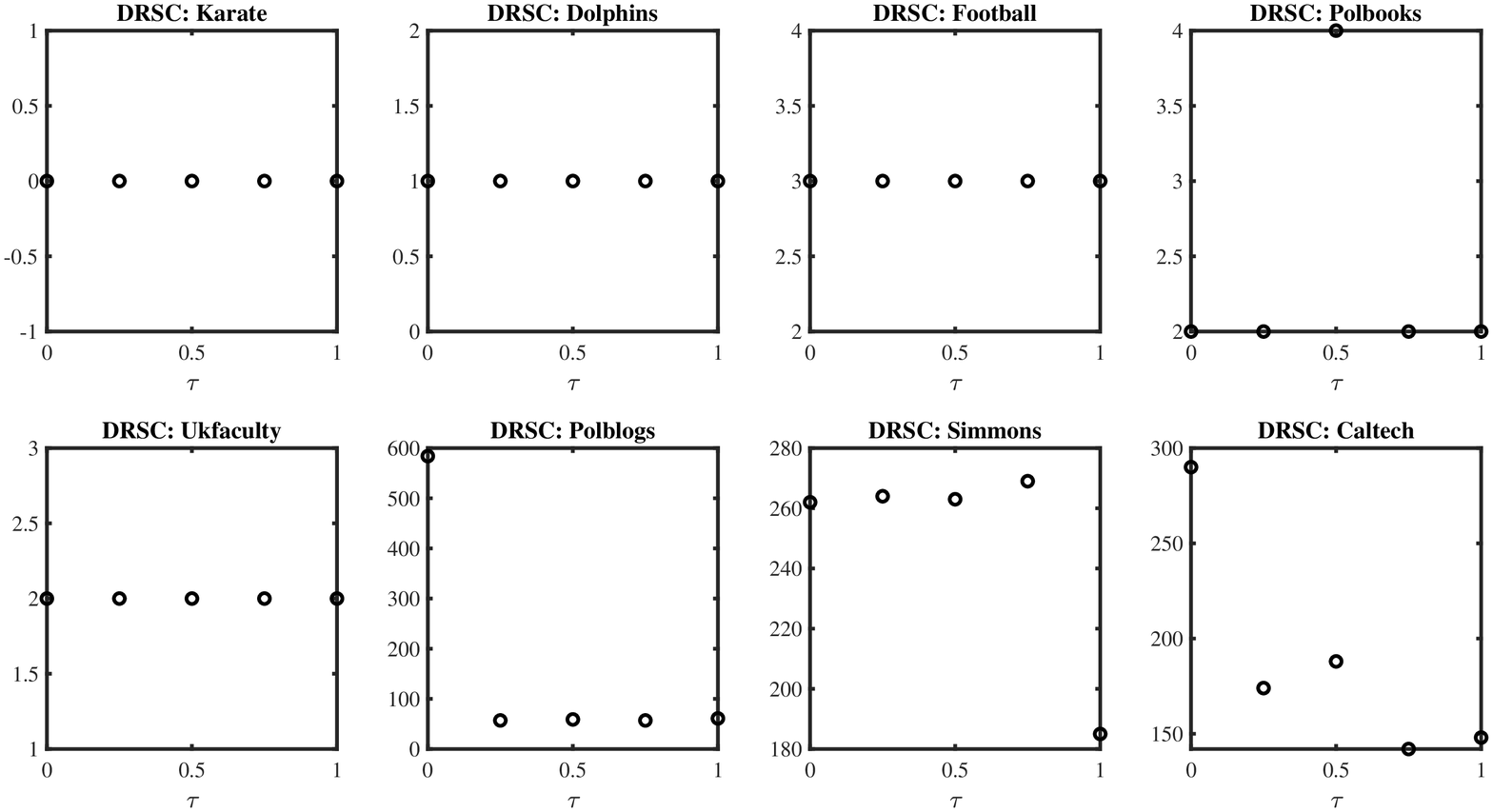}
\includegraphics[width=12cm,height=6cm]{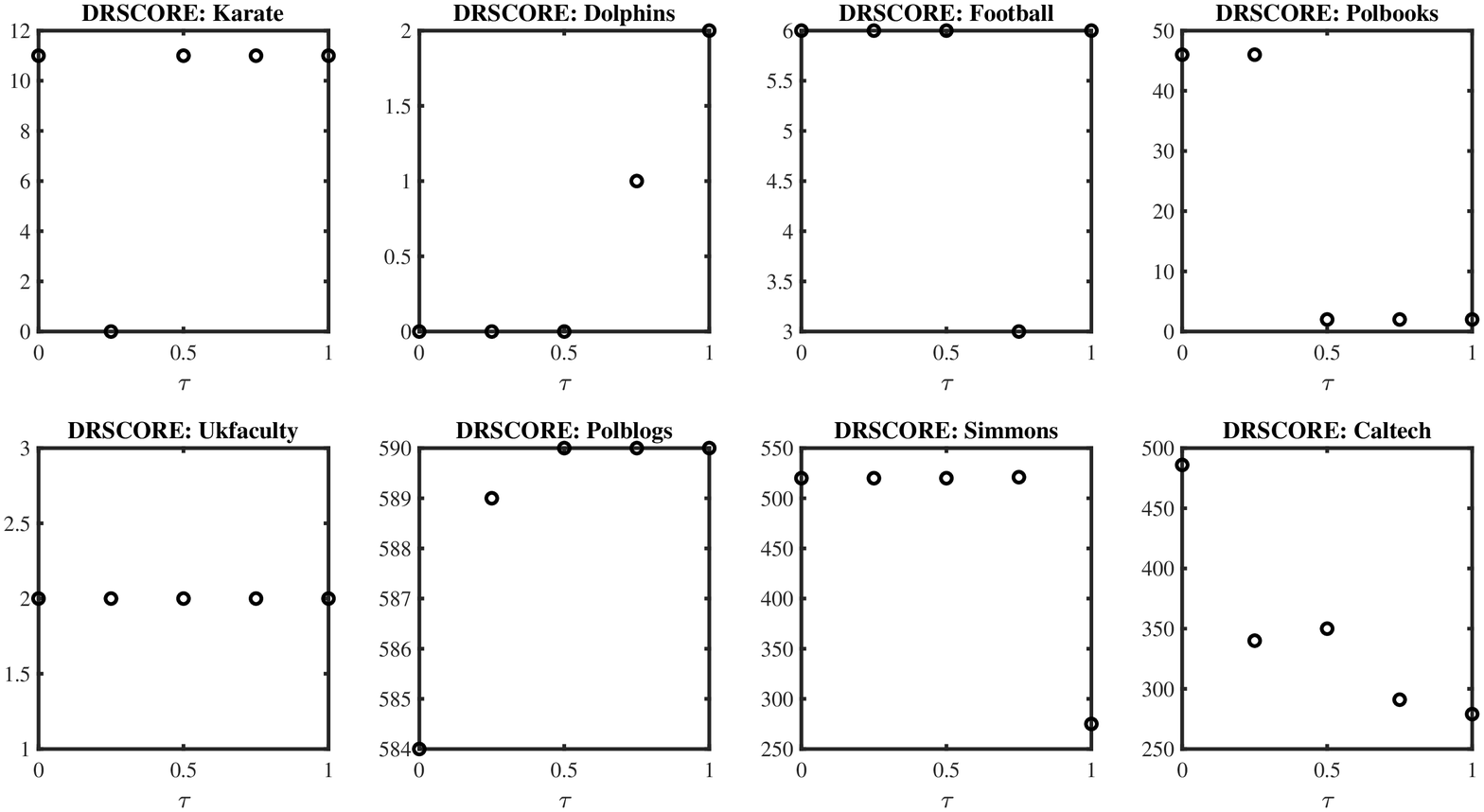}
\includegraphics[width=12cm,height=6cm]{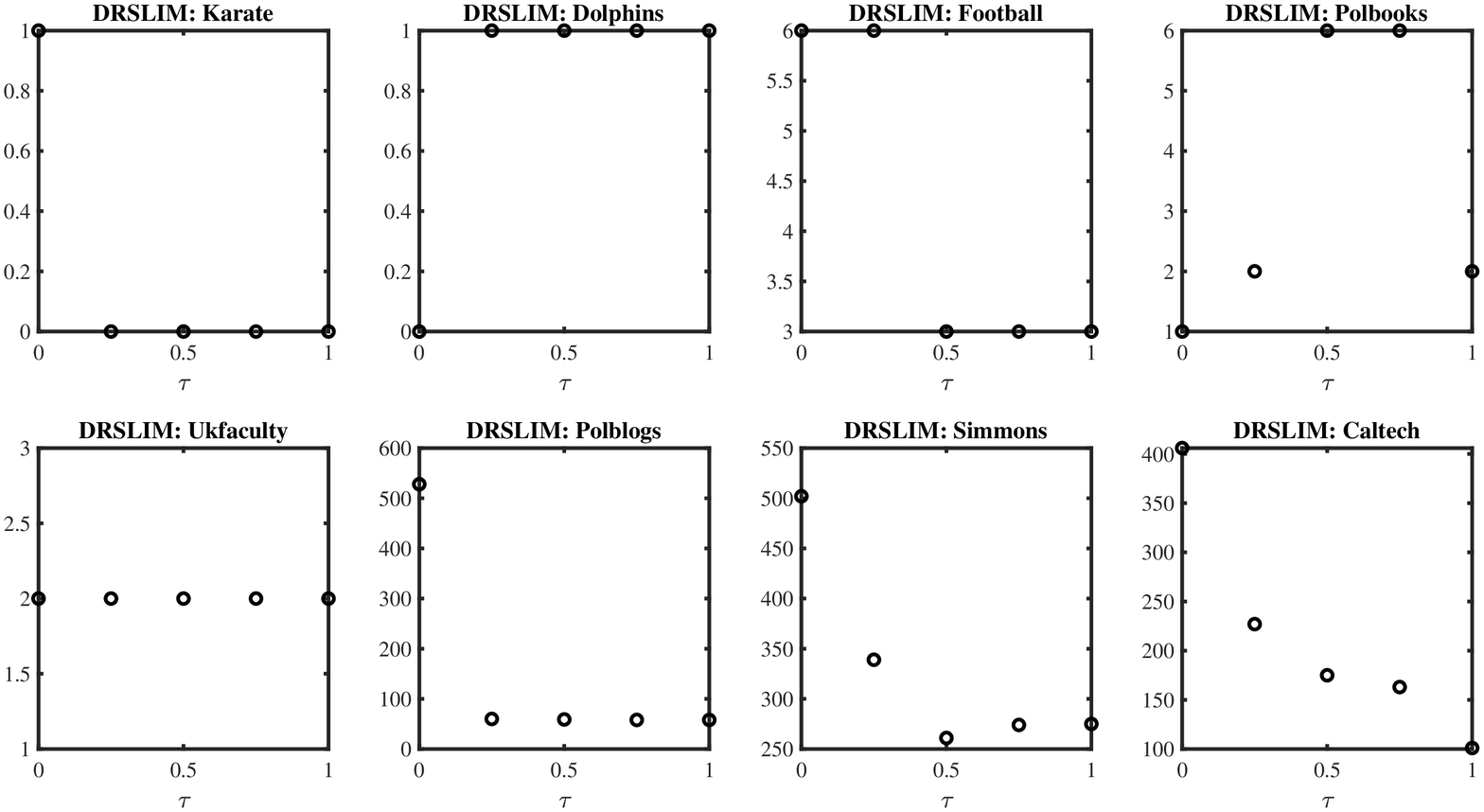}
\caption{Number errors on the eight empirical data sets for DRSC, DRSCORE and DRSLIM when $\tau$ is in $\{0,0.25, 0.5, 0.75, 11\}$. x-axis: $\tau$. y-axis: number errors.}
\label{tau1tau2_DR1}
\end{figure}

At present, there is no practical criterion to choose the optimal $\tau_{1}$ and $\tau_{2}$ in this paper. To have a better knowledge of the effect of $\tau_{1}$ and $\tau_{2}$ on the performances of DRSC, DRSCORE and DRSLIM, we study that whether the three procedures are sensitive to the choice of $\tau_{1}$ and $\tau_{2}$ here.  For convenience, set $\tau_{1}=\tau_{2}=\tau$. Figure \ref{tau1tau2_DR1} records error rates of the three dual regularized spectral clustering methods on the eight real-world datasets  when $\tau$ is in $\{0, 0.25, 0.5, 0.75, 1\}$ (i.e., $\tau$ is set as a small number).  From Figure \ref{tau1tau2_DR1}, we see that DRSC successfully detect Karate, Dolphins, Football, Polbooks and Ukfaculty. However, when $\tau_{1}=\tau_{2}=0$, DRSC has high error numbers for detecting Polblogs, Simmons and Caltech. Figure \ref{tau1tau2_DR1} shows that the  performances of DRSC on Simmons and Caltech are not satisfactory when $\tau$ is too small. When $\tau$ is too small, DRSCORE fails to detect Karate, Polbooks, Polblogs, Simmons and Caltech. Therefore DRSCORE is sensitive to the choice of $\tau$ if $\tau$ is too small. When $\tau$ is small, DRSLIM has similar performances on the eight empirical datasets as DRSC, hence it shares similar conclusions as DRSC.

\begin{figure}
	\centering
\includegraphics[width=12cm,height=6cm]{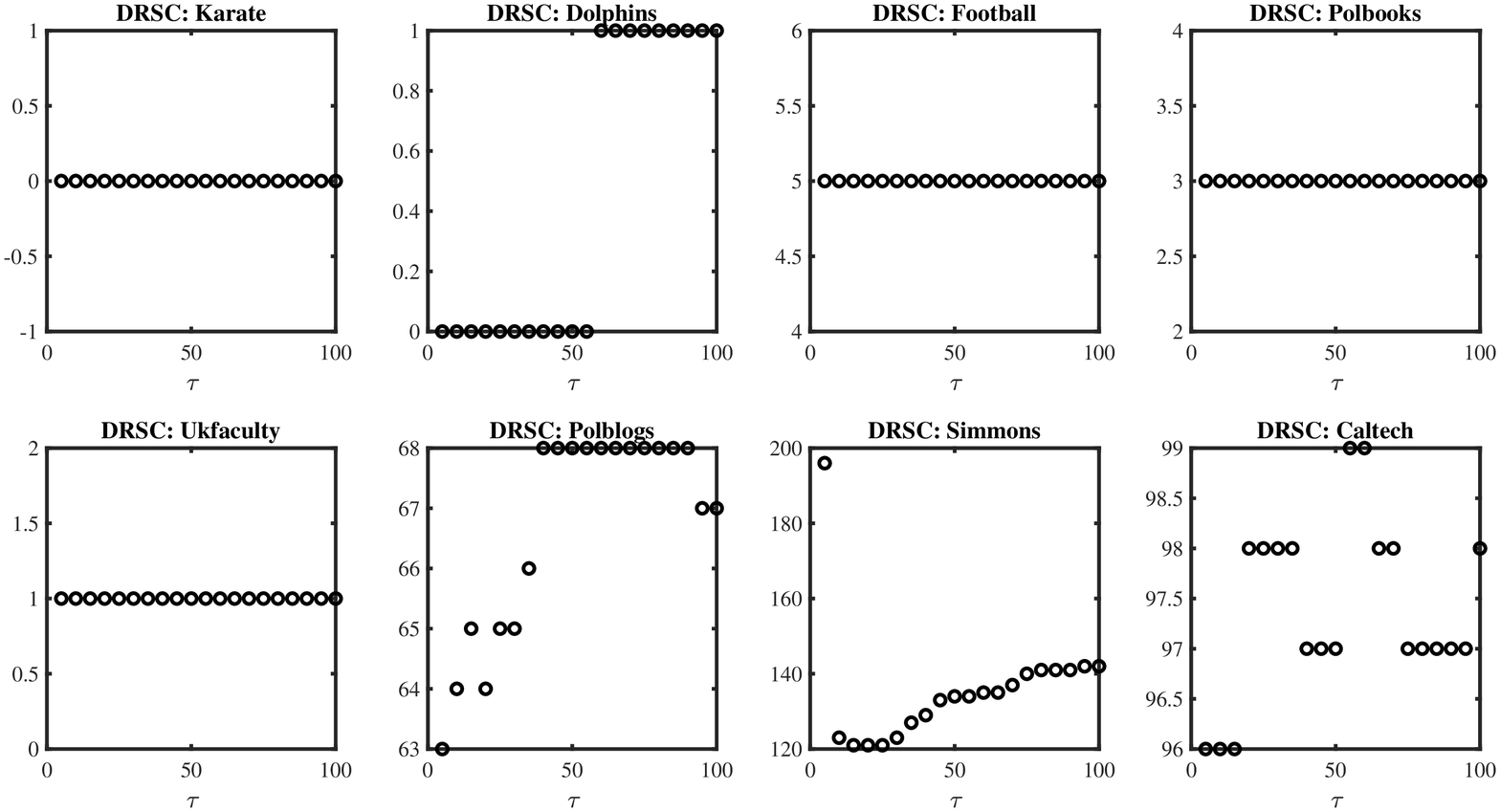}
\includegraphics[width=12cm,height=6cm]{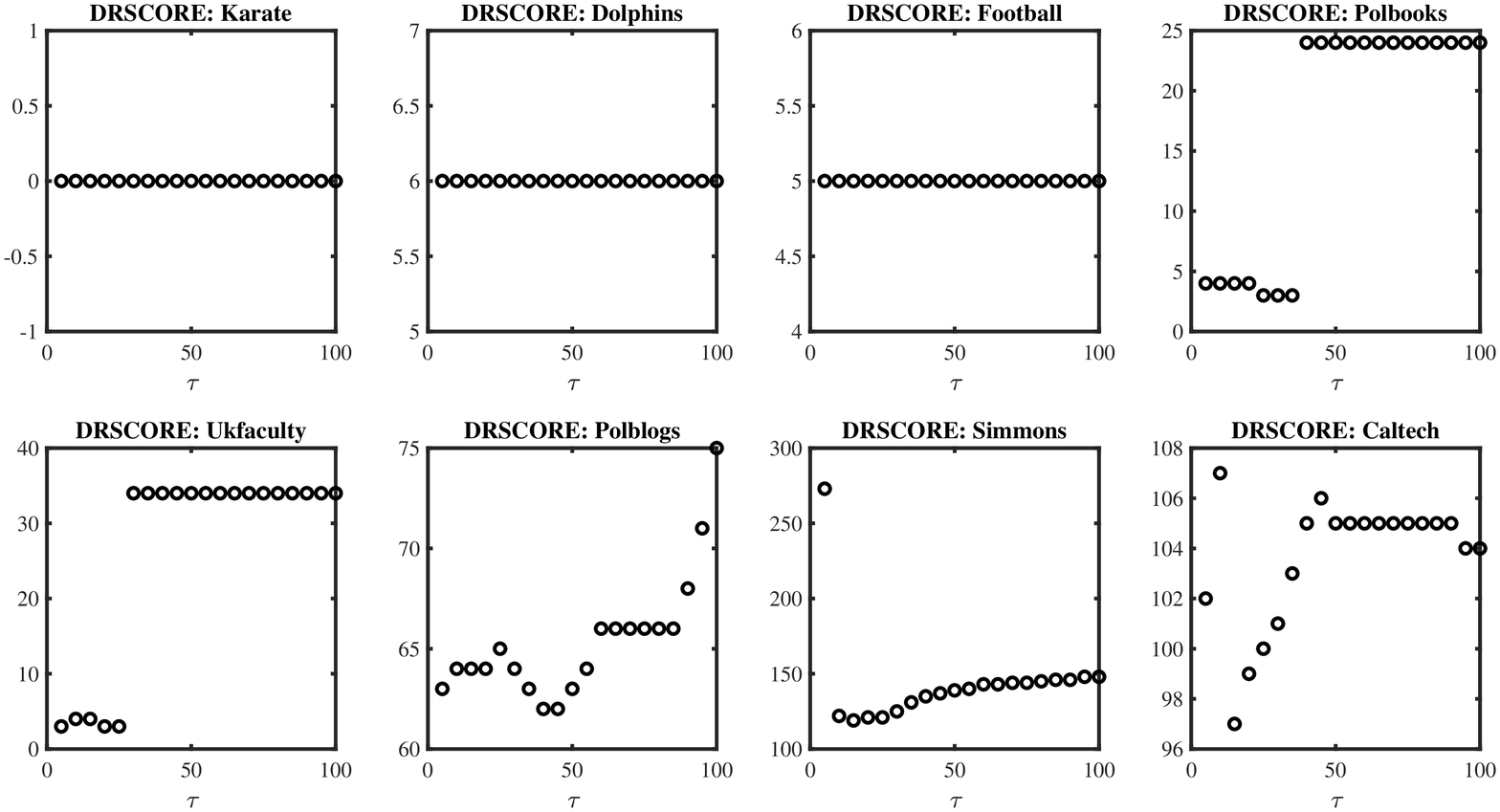}
\includegraphics[width=12cm,height=6cm]{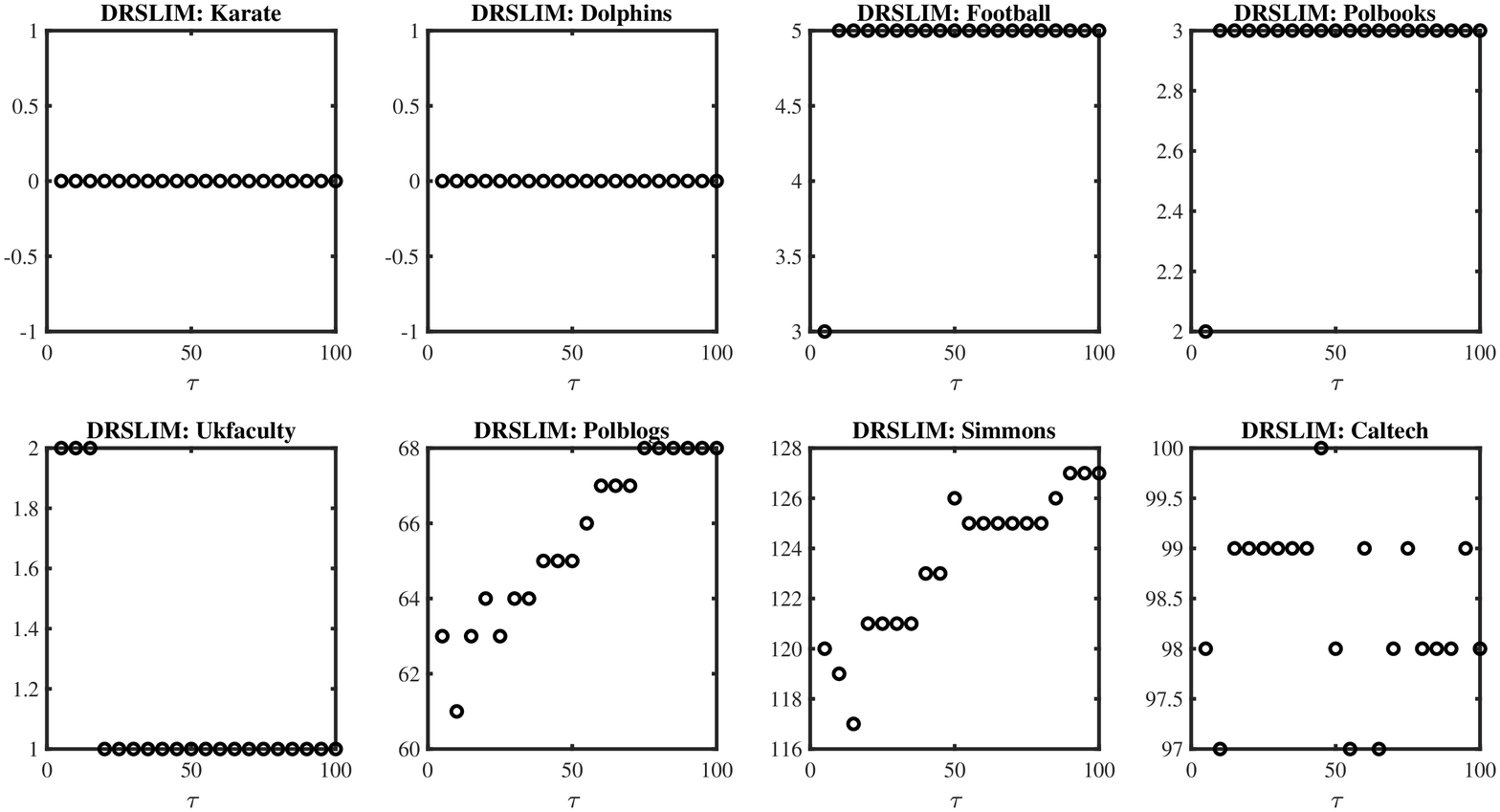}
\caption{Number errors on the eight empirical data sets for DRSC, DRSCORE and DRSLIM when $\tau$ is in $\{5,10, 15\ldots, 100\}$. x-axis: $\tau$. y-axis: number errors.}
\label{tau1tau2_DR}
\end{figure}

We also consider the cases when $\tau$ is a large number, that is, set $\tau_{1}=\tau_{2}=\tau \in \{5,10,\ldots, 100\}$, and the number errors of DRSC, DRSCORE and DRSLIM  on the eight empirical data sets are shown in Figure \ref{tau1tau2_DR}. We can find that DRSC and DRSLIM perform well with small number errors on the eight real-world networks when $\tau$ is slightly larger than 1. Unfortunately, DRSCORE is still sensitive to the choice of $\tau$ since it has larger number errors for Polbooks and UKfaculty even when $\tau$ is set quite large.  Combine the numerical results shown by Figure \ref{tau1tau2_DR1} and Figure \ref{tau1tau2_DR}, we can find that our DRSC and DRSLIM are insensitive to the choice of $\tau_{1}$ and $\tau_{2}$ when $\tau_{1}$ equals to $\tau_{2}$ as long as they are slightly larger than 1 while DRSCORE is sensitive to the choice of $\tau_{1}$ and $\tau_{2}$.

\begin{figure}%[H]
	\centering
\includegraphics[width=12cm,height=8cm]{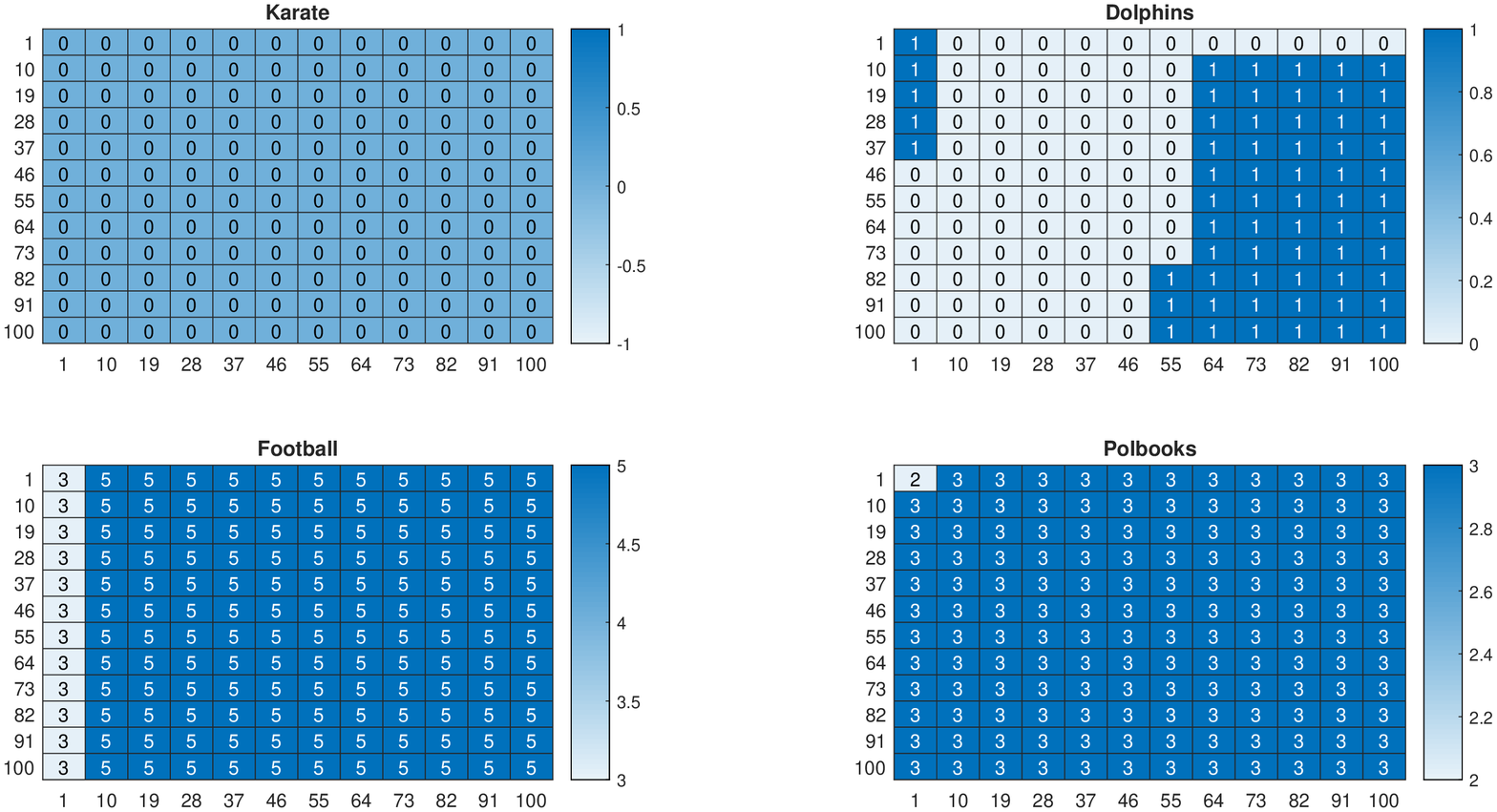}
\hspace{-2in}\\
\hspace{-2in}\\
\hspace{-2in}\\
\includegraphics[width=12cm,height=8cm]{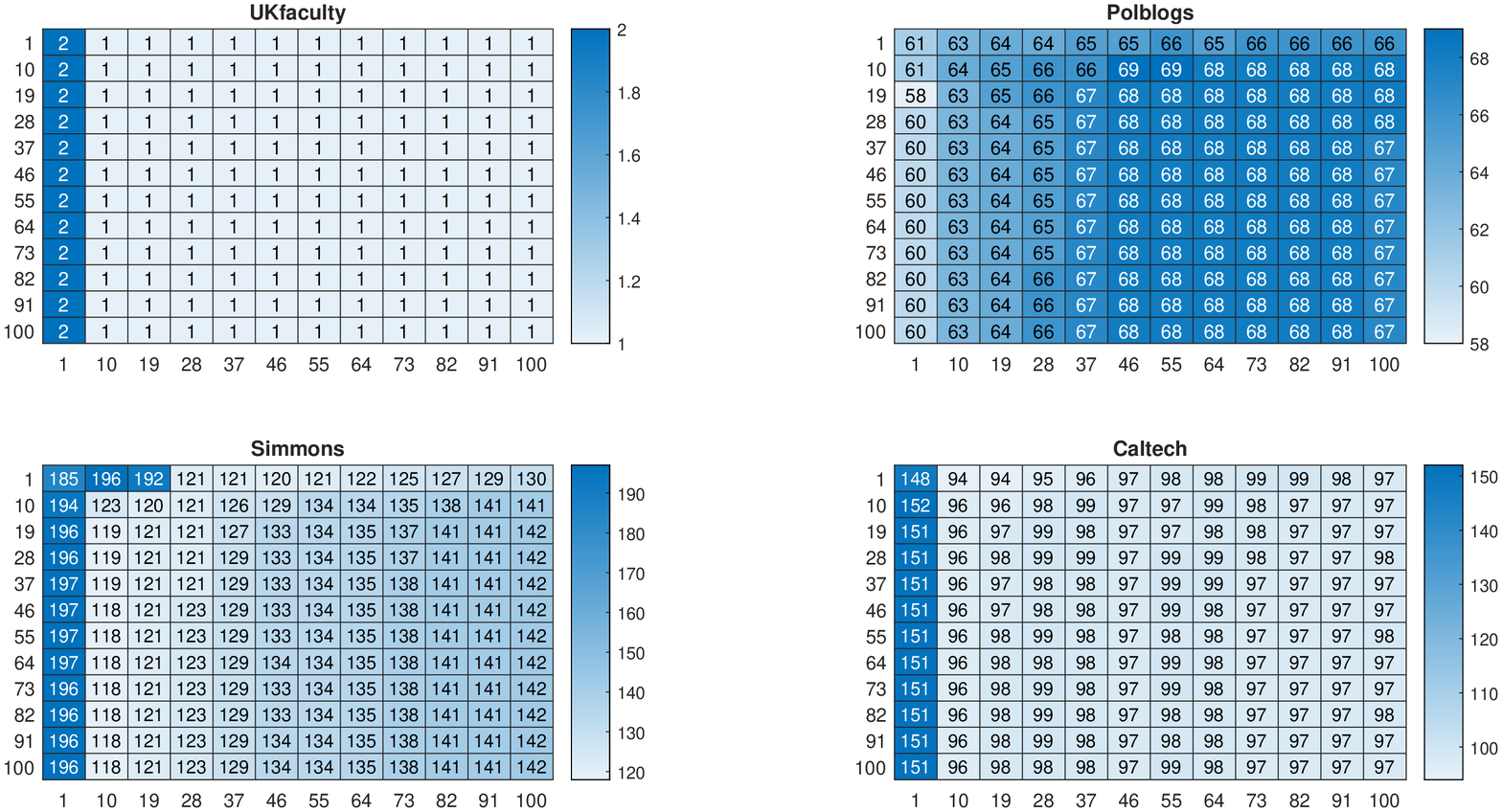}
\caption{The effect of $\tau_{1}$ and $\tau_{2}$ on the performances of DRSC for the eight empirical networks. x-axis: $\tau_{1}$. y-axis: $\tau_{2}$.}
\label{HeatmaptrueDRSC}
\end{figure}
Similar as Figure \ref{HeatmapdisL}, we can obtain the heatmap of number errors for the eight real-world datasets against $\tau_{1}$ and $\tau_{2}$, and the numerical results for DRSC are shown in Figure \ref{HeatmaptrueDRSC}. From Figure \ref{HeatmaptrueDRSC}, we can find that DRSC is insensitive to the choice of $\tau_{1}$ and $\tau_{2}$ as long as $\tau_{1}$ is slightly larger, since DRSC performs poor on Simmons when $\tau_{1}$ is 1 or 10, and DRSC performs unsatisfactory on Caltech when $\tau_{1}$ is 1. That is, DRSC is more sensitive on $\tau_{1}$ than on $\tau_{2}$, which is consistent with the findings in Figure \ref{HeatmapdisL}. The heatmaps of number errors for DRSCORE and DRSLIM are shown in Figure \ref{HeatmaptrueDRSCORE} and Figure \ref{HeatmaptrueDRSLIM} in Appendix \ref{HeadmapDRSCOEandDRSLIM}, which tell us that DRSCORE is sensitive to the choices of $\tau_{1}$ and $\tau_{2}$  while DRSLIM is insensitive.

For a conclusion of the above analysis on the choice of $\tau_{1}$ and $\tau_{2}$ for DRSC and DRSLIM, a safe choice for $\tau_{1}$ is the average degree for the given network (i.e., $\tau_{1}=\sum_{i,j=1}^{n}A_{ij}/n$), as suggested in \cite{RSC}, since the average degree for sparse networks are always lager than 1. Meanwhile, since DRSC and DRSLIM are insensitive to the choice of $\tau_{2}$ as long as it is positive, for convenience, we suggest that set $\tau_{2}$ as $\sum_{i,j=1}^{n}L_{\tau_{1}}(i,j)/n$ for DRSC and DRSLIM. As for DRSCORE, substantial numerical results show that DRSCORE has satisfactory numerical performances when setting $\tau_{1}=\sum_{i,j}A(i,j), \tau_{2}=\frac{\sum_{i,j}L_{\tau_{1}}(i,j)}{nK}$.
\subsection{Discussion on multiple regularized spectral clustering methods MRSC, MRSCORE and MRSLIM}
We study the performances of multiple regularized spectral clustering methods. Recall that our DRSC, DRSCORE and DRSLIM are designed based on dual regularized Laplacian matrix, an interesting idea comes naturally, we can design multiple  regularized spectral clustering methods and give respective theoretical framework (in Appendix \ref{PolMRSCMRSCORE}) just as our DRSC, DRSCORE and DRSLIM procedures.

Before introducing multiple regularized spectral methods, first we define the $M$-th \footnote{Without causing confusion with the matrix $M$ defined in the Ideal DRSLIM algorithm, here we use $M$ to denote a positive integer for multiple regularization.} regularized Laplacian matrix $L_{\tau_{M}}$ by iteration as:
\begin{align*}
  L_{\tau_{M}}=D_{\tau_{M}}^{-1/2}L_{\tau_{M-1}}D_{\tau_{M}}^{-1/2}, M=1,2,3,\ldots,
\end{align*}
where $L_{\tau_{0}}=A, D_{\tau_{M}}=D_{M}+\tau_{M}I$, $D_{M}$ is an $n\times n$ diagonal matrix whose $i$-th diagonal entry is given by $D_{M}(i,i)=\sum_{j}L_{\tau_{M-1}}(i,j)$.

First, we introduce the $M$-th multiple regularized spectral clustering (MRSC) method designed as below:

\textbf{MRSC}. Input:  $A, K$, and regularizers $\tau_{1}, \tau_{2},\ldots, \tau_{M}$. Output: nodes labels.

\textbf{Step 1}: Obtain the $M$-th graph Laplacian matrix $L_{\tau_{M}}$.

\textbf{Step 2}: Obtain the matrix of the product of the leading $\textbf{K+1}$ eigenvectors with unit-norm and the leading $\textbf{K+1}$ eigenvalues of  $L_{\tau_{M}}$ by
\begin{align*}
  \hat{X}=[\hat{\eta}_{1},\hat{\eta}_{2}, \ldots, \hat{\eta}_{K}, \hat{\eta}_{K+1}]\cdot \mathrm{diag}(\hat{\lambda}_{1}, \hat{\lambda}_{2}, \ldots, \hat{\lambda}_{K}, \hat{\lambda}_{K+1}).
\end{align*}
where $\hat{\lambda}_{i}$ is the $i$-th leading eigenvalue of $L_{\tau_{M}}$, and $\hat{\eta}_{i}$ is the respective eigenvector with unit-norm.

\textbf{Step 3}: Obtain $\hat{X}^{*}$ by normalizing each of $\hat{X}$' rows to have unit length.

\textbf{Step 4}: Apply K-means to $\hat{X}^{*}$,  assuming there are $K$ clusters.

Note that for MRSC, the default regularizer $\tau_{m}=\sum_{i,j}^{n}L_{\tau_{m-1}(i,j)}/n$ for $1\leq m\leq M$.

When $M$ is 2, the 2RSC is actually our DRSC approach. And when $M$ is 1, the 1RSC method is the regularized spectral clustering. Table \ref{MRSC} records number errors for the eight real-world datasets of MRSC (M= 1, 2, \ldots, 10) approaches.
\begin{table}[h!]
%\scriptsize
\centering
\caption{Error rates on the eight empirical datasets for MRSC when M=1, 2, \ldots, 10.}
\label{MRSC}
\resizebox{\columnwidth}{!}{
\begin{tabular}{cccccccccc}
\toprule
\textbf{Methods} &Karate&Dolphins&Football&Polbooks&UKfaculty&Polblogs&Simmons&Caltech\\
\midrule
1RSC&0/34&0/62&5/110&3/92&1/79&66/1222&133/1137&97/590\\
DRSC&0/34&1/62&5/110&3/92&2/79&63/1222&124/1137&95/590\\
3RSC&0/34&1/62&6/110&2/92&2/79&57/1222&193/1137&94/590\\
4RSC&3/34&1/62&3/110&2/92&2/79&60/1222&184/1137&93/590\\
5RSC&3/34&1/62&3/110&2/92&2/79&61/1222&182/1137&151/590\\
6RSC&3/34&1/62&6/110&2/92&2/79&64/1222&266/1137&151/590\\
7RSC&3/34&1/62&3/110&2/92&2/79&588/1222&266/1137&153/590\\
8RSC&3/34&1/62&3/110&1/92&2/79&588/1222&264/1137&159/590\\
9RSC&3/34&1/62&6/110&1/92&2/79&588/1222&264/1137&212/590\\
10RSC&3/34&1/62&3/110&1/92&2/79&588/1222&264/1137&204/590\\
\bottomrule
\end{tabular}}
\end{table}

From Table \ref{MRSC},  we can find that 1) all MRSC approaches perform similarly on the five networks Karate, Dolphins, Football, Polbooks and UKfaculty. 2) though 3RSC performs best on Polblogs with  error rates 57/1222 and 3RSC performs satisfactory on Caltech with error rates 94/590, it has poor performances on Simmons. 4RSC has similar performances as 3RSC, it also performs good on Polblogs and Caltech, while fails to detect Caltech. 3) Though 5RSC and 6 RSC performs satisfactory on Polblogs, the two approaches detect with high error rates for Simmons and Caltech even though they apply $(K+1)$ eigenvectors to construct $\hat{X}$ for clustering. 4) 7RSC, 8RSC, 9RSC and 10RSC fail to detect Polblogs, Simmons and Caltech. 5) 1RSC is the regular regularized spectral clustering, and it performs satisfactory on all the eight real-world datasets. When compared the performances of 1RSC with our DRSC, we can find that our DRSC outperforms 1RSC. Therefore, by the above analysis of Table \ref{MRSC}, generally speaking, our DRSC (the dual regularized spectral clustering method) has the best performances among all MRSC approaches. Though we can build respective theoretical frameworks for MRSC (when M is 3, 4, 5, etc.) following similar theoretical analysis procedures of DRSC  , it is tedious to build full theoretical framework for MRSC. Furthermore, it is unnecessary to find the theoretical bound for MRSC when M is large since our DRSC has the best performances. For reader's reference, we provide the population analysis for MRSC in Appendix \ref{PolMRSCMRSCORE} to show that MRSC returns perfect under the ideal case.

Then we introduce the $M$-th multiple regularized spectral clustering on ratios-of-eigenvectors (MRSCORE) method designed as follows:

\textbf{MRSCORE}. Input:  $A, K$, and regularizers $\tau_{1}, \tau_{2},\ldots, \tau_{M}$. Output: nodes labels.

\textbf{Step 1}: Obtain $L_{\tau_{M}}$.

\textbf{Step 2}: Obtain the matrix of the product of the leading $\textbf{K+1}$ eigenvectors with unit-norm and the leading $\textbf{K+1}$ eigenvalues of  $L_{\tau_{M}}$ by
\begin{align*}
  \hat{X}=[\hat{\eta}_{1},\hat{\eta}_{2}, \ldots, \hat{\eta}_{K}, \hat{\eta}_{K+1}]\cdot \mathrm{diag}(\hat{\lambda}_{1}, \hat{\lambda}_{2}, \ldots, \hat{\lambda}_{K}, \hat{\lambda}_{K+1}).
\end{align*}

\textbf{Step 3}: Obtain the $n\times \textbf{K}$ matrix $\hat{R}$ of entry-wise eigen-ratios such that $\hat{R}(i,k)=\hat{X}_{k+1}(i)/\hat{X}_{1}(i), 1\leq i\leq n, 1\leq k\leq K$.

\textbf{Step 4}: Apply K-means to $\hat{R}$,  assuming there are $K$ clusters.

Note that for MRSCORE, the default regularizer $\tau_{m}=\sum_{i,j}^{n}L_{\tau_{m-1}(i,j)}$ for $1\leq m\leq M-1$, and the default $\tau_{M}=\frac{\sum_{i,j}^{n}L_{\tau_{M-1}}}{nK}$. Numerical results show that MRSCORE is sensitive to the choice of regularizers, therefore we recommend applying the default regularizers for MRSCORE.

When $M$ is 2, the 2RSCORE is actually our DRSCORE approach. Table \ref{MRSCORE} records number errors for the eight real-world datasets of MRSCORE (M= 1, 2, \ldots, 6) approaches. From Table \ref{MRSCORE}, we see that MRSCORE performs similar when $M$ is different when dealing with the eight empirical datasets.
\begin{table}[h!]
\scriptsize
\centering
\caption{Error rates on the eight empirical datasets for MRSCORE when M=1, 2, \ldots, 6.}
\label{MRSCORE}
\resizebox{\columnwidth}{!}{
\begin{tabular}{cccccccccc}
\toprule
\textbf{Methods} &Karate&Dolphins&Football&Polbooks&UKfaculty&Polblogs&Simmons&Caltech\\
\midrule
1RSCORE&0/34&4/62&6/110&4/92&3/79&65/1222&117/1137&99/590\\
DRSCORE&0/34&4/62&5/110&4/92&3/79&65/1222&117/1137&99/590\\
3RSCORE&0/34&4/62&6/110&4/92&3/79&64/1222&117/1137&99/590\\
4RSCORE&0/34&4/62&3/110&4/92&3/79&64/1222&117/1137&99/590\\
5RSCORE&2/34&4/62&3/110&4/92&3/79&64/1222&117/1137&98/590\\
6RSCORE&2/34&4/62&3/110&4/92&3/79&64/1222&117/1137&99/590\\
\bottomrule
\end{tabular}}
\end{table}

Finally, we introduce the $M$-th multiple regularized symmetric Laplacian inverse matrix (MRSLIM) method designed as below:

\textbf{MRSLIM}. Input:  $A, K$, and regularizers $\tau_{1}, \tau_{2},\ldots, \tau_{M}$. Output: nodes labels.

\textbf{Step 1}: Obtain $L_{\tau_{M}}$.

\textbf{Step 2}: Obtain $\hat{W}=(I-e^{-\gamma}D^{-1}_{\tau_{M}}L_{\tau_{M}})^{-1}$. Set $\hat{M}=(\hat{W}+\hat{W}')/2$ and force $\hat{M}$'s diagonal entries to be 0.

\textbf{Step 3}: Obtain $\check{X}$ by $\check{X}=[\check{\eta}_{1},\check{\eta}_{2}, \ldots, \check{\eta}_{K+1}, \check{\eta}_{K+2}]\cdot \mathrm{diag}(\check{\lambda}_{1}, \check{\lambda}_{2}, \ldots, \check{\lambda}_{K+1}, \check{\lambda}_{K+2})$,
where $\check{\lambda}_{i}$ is the $i$-th leading eigenvalue of $\hat{M}$, and $\check{\eta}_{i}$ is the respective eigenvector with unit-norm.

\textbf{Step 4}: Obtain $\check{X}^{*}$ by normalizing each of $\check{X}$' rows to have unit length.

\textbf{Step 5}: Apply K-means to $\check{X}^{*}$,  assuming there are $K$ clusters.

Note that for MRSLIM, the default regularizer $\tau_{m}=\sum_{i,j}^{n}L_{\tau_{m-1}(i,j)}/n$ for $1\leq m\leq M$.

When $M$ is 2, the 2RSLIM is actually our DRSLIM approach. Table \ref{MRSLIM} records number errors for the eight real-world datasets of MRSLIM (M= 1, 2, \ldots, 6) approaches. From table \ref{MRSLIM}, we see that our DRSLIM almost always has the best performances among all MRSLIM approaches.
\begin{table}[h!]
\scriptsize
\centering
\caption{Error rates on the eight empirical datasets for MRSLIM when M=1, 2, \ldots, 6.}
\label{MRSLIM}
\resizebox{\columnwidth}{!}{
\begin{tabular}{cccccccccc}
\toprule
\textbf{Methods} &Karate&Dolphins&Football&Polbooks&UKfaculty&Polblogs&Simmons&Caltech\\
\midrule
1RSLIM&0/34&0/62&3/110&2/92&2/79&58/1222&121/1137&96/590\\
DRSLIM&0/34&0/62&3/110&2/92&2/79&59/1222&115/1137&98/590\\
3RSLIM&0/34&1/62&3/110&2/92&2/79&55/1222&122/1137&99/590\\
4RSLIM&0/34&1/62&3/110&2/92&2/79&58/1222&123/1137&97/590\\
5RSLIM&0/34&1/62&3/110&2/92&2/79&60/1222&277/1137&90/590\\
6RSLIM&0/34&1/62&3/110&5/92&2/79&61/1222&276/1137&100/590\\
\bottomrule
\end{tabular}}
\end{table}
\section{Discussion}\label{sec6}
In this paper, we give theoretical analysis, simulation and empirical results that demonstrate how a simple adjustment to the traditional spectral clustering methods RSC, SCORE and recently published spectral clustering method SLIM can give dramatically better results for community detection. %By applications of our new dual regularized Laplacian matrix $L_{\tau_{2}}$ to the traditional spectral clustering methods RSC and SCORE as well as the recent method LSIM, we design three spectral clustering methods DRSC, DRSCORE and DRSLIM. DRSC and DRSCORE are designed based on the production of the leading $(K+1)$ eigenvectors and eigenvalues of $L_{\tau_{2}}$, where there is a row-normalization step in DRSC and a entry-wise ratio step in  DRSCORE. For DRSLIM, we obtain the symmetric Laplacian inverse matrix $\hat{M}$ based on $L_{\tau_{2}}$, and then apply K-means on the row-normalization of the production of the leading $(K+2)$ eigenvectors and eigenvalues of $\hat{M}$ for clustering. Theoretical results show that DRSC and DRSLIM yields stable consistent community detection under mild conditions and DRSCORE ruturns perfect clustering under the ideal case.  Our methods DRSC, DRSCORE and DRSLIM outperform traditional spectral clustering methods (such as RSC, SCORE, and OCCAM) and the recent method SLIM both in simulation and empirical studies, especially for two weak signal networks Simmons and Caltech. Though there are two ridge regularizer $\tau_{1}$ and $\tau_{2}$ in these three procedures, numerical results show that our DRSC and DRSLIM are insensitive to the choice of $\tau_{1}$ and $\tau_{2}$ as long as they are slightly larger than 0 while DRSCORE is slightly sensitive to the choice of the two regularizers. Meanwhile, our three methods are designed based on the dual regularized Laplacian matrix, we also design MRSC, MRSCORE and MRSLIM approaches based on multiple regularized Laplacian matrix. Numerical results show that our DRSC and DRSLIM have the best performances among all MRSC and MRSLIM approaches, suggesting that it is unnecessary to independently study the theoretical and numerical performances of MRSC and MRSLIM for the case that when $M$ is larger than 2 while MRSCORE performs similar for all $M$.
There are several open problems for future work. For example, study the theoretical optimal values for $\tau_{1}$, $\tau_{2}$ and $\gamma$. Find the relationship between the eigenvalues of $\mathscr{L}_{\tau_{2}}$ (and $M$) and the two regularizers $\tau_{1}, \tau_{2}$. Since we only provide population analysis for DRSCORE and it is sensitive to the choice of $\tau_{1}$ and $\tau_{2}$, it is meaningful to build full theoretical framework for DRSCORE to study the question that whether there exists optimal regularizers for DRSCORE. Because numerical results show that our DRSC and DRSLIM can successfully detect dis-associative networks while we assume that the leading $K$ eigenvalues of $\mathscr{L}_{\tau_{2}}$ and $M$ should be positive in the theoretical analysis for DRSC and DRSLIM, it is meaningful to extend the Davis-Kahan theorem to disjoint sets of $S$ (defined in Lemma \ref{QinDK}). As in \cite{SLIM}, it is interesting to study the consistency of DRSC, DRSCORE and DRSLIM for both sparse networks and dense networks. Though we design three approaches DRSC, DRSCORE, DRSLIM based on the application of the dual regularized Laplacian matrix $L_{\tau_{2}}$ and build theoretical framework for them as well as investigate their performances via comparing with several spectral clustering methods, it is still unsolved that whether dual regularization can always provide with better performances than ordinary regularization in both theoretical analysis and numerical guarantees. Finally, it is interesting to study the question that whether there exists optimal $\alpha_{\mathrm{op}}$ and $\beta_{\mathrm{op}}$ such that spectral clustering methods based on the leading eigenvectors and eigenvalues of the matrix $D_{\tau_{2}}^{-\alpha_{\mathrm{op}}}L_{\tau_{1}}^{\beta_{\mathrm{op}}}D_{\tau_{2}}^{-\alpha_{\mathrm{op}}}$ outperform methods designed based on the matrix $D_{\tau_{2}}^{-\alpha}L_{\tau_{1}}^{\beta}D_{\tau_{2}}^{-\alpha}$ for any $\alpha$ and $\beta$.  We leave studies of these problems to our future work.
\bibliographystyle{Chicago}
\bibliography{reference}

\begin{thebibliography}{}

\bibitem[\protect\citeauthoryear{Adamic and Glance}{Adamic and
  Glance}{2005}]{Polblogs1}
Adamic, L.~A. and N.~Glance (2005).
\newblock The political blogosphere and the 2004 us election: divided they
  blog.
\newblock pp.\  36--43.

\bibitem[\protect\citeauthoryear{Airoldi, Blei, Fienberg, and Xing}{Airoldi
  et~al.}{2008}]{airoldi2008mixed}
Airoldi, E.~M., D.~M. Blei, S.~E. Fienberg, and E.~P. Xing (2008).
\newblock Mixed membership stochastic blockmodels.
\newblock {\em Journal of Machine Learning Research\/}~{\em 9}, 1981--2014.

\bibitem[\protect\citeauthoryear{{Amini}, {Chen}, {Bickel}, and
  {Levina}}{{Amini} et~al.}{2013}]{amini2013pseudo}
{Amini}, A.~A., A.~{Chen}, P.~J. {Bickel}, and E.~{Levina} (2013).
\newblock Pseudo-likelihood methods for community detection in large sparse
  networks.
\newblock {\em Annals of Statistics\/}~{\em 41\/}(4), 2097--2122.

\bibitem[\protect\citeauthoryear{{Bickel} and {Chen}}{{Bickel} and
  {Chen}}{2009a}]{PJBAC}
{Bickel}, P.~J. and A.~{Chen} (2009a).
\newblock {A nonparametric view of network models and Newman–Girvan and other
  modularities}.
\newblock {\em Proceedings of the National Academy of Sciences of the United
  States of America\/}~{\em 106\/}(50), 21068--21073.

\bibitem[\protect\citeauthoryear{{Bickel} and {Chen}}{{Bickel} and
  {Chen}}{2009b}]{bickel2009a}
{Bickel}, P.~J. and A.~{Chen} (2009b).
\newblock {A nonparametric view of network models and Newman–Girvan and other
  modularities}.
\newblock {\em Proceedings of the National Academy of Sciences of the United
  States of America\/}~{\em 106\/}(50), 21068--21073.

\bibitem[\protect\citeauthoryear{Borgatti and Everett}{Borgatti and
  Everett}{1997}]{borgatti1997network}
Borgatti, S.~P. and M.~G. Everett (1997).
\newblock Network analysis of 2-mode data.
\newblock {\em Social Networks\/}~{\em 19\/}(3), 243--269.

\bibitem[\protect\citeauthoryear{Burt}{Burt}{1976}]{burt1976positions}
Burt, R.~S. (1976).
\newblock Positions in networks.
\newblock {\em Social Forces\/}~{\em 55\/}(1), 93--122.

\bibitem[\protect\citeauthoryear{Chaudhuri, Chung, and Tsiatas}{Chaudhuri
  et~al.}{2012}]{chaudhuri2012spectral}
Chaudhuri, K., F.~Chung, and A.~Tsiatas (2012).
\newblock Spectral clustering of graphs with general degrees in the extended
  planted partition model.
\newblock pp.\  1--23.

\bibitem[\protect\citeauthoryear{Chen and Zhang}{Chen and
  Zhang}{2007}]{chen2007resistance}
Chen, H. and F.~Zhang (2007).
\newblock Resistance distance and the normalized laplacian spectrum.
\newblock {\em Discrete Applied Mathematics\/}~{\em 155\/}(5), 654--661.

\bibitem[\protect\citeauthoryear{{Chen}, {Li}, and {Xu}}{{Chen}
  et~al.}{2018}]{CMM}
{Chen}, Y., X.~{Li}, and J.~{Xu} (2018).
\newblock Convexified modularity maximization for degree-corrected stochastic
  block models.
\newblock {\em Annals of Statistics\/}~{\em 46\/}(4), 1573--1602.

\bibitem[\protect\citeauthoryear{Chung and Graham}{Chung and
  Graham}{1997}]{chung1997spectral}
Chung, F.~R. and F.~C. Graham (1997).
\newblock {\em Spectral graph theory}.
\newblock Number~92. American Mathematical Society.

\bibitem[\protect\citeauthoryear{{Daudin}, {Picard}, and {Robin}}{{Daudin}
  et~al.}{2008}]{daudin2008a}
{Daudin}, J.~J., F.~{Picard}, and S.~{Robin} (2008).
\newblock A mixture model for random graphs.
\newblock {\em Statistics and Computing\/}~{\em 18\/}(2), 173--183.

\bibitem[\protect\citeauthoryear{Dong, Thanou, Frossard, and
  Vandergheynst}{Dong et~al.}{2016}]{dong2016learning}
Dong, X., D.~Thanou, P.~Frossard, and P.~Vandergheynst (2016).
\newblock Learning laplacian matrix in smooth graph signal representations.
\newblock {\em IEEE Transactions on Signal Processing\/}~{\em 64\/}(23),
  6160--6173.

\bibitem[\protect\citeauthoryear{Doreian}{Doreian}{1985}]{doreian1985structural}
Doreian, P. (1985).
\newblock Structural equivalence in a psychology journal network.
\newblock {\em Journal of the Association for Information Science and
  Technology\/}~{\em 36\/}(6), 411--417.

\bibitem[\protect\citeauthoryear{Doreian, Batagelj, and Ferligoj}{Doreian
  et~al.}{1994}]{doreian1994partitioning}
Doreian, P., V.~Batagelj, and A.~Ferligoj (1994).
\newblock Partitioning networks based on generalized concepts of equivalence.
\newblock {\em Journal of Mathematical Sociology\/}~{\em 19\/}(1), 1--27.

\bibitem[\protect\citeauthoryear{Girvan and Newman}{Girvan and
  Newman}{2002}]{football}
Girvan, M. and M.~E. Newman (2002).
\newblock {Community structure in social and biological networks}.
\newblock {\em Proceedings of the national academy of sciences\/}~{\em
  99\/}(12), 7821--7826.

\bibitem[\protect\citeauthoryear{{Holland}, {Laskey}, and
  {Leinhardt}}{{Holland} et~al.}{1983}]{SBM}
{Holland}, P.~W., K.~B. {Laskey}, and S.~{Leinhardt} (1983).
\newblock Stochastic blockmodels: First steps.
\newblock {\em Social Networks\/}~{\em 5\/}(2), 109--137.

\bibitem[\protect\citeauthoryear{{Jin}}{{Jin}}{2015}]{SCORE}
{Jin}, J. (2015).
\newblock {Fast community detection by SCORE}.
\newblock {\em Annals of Statistics\/}~{\em 43\/}(1), 57--89.

\bibitem[\protect\citeauthoryear{{Jin}, {Ke}, and {Luo}}{{Jin}
  et~al.}{2018}]{SCORE+}
{Jin}, J., Z.~T. {Ke}, and S.~{Luo} (2018).
\newblock {SCORE+ for network community detection}.
\newblock {\em arXiv preprint arXiv:1811.05927\/}.

\bibitem[\protect\citeauthoryear{Jing, Li, Ying, and Yu}{Jing
  et~al.}{2021}]{SLIM}
Jing, B., T.~Li, N.~Ying, and X.~Yu (2021).
\newblock Community detection in sparse networks using the symmetrized
  laplacian inverse matrix (slim).
\newblock {\em Statistica Sinica\/}.

\bibitem[\protect\citeauthoryear{{Joseph} and {Yu}}{{Joseph} and
  {Yu}}{2016}]{joseph2016impact}
{Joseph}, A. and B.~{Yu} (2016).
\newblock Impact of regularization on spectral clustering.
\newblock {\em Annals of Statistics\/}~{\em 44\/}(4), 1765--1791.

\bibitem[\protect\citeauthoryear{{Karrer} and {Newman}}{{Karrer} and
  {Newman}}{2011}]{DCSBM}
{Karrer}, B. and M.~E.~J. {Newman} (2011).
\newblock Stochastic blockmodels and community structure in networks.
\newblock {\em Physical Review E\/}~{\em 83\/}(1), 16107.

\bibitem[\protect\citeauthoryear{Lorrain and White}{Lorrain and
  White}{1971}]{lorrain1971structural}
Lorrain, F. and H.~C. White (1971).
\newblock Structural equivalence of individuals in social networks.
\newblock {\em Journal of Mathematical Sociology\/}~{\em 1\/}(1), 49--80.

\bibitem[\protect\citeauthoryear{Lusseau}{Lusseau}{2003}]{dolphins1}
Lusseau, D. (2003).
\newblock {The emergent properties of a dolphin social network}.
\newblock {\em Proceedings of the Royal Society of London. Series B: Biological
  Sciences\/}~{\em 270\/}(suppl\_2), S186--S188.

\bibitem[\protect\citeauthoryear{Lusseau}{Lusseau}{2007}]{dolphins2}
Lusseau, D. (2007).
\newblock {Evidence for social role in a dolphin social network}.
\newblock {\em Evolutionary ecology\/}~{\em 21\/}(3), 357--366.

\bibitem[\protect\citeauthoryear{Lusseau, Schneider, Boisseau, Haase, Slooten,
  and Dawson}{Lusseau et~al.}{2003}]{dolphins0}
Lusseau, D., K.~Schneider, O.~J. Boisseau, P.~Haase, E.~Slooten, and S.~M.
  Dawson (2003).
\newblock {The bottlenose dolphin community of Doubtful Sound features a large
  proportion of long-lasting associations}.
\newblock {\em Behavioral Ecology and Sociobiology\/}~{\em 54\/}(4), 396--405.

\bibitem[\protect\citeauthoryear{{Luxburg}}{{Luxburg}}{2007}]{Tutorial}
{Luxburg}, U. (2007).
\newblock A tutorial on spectral clustering.
\newblock {\em Statistics and Computing\/}~{\em 17\/}(4), 395--416.

\bibitem[\protect\citeauthoryear{{Ma} and {Ma}}{{Ma} and {Ma}}{2017}]{LSCD}
{Ma}, Z. and Z.~{Ma} (2017).
\newblock {Exploration of large networks with covariates via fast and universal
  latent space model fitting}.
\newblock {\em arXiv preprint arXiv:1705.02372\/}.

\bibitem[\protect\citeauthoryear{{Nepusz}, {Petróczi}, {Négyessy}, and
  {Bazsó}}{{Nepusz} et~al.}{2008}]{nepusz2008fuzzy}
{Nepusz}, T., A.~{Petróczi}, L.~{Négyessy}, and F.~{Bazsó} (2008).
\newblock Fuzzy communities and the concept of bridgeness in complex networks.
\newblock {\em Physical Review E\/}~{\em 77\/}(1), 16107--16107.

\bibitem[\protect\citeauthoryear{Nepusz, Petr{\'o}czi, N{\'e}gyessy, and
  Bazs{\'o}}{Nepusz et~al.}{2008}]{UKfaculty}
Nepusz, T., A.~Petr{\'o}czi, L.~N{\'e}gyessy, and F.~Bazs{\'o} (2008).
\newblock {Fuzzy communities and the concept of bridgeness in complex
  networks}.
\newblock {\em Physical Review E\/}~{\em 77\/}(1), 016107.

\bibitem[\protect\citeauthoryear{{Newman}}{{Newman}}{2006}]{MN2006}
{Newman}, M. (2006).
\newblock Modularity and community structure in networks.
\newblock {\em Bulletin of the American Physical Society\/}.

\bibitem[\protect\citeauthoryear{Newman and Girvan}{Newman and
  Girvan}{2004}]{dolphinnewman}
Newman, M.~E. and M.~Girvan (2004).
\newblock {Finding and evaluating community structure in networks}.
\newblock {\em Physical review E\/}~{\em 69\/}(2), 026113.

\bibitem[\protect\citeauthoryear{Newman and Clauset}{Newman and
  Clauset}{2016}]{newman2016structure}
Newman, M. E.~J. and A.~Clauset (2016).
\newblock Structure and inference in annotated networks.
\newblock {\em Nature Communications\/}~{\em 7\/}(1), 11863--11863.

\bibitem[\protect\citeauthoryear{Ng, Jordan, and Weiss}{Ng
  et~al.}{2002}]{ng2002spectral}
Ng, A.~Y., M.~I. Jordan, and Y.~Weiss (2002).
\newblock On spectral clustering: Analysis and an algorithm.
\newblock {\em Advances in Neural Information Processing Systems\/}, 849--856.

\bibitem[\protect\citeauthoryear{{Qin} and {Rohe}}{{Qin} and
  {Rohe}}{2013}]{RSC}
{Qin}, T. and K.~{Rohe} (2013).
\newblock {Regularized spectral clustering under the degree-corrected
  stochastic blockmodel}.
\newblock In {\em Advances in Neural Information Processing Systems 26}, pp.\
  3120--3128.

\bibitem[\protect\citeauthoryear{Reichardt and White}{Reichardt and
  White}{2007}]{reichardt2007role}
Reichardt, J. and D.~R. White (2007).
\newblock Role models for complex networks.
\newblock {\em European Physical Journal B\/}~{\em 60\/}(2), 217--224.

\bibitem[\protect\citeauthoryear{{Snijders} and {Nowicki}}{{Snijders} and
  {Nowicki}}{1997}]{snijders1997estimation}
{Snijders}, T. A.~B. and K.~{Nowicki} (1997).
\newblock {Estimation and prediction for stochastic blockmodels for graphs with
  latent block structur}.
\newblock {\em Journal of Classification\/}~{\em 14\/}(1), 75--100.

\bibitem[\protect\citeauthoryear{Tang, Zhou, Zheng, Zhang, and Sha}{Tang
  et~al.}{2019}]{tang2019dual}
Tang, C., H.~Zhou, X.~Zheng, Y.~Zhang, and X.~Sha (2019).
\newblock Dual laplacian regularized matrix completion for microrna-disease
  associations prediction.
\newblock {\em RNA Biology\/}~{\em 16\/}(5), 601--611.

\bibitem[\protect\citeauthoryear{{Traud}, {Kelsic}, {Mucha}, and
  {Porter}}{{Traud} et~al.}{2011}]{traud2011comparing}
{Traud}, A.~L., E.~D. {Kelsic}, P.~J. {Mucha}, and M.~A. {Porter} (2011).
\newblock {Comparing community structure to characteristics in online
  collegiate social network}.
\newblock {\em Siam Review\/}~{\em 53\/}(3), 526--543.

\bibitem[\protect\citeauthoryear{Traud, Mucha, and Porter}{Traud
  et~al.}{2012}]{traud2012social}
Traud, A.~L., P.~J. Mucha, and M.~A. Porter (2012).
\newblock Social structure of facebook networks.
\newblock {\em Physica A-statistical Mechanics and Its Applications\/}~{\em
  391\/}(16), 4165--4180.

\bibitem[\protect\citeauthoryear{Von~Luxburg}{Von~Luxburg}{2007}]{von2007tutorial}
Von~Luxburg, U. (2007).
\newblock A tutorial on spectral clustering.
\newblock {\em Statistics and Computing\/}~{\em 17\/}(4), 395--416.

\bibitem[\protect\citeauthoryear{Wang and Wong}{Wang and
  Wong}{1987}]{wang1987stochastic}
Wang, Y.~J. and G.~Y. Wong (1987).
\newblock Stochastic blockmodels for directed graphs.
\newblock {\em Journal of the American Statistical Association\/}~{\em
  82\/}(397), 8--19.

\bibitem[\protect\citeauthoryear{{Weyl}}{{Weyl}}{1912}]{weyl1912das}
{Weyl}, H. (1912).
\newblock Das asymptotische verteilungsgesetz der eigenwerte linearer
  partieller differentialgleichungen (mit einer anwendung auf die theorie der
  hohlraumstrahlung).
\newblock {\em Mathematische Annalen\/}~{\em 71\/}(4), 441--479.

\bibitem[\protect\citeauthoryear{Xiao, Luo, Liang, Cai, and Ding}{Xiao
  et~al.}{2018}]{xiao2018graph}
Xiao, Q., J.~Luo, C.~Liang, J.~Cai, and P.~Ding (2018).
\newblock A graph regularized non-negative matrix factorization method for
  identifying microrna-disease associations.
\newblock {\em Bioinformatics\/}~{\em 34\/}(2), 239--248.

\bibitem[\protect\citeauthoryear{Yankelevsky and Elad}{Yankelevsky and
  Elad}{2016}]{yankelevsky2016dual}
Yankelevsky, Y. and M.~Elad (2016).
\newblock Dual graph regularized dictionary learning.
\newblock {\em IEEE Transactions on Signal and Information Processing over
  Networks\/}~{\em 2\/}(4), 611--624.

\bibitem[\protect\citeauthoryear{Zachary}{Zachary}{1977}]{karate}
Zachary, W.~W. (1977).
\newblock An information flow model for conflict and fission in small groups.
\newblock {\em Journal of Anthropological Research\/}~{\em 33\/}(4), 452--473.

\bibitem[\protect\citeauthoryear{{Zhang}, {Levina}, and {Zhu}}{{Zhang}
  et~al.}{2020}]{OCCAM}
{Zhang}, Y., E.~{Levina}, and J.~{Zhu} (2020).
\newblock {Detecting overlapping communities in networks using spectral
  methods}.
\newblock {\em SIAM Journal on Mathematics of Data Science\/}~{\em 2\/}(2),
  265--283.

\end{thebibliography}
%\appendix
\begin{appendices}
\section{Heatmaps for DRSCORE and DRSLIM}\label{HeadmapDRSCOEandDRSLIM}
\begin{figure}[H]
\centering
\includegraphics[width=12cm,height=8cm]{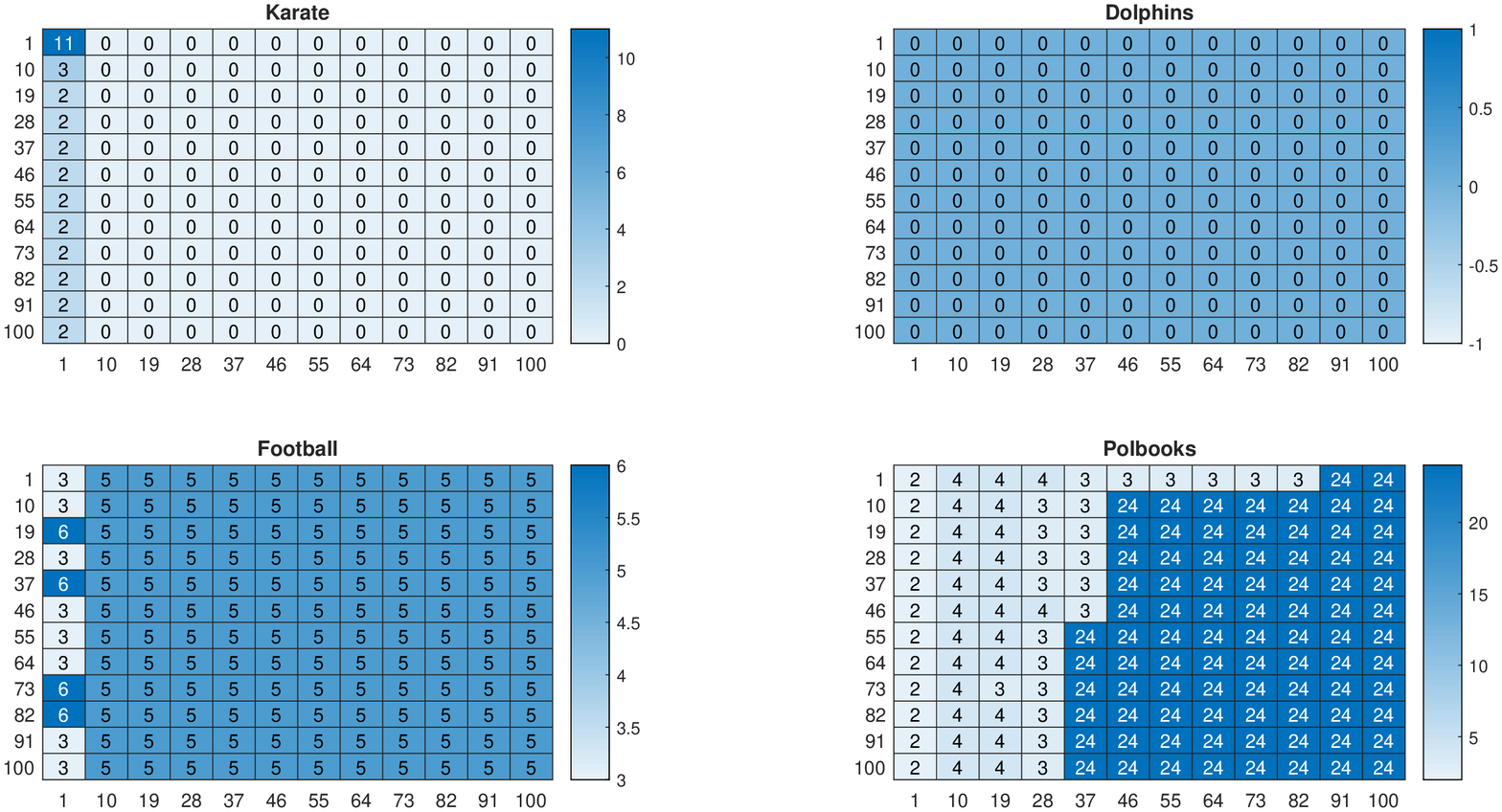}
\hspace{-2in}\\
\hspace{-2in}\\
\hspace{-2in}\\
\includegraphics[width=12cm,height=8cm]{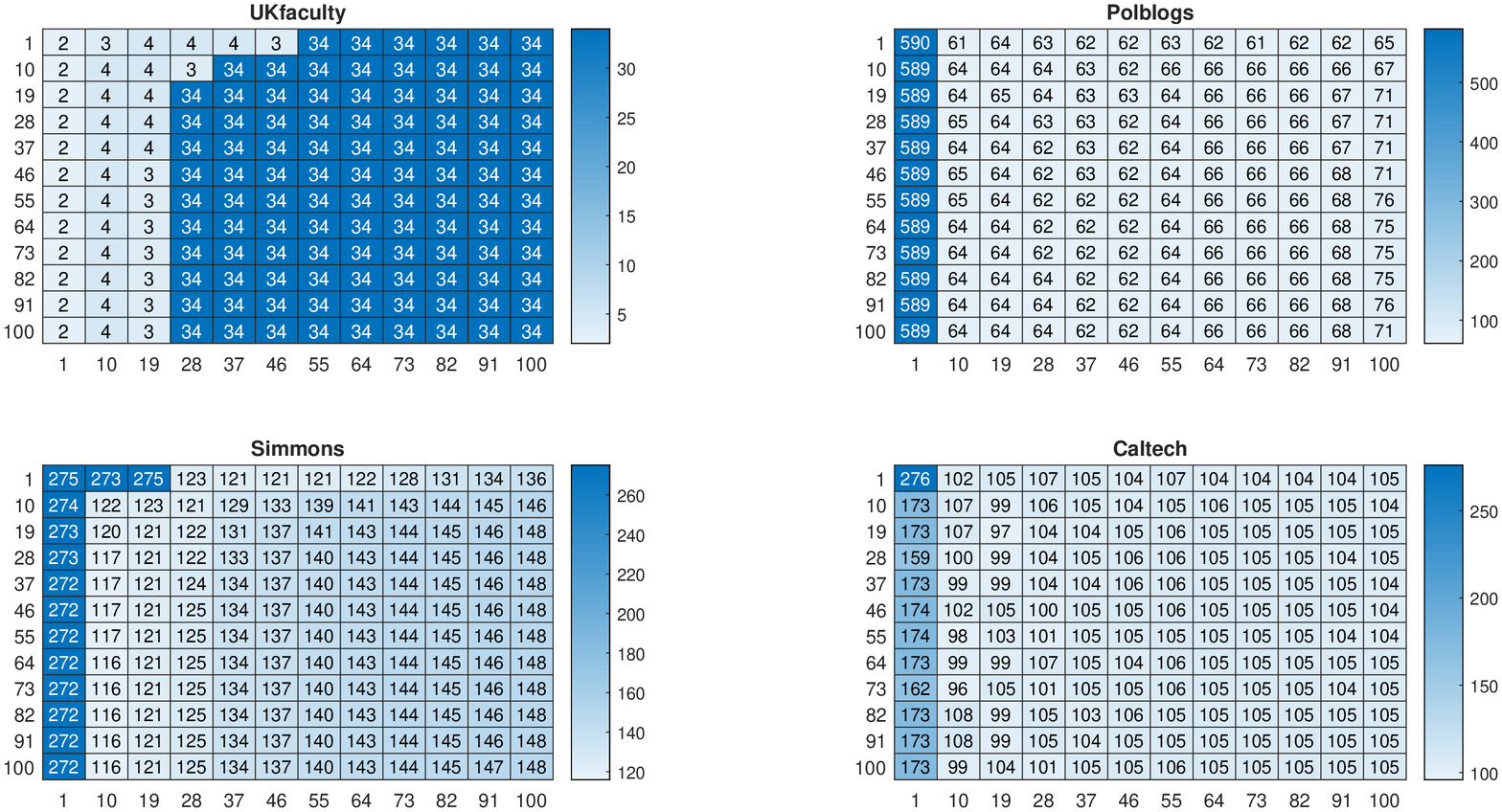}
\caption{The effect of $\tau_{1}$ and $\tau_{2}$ on the performances of DRSCORE for the eight empirical networks. x-axis: $\tau_{1}$. y-axis: $\tau_{2}$.}
\label{HeatmaptrueDRSCORE}
\end{figure}
\begin{figure}[H]
\centering
\includegraphics[width=12cm,height=8cm]{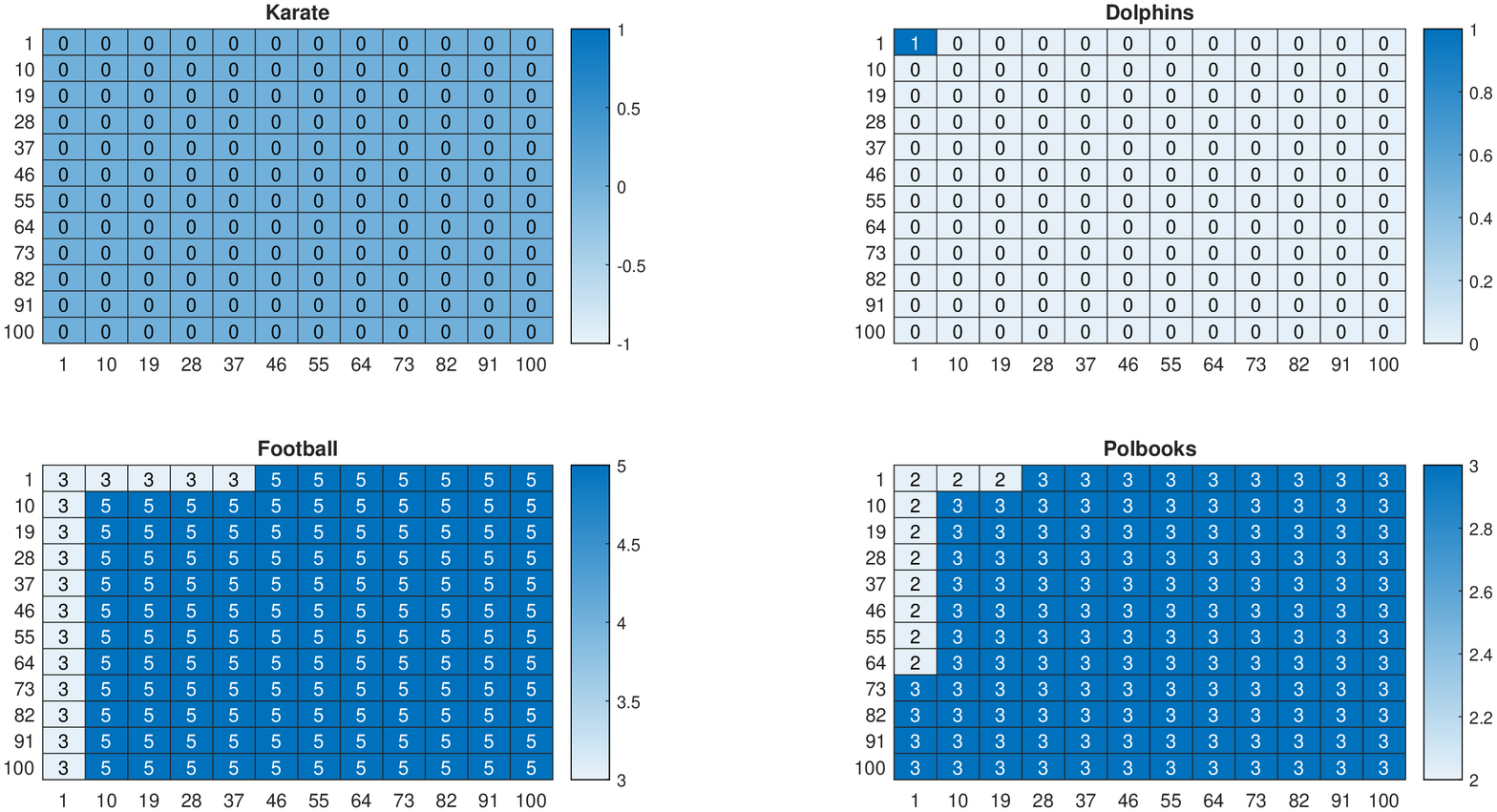}
\hspace{-2in}\\
\hspace{-2in}\\
\hspace{-2in}\\
\includegraphics[width=12cm,height=8cm]{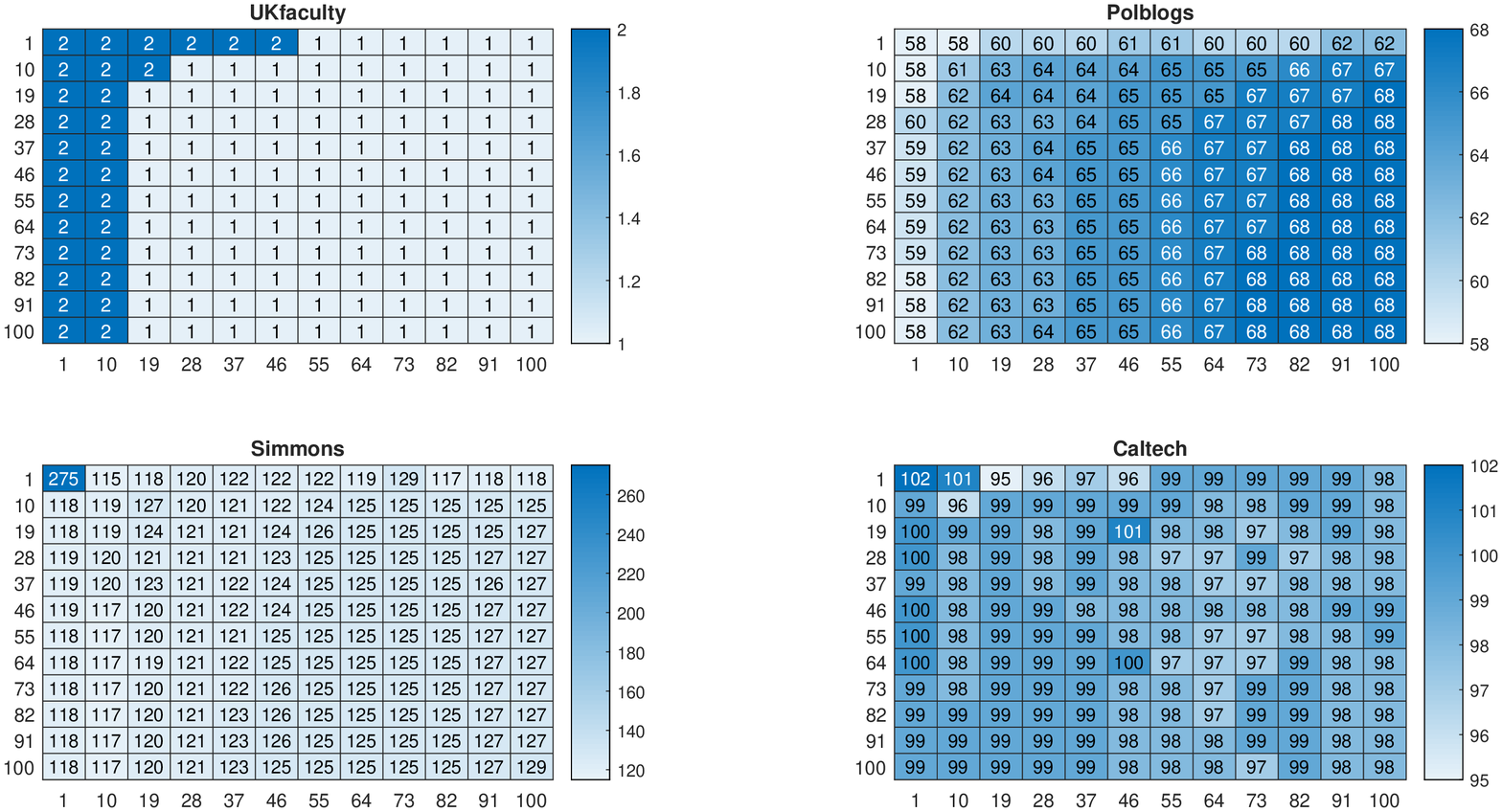}
\caption{The effect of $\tau_{1}$ and $\tau_{2}$ on the performances of DRSLIM for the eight empirical networks. x-axis: $\tau_{1}$. y-axis: $\tau_{2}$.}
\label{HeatmaptrueDRSLIM}
\end{figure}

\section{Description of eight real-word data}\label{dereal8}
\begin{itemize}
	\item \textbf{Karate}: this network consists of 34 nodes where each
	node denotes a member in the karate club \citep{karate}. As there is a conflict in the club, the network divides into two communities: Mr. Hi’s group and John’s group. \cite{karate} records all labels for each member and we use them as the true labels.
\item \textbf{Dolphins}: this network consists of frequent associations between 62 dolphins in a community living off Doubtful Sound. In Dolphins network, node denotes a dolphin, and edge stands for companionship \cite{dolphins0, dolphins1, dolphins2}. The network splits naturally into two large groups females and males \cite{dolphins1, dolphinnewman}, which are seen as the ground truth in our analysis.
\item \textbf{Football}: this network is for American
football games between Division I-A college teams during the regular football season of Fall \cite{football}. Nodes in Football denote teams and edges represent regular-season games between any two teams \cite{football}. The original network contains 115 nodes in total, since 5 of them are called ``Independent'' and the remaining 110 nodes are manually divided into 11 conferences for administration purpose, for community detection, we remove the 5 independent teams in this paper.
\item \textbf{Polbooks}: this network is about US politics
    published around the 2004 presidential election and sold by the online bookseller Amazon.com. In Polbooks, nodes represent books, edges represent frequent co-purchasing of books by the same buyers. Full information about edges and labels can be downloaded from \url{http://www-personal.umich.edu/~mejn/netdata/}. The original network contains 105 nodes labeled as either ``Conservative”, ``Liberal”, or ``Neutral”. Nodes labeled ``Neutral'' are removed for community detection in this paper.
\item \textbf{UKfaculty}: this network reflects the friendship among academic staffs of a given Faculty in a UK university consisting of three separate schools \cite{UKfaculty}. The original network contains 81 nodes, in which the smallest group only has 2 nodes. The smallest group is removed for community detection in this paper.
\item \textbf{Polblogs}: this network consists of political blogs during the
	2004 US presidential election \citep{Polblogs1}. Each blog belongs to one of the two parties liberal or
	conservative. As suggested by \cite{DCSBM}, we only consider the largest connected component with 1222  nodes and ignore the edge direction for community detection.
	\item \textbf{Simmons}: this network contains one largest connected
	component with 1137 nodes. It is observed in \cite{traud2011comparing, traud2012social} that the community structure of the Simmons College network exhibits a strong correlation with the graduation year-students since students in the same year are more likely to be friends.
	\item \textbf{Caltech}: this network has one largest connected component with 590 nodes. The community structure is highly correlated with which of the 8 dorms a user is from, as observed in \cite{traud2011comparing, traud2012social}.
\end{itemize}
\end{appendices}
\end{document}